\PassOptionsToPackage{square,comma,numbers,sort&compress}{natbib}
\documentclass{article}

\usepackage[final]{neurips_2020}

\usepackage[utf8]{inputenc} 
\usepackage[T1]{fontenc}    
\usepackage[colorlinks,bookmarks=false,urlcolor=black]{hyperref}
\usepackage{url}            
\usepackage{booktabs}       
\usepackage{amsfonts}       
\usepackage{nicefrac}       
\usepackage{microtype}      

\usepackage{amsfonts}       
\usepackage{algorithm}
\usepackage{algorithmic}
\usepackage{amsmath}
\usepackage{graphicx}
\usepackage{multirow}
\usepackage{bm}
\usepackage{color}
\usepackage{amsthm}
\usepackage{tabularx}
\usepackage{wrapfig}
\usepackage{footnote}
\usepackage{threeparttable}
\usepackage{tablefootnote}
\makesavenoteenv{tabular}
\makesavenoteenv{table}
\usepackage{caption}
\usepackage{fancyhdr}
\usepackage{amssymb}
\usepackage{pifont}
\usepackage{xr}
\usepackage{wrapfig}
\usepackage{setspace} 

\usepackage{exscale}
\usepackage{relsize}
\usepackage{xr}

\newcommand{\comment}[1]{}

\newcommand{\et}{\emph{et al.}}
\newcommand{\ie}{\emph{i.e.}}
\newcommand{\eg}{\emph{e.g.}}

\newcommand{\diagop}{\operatorname{diag}}
\newcommand{\diag}[1]{\ensuremath{\diagop\left(#1\right)}}

\newtheorem{thm}{Theorem}
\newtheorem{lem}{Lemma}
\newtheorem{defn}{Definition}

\newtheorem{assum}{Assumption}
\newcommand{\led}[1]{\overset{\text{\ding{#1}}}{\leq}}

\newcommand{\lee}[1]{\overset{\text{\ding{#1}}}{=}}
\newcommand{\ged}[1]{\overset{\text{\ding{#1}}}{\geq}}

\newcommand{\Rs}[1]{{\mathbb{R}^{#1}}}

\newcommand{\step}[1]{\textbf{\noindent{#1}}}
\newcommand{\Bi}[1]{\mathcal{S}_{#1}}
\newcommand{\Sas}{\mathcal{S}\alpha\mathcal{S}}

\newcommand{\Sgd}{\textsc{Sgd}}
\newcommand{\Sgdm}{\textsc{Sgd-M}}
\newcommand{\AdaGrad}{\textsc{AdaGrad}}
\newcommand{\RMSProp}{\textsc{RMSProp}}
\newcommand{\Adam}{\textsc{Adam}}

\newcommand{\Grd}{\bm{\Omega}^{-\varepsilon^{\gamma}}}
\newcommand{\GG}{\bm{\Omega}^{-2\varepsilon^{\gamma}}}
\newcommand{\Omegas}{\bm{\Omega}}
\newcommand{\G}{\bm{\Omega}}
\newcommand{\dis}{\textsf{dis}}
\newcommand{\ells}{\ell}

\newcommand{\levyp}{{L\'evy}}
\newcommand{\levy}{L}
\newcommand{\levyi}[1]{L_{#1}}

\newcommand{\Hui}[1]{\Theta(#1)}

\newcommand{\epsi}[1]{{\varepsilon^{#1}}}

\newcommand{\jumpi}[1]{J_{#1}}
\newcommand{\psiis}[2]{\Theta(#1^{-#2})}

\newcommand{\indi}[1]{\mathbb{I}\left\{#1\right\}}
\newcommand{\lam}{\bm{\lambda}}

\newcommand{\lame}{\lambda_{\varepsilon}}
\newcommand{\ve}{v_{\varepsilon}}
\newcommand{\mue}{\mu_{\varepsilon}}
\newcommand{\Te}{T_{\varepsilon}}
\newcommand{\ttt}{t\land v_{\varepsilon} \land \sigma_{1}}
\newcommand{\ttte}{t\land T_{\varepsilon} \land \sigma_{1}}
\newcommand{\ke}{k_{\varepsilon}}
\newcommand{\te}{\Gamma}
\newcommand{\tea}{\widetilde{\Xi}}
\newcommand{\aae}{a_{\varepsilon}}

\newcommand{\rhoa}{\bar{\rho}}

\newcommand{\threslemma}{\epsilon'}
\newcommand{\clip}{\kappa_1}

\newcommand{\jumptime}[1]{\sigma_{#1}}
\newcommand{\Emi}[1]{\bm{E}_{#1}}
\newcommand{\W}{\bm{\mathcal{W}}}

\newcommand{\Resi}[1]{\text{Res}_{#1}}

\newcommand{\Mmi}[1]{M_{#1}}

\newcommand{\batchsize}{S}

\newcommand{\wmxi}[1]{\bm{\theta}_{#1}^{\bm{\xi}}}
\newcommand{\gmii}[2]{\bm{g}_{#1}^{#2}}
\newcommand{\wgti}[2]{\widehat{\bm{\theta}}_{#1}(#2)}

\newcommand{\Fbi}[1]{f_{\mathcal{S}_{#1}}(\bm{\theta}_{#1})}
\newcommand{\Fbhi}[1]{f_{\mathcal{S}_{#1}}(\widehat{\bm{\theta}}_{#1})}

\newcommand{\mmi}[1]{\bm{m}_{#1}}
\newcommand{\vmi}[1]{\bm{v}_{#1}}

\newcommand{\wgi}[1]{\widehat{\bm{\theta}}_{#1}}

\newcommand{\mgi}[1]{\widehat{\bm{m}}_{#1}}
\newcommand{\vgi}[1]{\widehat{\bm{v}}_{#1}}

\newcommand{\omegai}[1]{\omega_{#1}}
\newcommand{\mui}[1]{\mu_{#1}}
\newcommand{\varepi}[1]{\varepsilon_{#1}}

\newcommand{\betag}{\beta_1}
\newcommand{\betav}{\beta_2}

\newcommand{\thres}{\epsilon}

\newcommand{\vmax}{v_{\max}}
\newcommand{\vmin}{v_{\min}}

\newcommand{\taum}{\tau_m}

\newcommand{\smi}[1]{\widehat{\bm{s}}_{#1}}

\newcommand{\zetai}[1]{\bm{\zeta}_{#1}}
\newcommand{\xii}[1]{\bm{\xi}_{#1}}
\newcommand{\xwi}[1]{\widehat{\bm{\xi}}_{#1}}
\newcommand{\xwii}[2]{\widehat{\bm{\xi}}_{#1}^{#2}}

\newcommand{\betai}[1]{\beta_{#1}}

\newcommand{\wms}{\bm{\theta}^{*}}
\newcommand{\wmi}[1]{\bm{\theta}_{#1}}

\newcommand{\ds}{\textsf{d}}
\newcommand{\etai}[1]{\eta_{#1}}
\newcommand{\Sigmai}[1]{\bm{\Sigma}_{#1}}
\newcommand{\Sigmabi}[1]{\bar{\bm{\Sigma}}_{#1}}

\newcommand{\umi}[1]{\bm{u}_{#1}}

\newcommand{\ymi}[1]{\bm{y}_{#1}}

\newcommand{\gmi}[1]{\bm{g}^{#1}}

\newcommand{\LL}{\mathcal{L}}
\newcommand{\U}{\mathcal{U}}
\newcommand{\V}{\mathcal{V}}

\newcommand{\dm}{\bm{d}}

\newcommand{\wm}{\bm{\theta}}
\newcommand{\xm}{\bm{x}}
\newcommand{\ym}{\bm{y}}

\newcommand{\zm}{\bm{z}}

\newcommand{\Am}{\bm{A}}

\newcommand{\Hm}{\bm{H}}
\newcommand{\Imm}{\bm{I}}

\newcommand{\Fm}{\bm{F}}

\newcommand{\Qmi}[1]{\bm{Q}_{#1}}
\newcommand{\Qgi}[1]{\widehat{\bm{Q}}_{#1}}

\newcommand{\X}{\bm{\mathcal{X}}}

\newcommand{\EE}{\mathbb{E}}
\newcommand{\EEi}[1]{\mathbb{E}\left[#1\right]}

\newcommand{\PPi}[1]{\mathbb{P}\left(#1\right)}

\newcommand{\setall}{
\makeatletter
\renewcommand\normalsize{%
	\@setfontsize\normalsize\@xpt\@xiipt
	\abovedisplayskip 7\p@ \@plus1\p@ \@minus6\p@
	\abovedisplayshortskip \z@ \@plus3\p@
	\belowdisplayshortskip 3\p@ \@plus3\p@ \@minus3\p@
	\belowdisplayskip \abovedisplayskip
	\let\@listi\@listI}
\makeatother
}

\title{Towards Theoretically Understanding Why \Sgd\ Generalizes Better Than \Adam~in Deep Learning}

\author{ Pan Zhou$^{*}$,  \ \ \normalsize{Jiashi~Feng$^{\dagger}$,}  \ \  \normalsize{Chao Ma$^{\ddagger}$}, \ \    \normalsize{Caiming Xiong$^{*}$}, \ \     \normalsize{Steven HOI$^{*}$},\ \     \normalsize{Weinan E$^{\ddagger}$}\\
	{$^{*}$Salesforce Research,\ \ $^{\dagger}$ National University of Singapore,\ \ $^{\ddagger}$ Princeton University} \\
	{\small{\texttt\{pzhou,shoi,cxiong\}@salesforce.com}  \  {\texttt\tt elefjia@nus.edu.sg}   \   {\texttt\tt \{chaom@, weinan@math.\}princeton.edu} }
}

\begin{document}

\maketitle

\begin{abstract}
	It is not clear yet why \Adam-alike  adaptive gradient algorithms suffer from worse generalization performance than  \Sgd~despite their faster training speed.   This work aims to provide understandings on this generalization gap by  analyzing 	their local convergence behaviors. Specifically,   we  observe the heavy tails of gradient noise in these algorithms. This motivates us to  analyze these algorithms through their  \levyp-driven stochastic differential equations (SDEs)  because of the similar convergence behaviors of an algorithm and its SDE. Then we  establish the escaping time of these SDEs  from a local basin. The result shows that (1) the escaping time of both \Sgd~and \Adam~depends on the  Radon measure of the basin positively and the heaviness of gradient noise negatively;  (2) for the same basin, \Sgd~enjoys  smaller escaping time than \Adam, mainly because  (a) the geometry adaptation in \Adam~via adaptively scaling each gradient coordinate well diminishes the   anisotropic structure in gradient noise  and results in larger Radon measure of a basin; (b)  the  exponential gradient average in \Adam~smooths its gradient  and  leads to  lighter gradient noise tails than \Sgd.  So \Sgd~is more locally unstable than \Adam~at sharp minima  defined as the minima whose local basins have small Radon measure, and  can better escape from them to flatter ones with larger Radon measure.   As  flat minima here which often refer to the minima  at  flat or asymmetric basins/valleys  often generalize better  than sharp ones~\cite{keskar2016large,he2019asymmetric}, our result  explains the better generalization performance of \Sgd~over \Adam.    Finally,  experimental results confirm our heavy-tailed gradient noise assumption and theoretical affirmation.
\end{abstract}

\section{Introduction}\label{introduction}

Stochastic gradient descent (\Sgd)~\cite{robbins1951stochastic,bottou1991stochastic} has become one of the most popular algorithms for training deep neural networks~\cite{bengio2009learning,hinton2012deep,lecun2015deep,zhou2018DSCN,he2016deep,zhou2019metalearning,zhou2020NAS}. In spite of its simplicity and effectiveness,   \Sgd~uses one learning rate for all gradient coordinates    and could suffer from unsatisfactory convergence performance, especially for ill-conditioned problems~\cite{keskar2017improving}.  To avoid this issue, a variety of adaptive  gradient algorithms have been developed that adjust learning rate  for each gradient coordinate  according to the current     geometry curvature of the objective function \cite{duchi2011adaptive,hiton2012,kingma2014adam,reddi2019convergence}. These   algorithms, especially for \Adam,  have achieved  much faster convergence speed than vanilla \Sgd~in practice.  

Despite their faster convergence behaviors, these adaptive gradient  algorithms usually suffer from worse generalization performance than  \Sgd~\cite{wilson2017marginal,keskar2017improving,luo2019adaptive}. Specifically, adaptive gradient algorithms often show faster   progress in the training phase but their performance quickly reaches a plateaus on test data. Differently,  \Sgd~usually improves model performance slowly but could achieve higher test performance.  
One   empirical  explanation~\cite{keskar2016large,hochreiter1997flat,izmailov2018averaging,li2018visualizing} for this generalization  gap is that adaptive gradient algorithms 
tend to converge to sharp minima whose local basin has large curvature 	and usually generalize poorly, while \Sgd~prefers to find flat minima  
and thus  generalizes better.  
However,  recent evidence~\cite{he2019asymmetric,sagun2017empirical} shows that (1) for  deep neural networks, the minima  at the  asymmetric basins/valleys   where  both steep  and flat directions exist   also generalize well though they are sharp  in terms of  their  local curvature, and (2) \Sgd~often converges to these minima.  So the  argument of the  conventional ``flat" and ``sharp" minima defined on curvature   cannot  explain these new results. Thus  the reason for   the generalization gap between adaptive gradient methods and \Sgd~is  still  unclear. 

\begin{figure}[tb]
	\begin{center}
		\setlength{\tabcolsep}{0.8pt}  
		\begin{tabular}{cc}
			\includegraphics[width=0.49\linewidth]{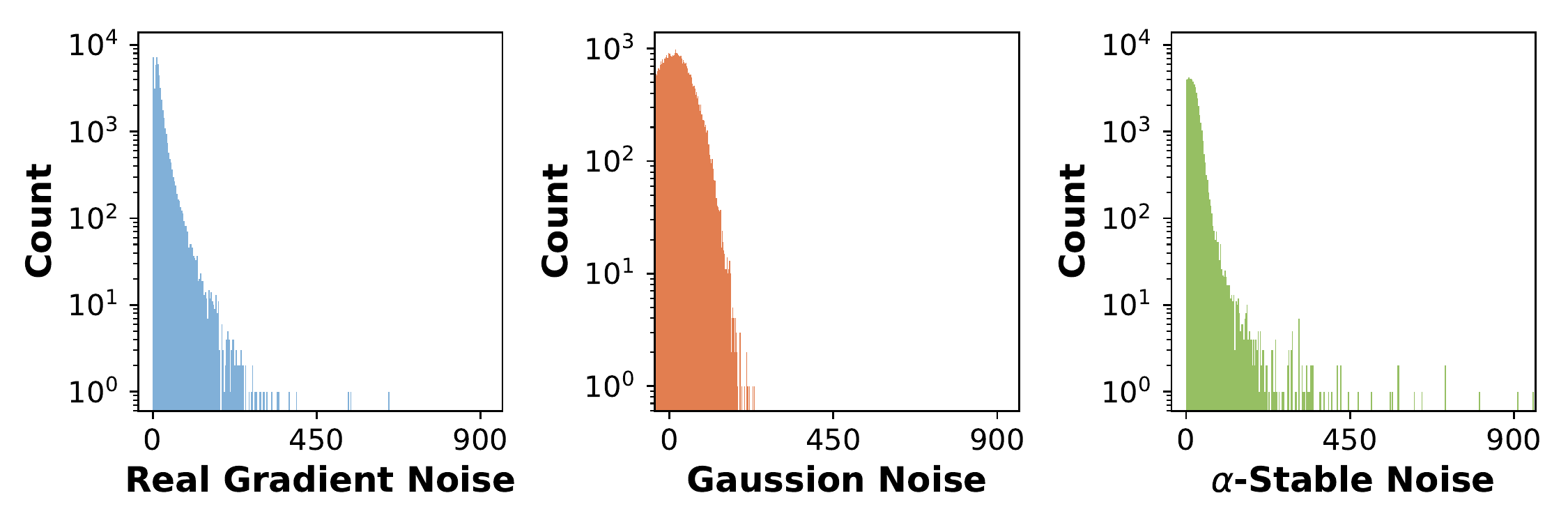} & \includegraphics[width=0.49\linewidth]{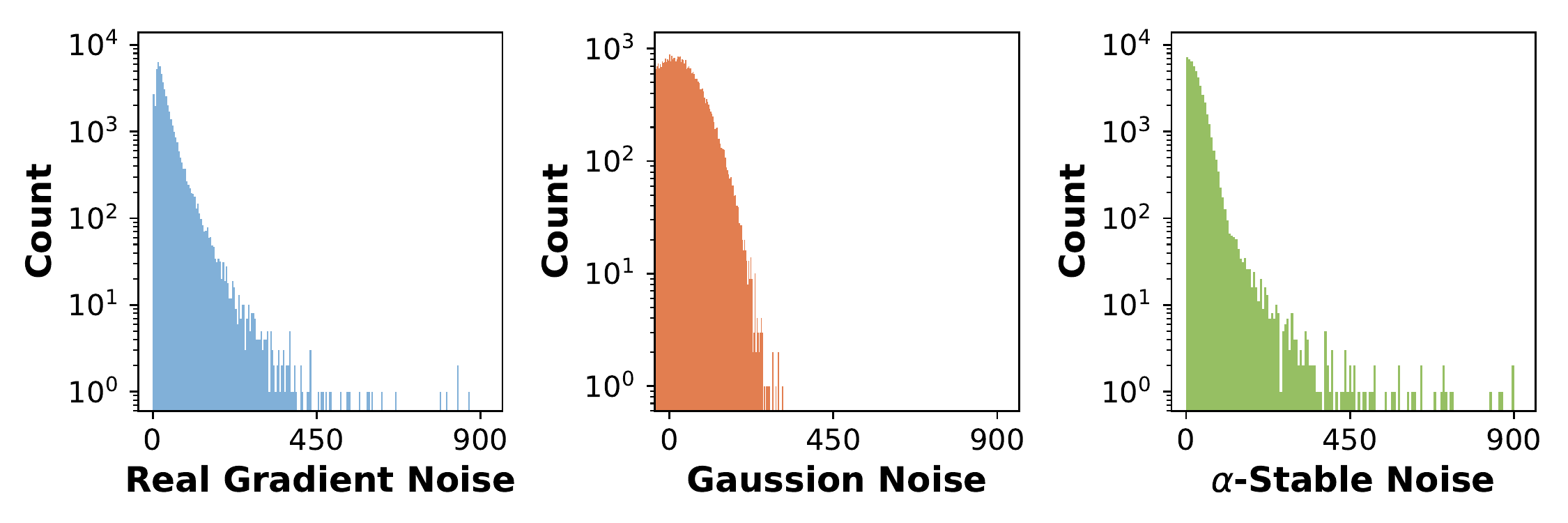} \vspace{-0.1em}\\
			{\small{(a) \Adam}}  & 	{\small{(b) \Sgd}} \\
		\end{tabular}
	\end{center}
	\vspace{-0.4em}
	\caption{Illustration of gradient noise in \Adam~and \Sgd~on AlexNet trained with CIFAR10. (b) is produced under the same setting in~\cite{simsekli2019tail}.  
		By comparison, one can observe (1) $\alpha$-stable noise  can better characterize  real gradient noise and (2) \Sgd~has heavier gradient noise tails than \Adam. } 
	\label{illustration_grdientnoise}
	\vspace{-0.6em}
\end{figure}

In this work, we provide a new viewpoint for understanding the generalization performance gap. We first formulate \Adam~and \Sgd~as  \levyp-driven  stochastic differential equations (SDEs), since  the SDE of an algorithm shares similar convergence behaviors of  the algorithm and can be analyzed more easily than directly analyzing the algorithm.  Then we  analyze  the escaping behaviors of these SDEs  at local minima  to investigate the generalization gap between  \Adam~and \Sgd, as escaping behaviors determine which basin that an algorithm finally converges to and thus affect the generalization performance of the algorithm.  By analysis, we find that compared with  \Adam, \Sgd~is more locally unstable  and is more likely to converge to the minima at the flat or asymmetric basins/valleys which often  have better generalization performance over other type minima. So our results can explain the better generalization performance of \Sgd~over \Adam. Our contributions are highlighted below.

Firstly, this work is  the first one that adopts \levyp-driven SDE which better characterizes the algorithm gradient noise in practice, to analyze the adaptive gradient algorithms. Specifically, Fig.~\ref{illustration_grdientnoise}   shows that the gradient noise  in \Adam~and \Sgd, \ie~the difference between the full  and stochastic gradients, has heavy tails   and can be well characterized by systemic $\alpha$-stable  ($\Sas$) distribution~\cite{levy1954theorie}.  
Based on this observation, we  view \Adam~and \Sgd~as discretization of the continuous-time processes   and  formulate the processes as \levyp-driven  SDEs to analyze their behaviors.  
Compared with Gaussian gradient noise assumption in \Sgd~\cite{mandt2016variational,jastrzkebski2017three,chaudhari2018stochastic}, $\Sas$ distribution assumption can  characterize the heavy-tailed gradient noise  in practice more accurately as shown in Fig.~\ref{illustration_grdientnoise}, and also better explains the different generalization performance of \Sgd~and \Adam~as discussed in Sec.~\ref{levySDEs}.   
This work  extends \cite{simsekli2019tail,Pavlyukevich2007levy} from \Sgd~on the over-simplified one-dimensional  problems to much more complicated adaptive   algorithms on high-dimensional problems.  It also differs from~\cite{pavlyukevich2011first}, as \cite{pavlyukevich2011first} considers   escaping behaviors of \Sgd~along several fixed directions, while this work analyzes the dynamic underlying structures in gradient noise that plays an important role in the local escaping behaviors of both \Adam~and~\Sgd.

Next, we theoretically prove that for the \levyp-driven SDEs of  \Adam~and \Sgd, their escaping time $\te$ from a local basin $\Omegas$, namely the least  time for escaping from the inner of $\Omegas$ to  its outside, is at the order of $\mathcal{O}(\epsi{-\alpha} \!/m(\W) )$, where  the constant $\epsi{} \!\in\!(0,1)$ relies on the learning rate of algorithms and  $\alpha$ denotes the tail index of $\Sas$ distribution. Here   $m(\W)$ is a non-zero Radon measure on the escaping set  $\W$ of \Adam~ and \Sgd~ at the local basin $\Omegas$ (see Sec.~\ref{flatone}), and actually negatively relies on the  Radon measure of $\Omegas$. So   both  \Adam~and \Sgd~have small escaping time at the ``sharp" minima  whose corresponding basins $\Omegas$ have small Radon measure. It means that \Adam~and \Sgd~are actually unstable  at ``sharp" minima and would escape them to ``flatter" ones.  Note, the Radon measure of $\Omegas$   positively depends on the volume of $\Omegas$. So these results also well explain the observations  in~\cite{keskar2016large,he2019asymmetric,izmailov2018averaging,li2018visualizing} that the minima of   deep networks  found by \Sgd~often locate at the flat or asymmetric valleys, as their corresponding basins have large volumes and thus large Radon measure. 

Finally, our results can   answer why \Sgd~often converges to flatter minima than \Adam~in terms of Radon measure,  and thus  explain the generalization gap between \Adam~and \Sgd.  Firstly, our analysis shows that even for the same basin $\Omegas$,  \Adam~often has smaller Radon measure  $m(\W)$ on the escaping set $\W$ at   $\Omegas$ than \Sgd, as the geometry adaptation in \Adam~via adaptively scaling each gradient coordinate well diminishes   underlying  anisotropic structure in gradient noise  and  leads to smaller $m(\W)$. 
Secondly, the empirical results in Sec.~\ref{experiments} and Fig.~\ref{illustration_grdientnoise} show that  \Sgd~often 
has much smaller tail index $\alpha$ of gradient noise than \Adam~for some optimization iterations and thus enjoys smaller factor $\epsi{-\alpha}$.  
These results together show that   \Sgd~is more locally unstable and would  like to converge to flatter minima with larger measure $m(\W)$ which often refer to the minima  at the flat and asymmetric basins/valleys, according with empirical evidences in~\cite{keskar2017improving,merity2017regularizing,loshchilov2016sgdr,wilson2017marginal}. Considering the  observations in~\cite{keskar2016large,hochreiter1997flat,izmailov2018averaging,li2018visualizing} that the minima at the flat and asymmetric basins/valleys often generalize better, our results  well explain the generalization gap between \Adam~and \Sgd.  Besides, our results also show  that \Sgd~benefits from its anisotropic gradient noise on its escaping behaviors, while \Adam~does not.

\section{Related Work}
 
Adaptive gradient algorithms  have become the default optimization tools in deep learning because of their fast convergence speed.  
But they often suffer from worse generalization performance than \Sgd~\cite{keskar2017improving,wilson2017marginal,merity2017regularizing,loshchilov2016sgdr}.  Subsequently, most works~\cite{keskar2017improving,merity2017regularizing,loshchilov2016sgdr,wilson2017marginal,luo2019adaptive} empirically analyze this issue from  the argument of flat and sharp minima defined on local curvature  in~\cite{hochreiter1997flat} that flat minima often generalize better than sharp ones, as  they observed that \Sgd~often converges  to flatter minima than adaptive gradient  algorithms,~\eg~\Adam. 
However, Sagun~\et~\cite{sagun2017empirical}   and He~\et~\cite{he2019asymmetric} observed that the minima of modern deep networks  at the asymmetric valleys where  both steep  and flat directions exist also  generalize well, and \Sgd~often converges to these minima.  So the conventional  flat and sharp argument cannot  explain these new results. This work theoretically shows that \Sgd~tends to converge to the minima whose local basin has larger Radon measure. It  well explains the above  new observations, as the minima with larger Radon measure often locate  at the flat and asymmetric basins/valleys. Moreover, based on our results, exploring  invariant Radon measure to  parameter scaling in networks could resolve the issue in  \cite{dinh2017sharp} that flat minima could become sharp via parameter scaling. See more details in   Appendix~\ref{comparison}.  Note,   \Adam~could achieve better performance than \Sgd~when gradient clipping is required~\cite{zhang2019adam}, \eg~attention models with gradient exploding issue, as adaptation in \Adam~provides a clipping effect. This work considers  a general non-gradient-exploding setting, as it is more practical across many important tasks, \eg~classification.

For theoretical generalization analysis, most works~\cite{mandt2016variational,chaudhari2018stochastic,jastrzkebski2017three,zhu2018anisotropic} only focus on analyzing \Sgd. They  formulated \Sgd~into Brownian motion based SDE via assuming  gradient noise to be Gaussian. 
For instance,  
Jastrzkebski~\et~\cite{jastrzkebski2017three} proved that the larger ratio of learning rate to mini-batch size in \Sgd~leads to flatter minima and  better generalization.   
But  Simsekli~\et~\cite{simsekli2019tail} empirically found that the gradient noise has heavy tails and can be  characterized by $\Sas$ distribution instead of Gaussian distribution.   Chaudhari~\et~\cite{chaudhari2018stochastic} also  claimed that the trajectories of \Sgd~in deep networks are not  Brownian motion. Then Simsekli~\et~\cite{simsekli2019tail}~formulated \Sgd~as a \levyp-driven SDE~and adopted the results in \cite{Pavlyukevich2007levy} to show that \Sgd~tends to converge to flat  minima on one dimensional problems.  Pavlyukevich~\et~\cite{pavlyukevich2011first} extended  the one-dimensional SDE in~\cite{Pavlyukevich2007levy} and analyzed   escaping behaviors of \Sgd~along several fixed directions, differing from this work that analyzes  dynamic underlying structures in gradient noise that greatly affect  escaping behaviors of both \Adam~and~\Sgd.
 
The literature targeting theoretically understanding the generalization degeneration  of adaptive gradient algorithms are limited mainly due to their more complex algorithms. Wilson~\et~\cite{wilson2017marginal} constructed a binary classification problem  and 
showed that \AdaGrad~\cite{duchi2011adaptive} tend to give undue influence to spurious features that have no effect on out-of-sample generalization.  Unlike the above theoretical works that focus on analyzing \Sgd~only or special problems, we target at revealing the different convergence behaviors of   adaptive gradient algorithms and \Sgd~and also analyzing their different generalization performance, which is of more practical interest especially in deep learning.

\section{\levyp-driven SDEs of  Algorithms in Deep Learning}\label{levySDEs}
 
In this section, we  first briefly introduce \Sgd~and \Adam, and formulate them as discretization of stochastic differential equations (SDEs) which is a popular approach to analyze algorithm behaviors.  
Suppose the  objective function of $n$ components in deep learning models is formulated as 
\begin{equation}\label{DLproblem} 
\min\nolimits_{\wm\in\Rs{d}}\ \Fm(\wm) :={\frac{1}{n}\sum_{i=1}^n f_i(\wm),}
\end{equation}
where $f_i(\wm)$ is the loss of the $i$-th sample.  Subsequently, we  focus on analyzing \Sgd\ and \Adam. Note our analysis technique is  applicable to other adaptive  algorithms with similar results as \Adam.  

\vspace{-0.5em}
\subsection{\Sgd~and \Adam}\label{Preliminaries}
\vspace{-0.4em} 

As one of the most effective algorithms, \Sgd~\cite{robbins1951stochastic} solves problem~\eqref{DLproblem} by sampling a  data mini-batch $\Bi{t}$ of size $\batchsize$   and then running one  gradient descent step:
\begin{equation}\label{SGD} 
\wmi{t+1}=\wmi{t}-\etai{} \nabla \Fbi{t}, 
\end{equation}
where  $\nabla \Fbi{t} \!=\!\frac{1}{\batchsize}{\sum_{i\in\Bi{t}}}\!\nabla f_i(\wmi{t})$ denotes the gradient on mini-batch $\Bi{t}$, and $\etai{}$ is the learning rate.  
Recently, to  improve the efficiency of \Sgd, adaptive gradient algorithms, such as~\AdaGrad~\cite{duchi2011adaptive}, \RMSProp~\cite{hiton2012} and \Adam~\cite{kingma2014adam}, are developed which  adjust the learning rate of each gradient coordinate according to the current geometric curvature.  
Among them, \Adam~has become the default\\ training algorithm in deep learning. 
Specifically,  \Adam~estimates the current gradient $\nabla \Fm(\wmi{t})$ as 
\begin{equation*} 
\mmi{t}=\betai{1} \mmi{t-1} + (1-\betai{1})\nabla\Fbi{t} \ \quad \text{with}\ \mmi{0}=\bm{0} \ \ \text{and}\ \  \betai{1}\!\in\!(0,1).
\end{equation*} 
Then like natural gradient descent~\cite{amari1998natural}, \Adam~adapts itself to the function geometry via  a  diagonal
Fisher  matrix approximation $\diag{\vmi{t}}$  which serves as a preconditioner and is defined as 
\begin{equation*} 
\vmi{t}=\betai{2} \vmi{t-1} + (1-\betai{2}) [\nabla \Fbi{t}]^2\ \quad \text{with}\  \vmi{0}=\bm{0} \ \ \text{and}\ \  \betai{2}\in (0,1).
\end{equation*} 
Next \Adam~preconditions the problem by scaling each gradient coordinate, and updates the variable    
\begin{equation}\label{ADSGD} 
\wmi{t+1}=\wmi{t}-\etai{}  \mmi{t}/(1-\betai{1}^t)  / \Big({\sqrt{\vmi{t}/(1-\betai{2}^t)}}+\epsilon\Big)\ \quad \ \text{with a small  constant}\ \epsilon.
\end{equation}  

\subsection{\levyp-driven SDEs}\label{leverydirvensde} 
Let $\umi{t}\!=\!\nabla \Fm(\wmi{t}) \!-\! \nabla\! \Fbi{t}$ denote gradient noise.  From Sec.~\ref{Preliminaries}, we can formulate \Sgd~as follows
\begin{equation*}  
\wmi{t+1}\!=\!\wmi{t}-\etai{}  \nabla \Fm(\wmi{t}) + \etai{}  \umi{t}. 
\end{equation*}  
To analyze  behaviors of an algorithm, one effective approach is  to obtain its SDE  via making assumptions on $\umi{t}$ and then analyze its SDE. For instance, to  analyze \Sgd, most works~\cite{mandt2016variational,chaudhari2018stochastic, jastrzkebski2017three,zhu2018anisotropic} assume that $\umi{t}$ obeys a Gaussian distribution $\mathcal{N}(\bm{0},\! \Sigmai{t})$ with covariance matrix
\begin{equation*}  
\Sigmai{t}={ \frac{1}{\batchsize}\Big[\frac{1}{n} \sum\nolimits_{i=1}^n \nabla f_i(\wmi{t}) \nabla f_i(\wmi{t})^T - \nabla \Fm(\wmi{t}) \nabla \Fm(\wmi{t})^T\Big].}
\end{equation*} 
However, both recent work~\cite{simsekli2019tail} and Fig.~\ref{illustration_grdientnoise} show that the gradient noise $\umi{t}$ has  heavy tails and can be better characterized by $\Sas$ distribution~\cite{levy1954theorie}. Moreover, the heavy-tail assumption   can also better explain the  behaviors of   \Sgd~than   Gaussian noise assumption. Concretely, for the SDE of \Sgd~on the one-dimensional  problems,  under Gaussian noise assumption its escaping time from  a simple quadratic basin  respectively exponentially and polynomially depends on the height and width of the basin~\cite{imkeller2010hierarchy},  indicating that \Sgd~gets stuck at deeper minima as opposed to wider/flatter minima. This contradicts with the  observations  in~\cite{keskar2016large,hochreiter1997flat,izmailov2018averaging,li2018visualizing} that \Sgd~often converges to flat minima. By  contrast, on the same problem, for \levyp-driven SDE, both \cite{simsekli2019tail} and this work show that \Sgd~tends to converge to flat minima instead of deep minima, well explaining the convergence behaviors of \Sgd.  

Following~\cite{simsekli2019tail}, we also assume $\umi{t}$ obeys $\Sas$ distribution but with a time-dependent covariance matrix $\Sigmai{t}$ to better characterize the  underlying structure in the gradient noise $\umi{t}$.  In this way, when the learning rate $\eta$ is small and $\varepi{}=\eta^{(\alpha-1)/\alpha}$, we can write the \levyp-driven SDE of \Sgd~as 
\begin{equation}\label{assumption_stochastic_RMSP} 
\ds \wmi{t} =  -  \nabla \Fm(\wmi{t}) + \varepi{} \Sigmai{t}\ds \levyi{t}.
\end{equation}
Here the \levyp~motion  $\levyi{t}\in\Rs{d}$ is a random vector and  its $i$-th entry $\levyi{t,i}$ obeys the $\Sas(1)$ distribution which is defined through the characteristic function $\EE[\exp(i\omega x)]=\exp(-\sigma^{\alpha} |\omega|^{\alpha})$ if $x\sim\Sas(\sigma)$. Intuitively, the $\Sas$ distribution is a heavy-tailed distribution with a decay density like $1/|x|^{1+\alpha}$.  When the tail index  $\alpha$ is 2,  $\Sas(1)$ becomes a Gaussian distribution and thus has stronger data-fitting capacity over Gaussian distribution.   In this sense, the SDE of \Sgd~in~\cite{mandt2016variational,chaudhari2018stochastic,jastrzkebski2017three,li2017stochastic,zhu2018anisotropic}   is actually a special case of the \levyp-driven SDE in this work.  Moreover, Eqn.~\eqref{assumption_stochastic_RMSP} extends the one-dimensional SDE of \Sgd~in~\cite{simsekli2019tail}.  Note, Eqn.~\eqref{assumption_stochastic_RMSP}  differs from \cite{pavlyukevich2011first}, since it considers dynamic covariance matrix $\Sigmai{t}$ in gradient noise and shows great effects of its underlying structure to the escaping behaviors in both \Adam~and~\Sgd, while \cite{pavlyukevich2011first}   analyzed  escaping behaviors of \Sgd~along several fixed directions.

Similarly, we can derive the SDE of \Adam.  For brevity, we define $ \mmi{t}' \!= \! \betag \mmi{t-1}' \!+\!(1- \betag) \nabla\Fm(\wmi{t})$ with $\mmi{0}'=\bm{0}$. Then by the definitions of  $ \mmi{t}$ and $\mmi{t}'$, we can compute 
\begin{equation*}\label{mmt} 
\mmi{t}'- \mmi{t} ={ (1-\betag) \sum_{i=0}^{t} \betag^{t-i} [\nabla\Fm(\wmi{i}) - \nabla\Fbi{i}]}={ (1-\betag) \sum_{i=0}^{t} \betag^{t-i}\umi{i}}.
\end{equation*}
As noise $\umi{t}$  has  heavy tails,  their exponential average  should  have similar behaviors, which is also illustrated by Fig.~\ref{illustration_grdientnoise}.  So we also assume ${\frac{1}{1-\beta_{{\scriptscriptstyle{1}}}^t}}(\mmi{t}'- \mmi{t})$ obeys $\Sas(1)$ distribution with covariance matrix $\Sigmai{t}$.   
Meanwhile,  we can write \Adam~as 
\begin{equation*}  
\wmi{t+1}=\wmi{t}-{\etai{}   \mmi{t}'/\zm_t+ \etai{}   (\mmi{t}'-\mmi{t})/\zm_t} \quad \text{with} \quad \zm_t\!=\!(1\!-\!\betag^t)  \Big({\sqrt{\vmi{t}/(1-\betai{2}^t)}} + \epsilon\Big).
\end{equation*}  
So we can derive the \levyp-driven SDE of \Adam:
\begin{equation}\label{assumption_stochastic_adam} 
\ds \wmi{t} =  -\mui{t} \Qmi{t}^{-1}\mmi{t} \!+\! \varepi{}\Qmi{t}^{-1} \Sigmai{t}\ds \levyi{t}, \ \ 
\ds \mmi{t} =   \betag(\nabla\Fm(\wmi{t})  \!-\! \mmi{t}),\ \ 
\ds \vmi{t} =  \betav([\nabla \Fbi{t}]^2\!-\!\vmi{t}), 
\end{equation} 
where $\varepi{}\!=\!\eta^{(\alpha-1)/\alpha}$,   $\Qmi{t}\!=\!\diag{\sqrt{\omegai{t}\vmi{t}} +\thres}$, $\mui{t}\!=\!1/(1-e^{-\betag t})$ and $\omegai{t}=1/(1-e^{-\betav t})$  are two constants to correct the bias in $\mmi{t}$ and $\vmi{t}$. Note, here we replace $\mmi{t}'$ with $\mmi{t}$ for brevity. Appendix~\ref{constructiondetails} provides more construction details, randomness discussion and shows the fitting capacity of this SDE to \Adam. Subsequently, we will   analyze escaping behaviors of the SDEs in Eqns.~\eqref{assumption_stochastic_RMSP} and~\eqref{assumption_stochastic_adam}.

\section{Analysis for Escaping Local Minima}\label{volumeanalysis} 
Now we analyze the local stability of \Adam-alike adaptive  algorithms and \Sgd. Suppose the process $\wmi{t}$  in Eqns.~\eqref{assumption_stochastic_RMSP} and~\eqref{assumption_stochastic_adam}  starts at  a local basin $\Omegas$ with a minimum $\wms$, \ie~$\wmi{0}\!\in\!\Omegas$.  Here  we are particularly interested in the first escaping time $\te$ of $\wmi{t}$ produced by an algorithm  which reveals  the convergence behaviors  and    generalization performance of the algorithm. Formally, let $\Grd\!\!=$ $\{\ym \!\in\! \G \ |\ \dis(\partial \G, \ym)\! \geq\! \epsi{\gamma}\}$  denote the inner part of $\G$. Then we give two important definitions,~\ie~(1) the escaping time $\te$ of the process $\wmi{t}$ from the local basin $\G$ and (2) the escaping set $\W$ at  $\Omegas$, as 
\begin{equation}\label{escapingset} 
\te=\inf\{t\geq 0\ |\ \wmi{t}\notin \Grd\}\quad \text{and}\quad \W=\{\ym\in\Rs{d}\ | \ \Qmi{\wms}^{-1} \Sigmai{\wms}\ym \notin\Grd\},
\end{equation}
where the constant  $\gamma>0$ satisfies $\lim_{\epsi{}\rightarrow 0} \epsi{\gamma}=0$,   $\Sigmai{\wms}$ $=\!\lim_{\wmi{t}\rightarrow \wms}\! \Sigmai{t}$ for both \Sgd~and \Adam, and $\Qmi{\wms}\!\!=\Imm$ for \Sgd~and $\Qmi{\wms}\!\!=\!\lim_{\wmi{t}\rightarrow \wms}\! \Qmi{t}$ for \Adam.  Then we define  Radon measure~\cite{simon1983lectures}.
\begin{defn}\label{radonmeasure}
	If a measure $m(\V)$ defined on Hausdorff topological space $\X$ obeys (1) inner regular, \ie~$m(\V) \!=\! \sup_{\U\subseteq\V}\! m(\U)$,  (2) outer regular, \ie~$m(\V) \!=\! \inf_{\V\subseteq\U} \!m(\U)$, and (3)  local finiteness, \ie~every point of $\X$ has a neighborhood $\U$ with finite $m(\U)$, then  $m(\V)$ is a Radon measure. \vspace{-0.5em}
\end{defn}
Then we  define non-zero Radon measure which further  obeys $m(\U)\!<\! m(\V)$ if $\U\!\subset\! \V$. Since larger set has larger volume,     $m(\U)$  positively depends on the volume of the set $\U$. Let $m(\W)$ be a non-zero Radon measure on the set $\W$. Then 
we first introduce two mild assumptions for analysis. 
 
\begin{assum}\label{function_assumption}
	For both \Adam~and \Sgd, suppose the objective $\Fm(\wm)$ is a  upper-bounded non-negative loss, and is  locally $\mu$-strongly convex and $\ells$-smooth in the  basin $\Omegas$.  
\end{assum}
\begin{assum}\label{algorithm_assumption}
	For \Adam, suppose its  process $\wmi{t}$ satisfies ${\int_{0}^{\te} \big\langle \frac{\nabla \Fm(\wmi{s})}{1+ \Fm(\wmi{s})}, } \mui{s} {\Qmi{s}^{-1} } \mmi{s}   \big\rangle\ds s\geq 0$ almost sure, and its parameters $\betag$ and $\betav$ obey $\betag\leq \betav\leq 2\betag$. Moreover, for \Adam, we assume $\|\mmi{t}-\mgi{t}\|\leq \taum \|{\int_{0}^{t-}}(\mmi{s}-\mgi{s})\ds s\|$ and $\|\mgi{t}\|\geq \tau \|\nabla \Fm(\wgi{t})\|$ where $\mgi{t}$ and $\wgi{t}$ are obtained by Eqn.~\eqref{assumption_stochastic_adam} with $\epsi{}=0$. Each coordinate $\vmi{t,i}$ of $\vmi{t}$ in \Adam~obeys $\vmin \leq  \sqrt{\vmi{t,i}}\leq \vmax$ $(\forall i, t)$.  
\end{assum} 

\begin{wrapfigure}{r}{0.4\linewidth} 
	\vspace{-1.9em}
	\begin{center}
		\setlength{\tabcolsep}{0.1pt}  
		\begin{tabular}{c}
			\includegraphics[width=0.9\linewidth]{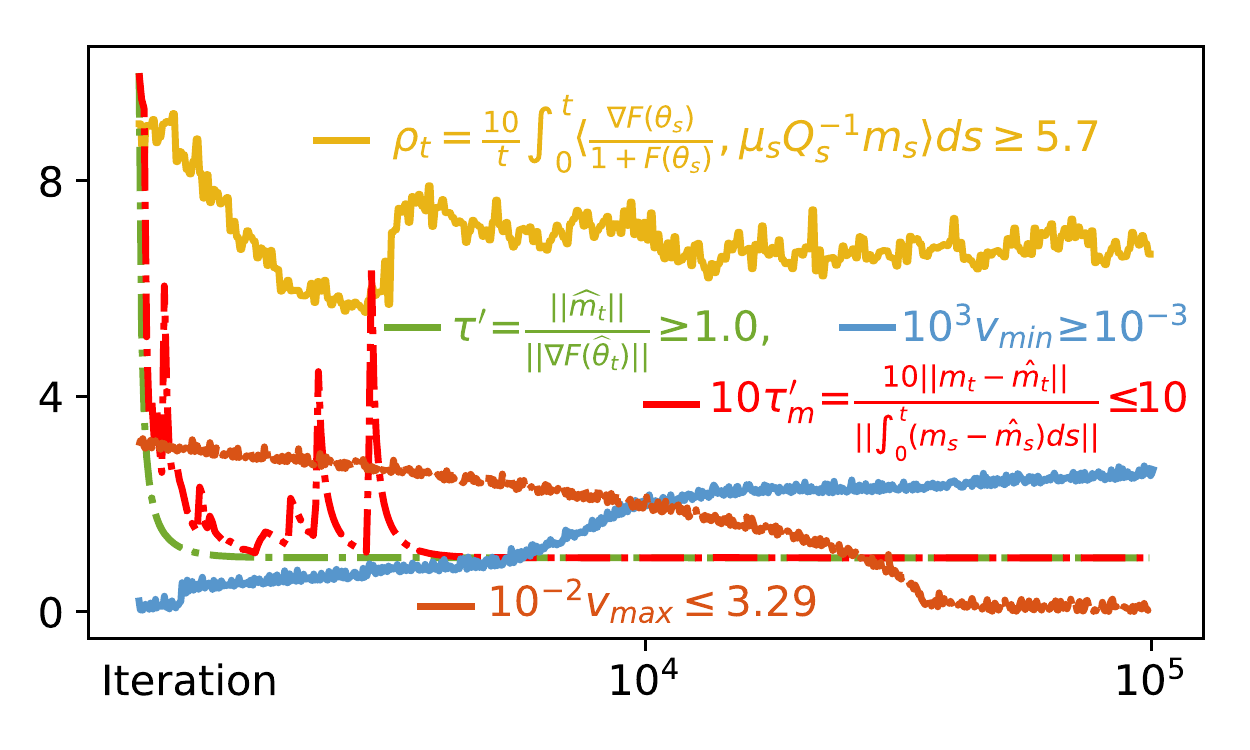}
		\end{tabular}
	\end{center}
	\vspace{-0.8em}
	\caption{Empirical investigation of Assumption~\ref{algorithm_assumption} on \Adam.   
	}
	\label{assumption_verification}
	\vspace{-1.1em}
\end{wrapfigure}
Assumption~\ref{function_assumption} is very standard for analyzing  stochastic   optimization~\cite{ghadimi2013stochastic,zhou2018hardthresholding,SVRG,zhou2018HSGDHT,zhou2019Riemannian,zhou2020hybird} and network analysis~\cite{du2018gradient,tian2017analytical,zhou2018analysiscnns,zhou2018analysisdnn}. In Assumption~\ref{algorithm_assumption}, we indeed require similar directions of  gradient estimate $\mmi{t}$ and  full gradient $\nabla \Fm(\wmi{t})$ in  \Adam~in most cases, as we  assume their inner product is non-negative along the iteration trajectory. So this assumption  can be satisfied in practice.  To analyze  the processes $\wmi{t}$ and $\wgi{t}$ in \Adam, we make an assumption on the distance between their corresponding gradient estimates $\mmi{t}$ and $\mgi{t}$ which can be easily fulfilled by their definitions. Then for \Adam, we mildly assume  its estimated  $\vmi{t}$ to be bounded. For $\vmin$, we indeed allow $\vmin=0$ because of the  small  constant $\epsilon$. The relation $\betag\leq \betav\leq 2\betag$ is also satisfied under the default setting of \Adam.  Actually, we also empirically investigate Assumption~\ref{algorithm_assumption} on \Adam. In Fig.~\ref{assumption_verification}, we report the values of $\rho_{t}=\frac{10}{t}\int_{0}^{\te} \!\!\big\langle \frac{\nabla \Fm(\wmi{s})}{1+ \Fm(\wmi{s})}, \mui{s}{\Qmi{s}^{-1} } \mmi{s}   \big\rangle d s$, $\taum'=\frac{\|\mmi{t}-\mgi{t}\|}{\|{\int_{0}^{t-}}\!\!(\mmi{s}-\mgi{s}) \ds s\|}$, $\tau'=\frac{\|\mgi{t}\|}{ \|\!\nabla\! \Fm(\wgi{t})\|}$, $\vmin= \min_{i} \sqrt{\vmi{t,i}},  \vmax= \max_{i} \sqrt{\vmi{t,i}}$ in the SDE of \Adam~on the 4-layered fully connected networks with width 20.  Note that we scale some values of $\rho_{t}$, $\taum'$, $ \tau'$, $ \vmin$ and $ \vmax$ so that we can plot them in one figures. From~Fig.~\ref{assumption_verification}, one can observe that $\rho_{t}, \tau'$ and $ \vmin$ are well lower bounded, and $\taum'$ and $ \vmax$ are well upper bounded.  These results demonstrate the  validity  of Assumption~\ref{algorithm_assumption}.

With these two assumptions, we analyze the escaping time $\te$ of process $\wmi{t}$  and summarize the main results in Theorem~\ref{upperlowerbound}. For brevity, we define a group of key constants for \Sgd~and \Adam: $\clip\!=\!\ell$ and $\kappa_2\!=\!2\mu$  in \Sgd,  $\clip\!=\!\frac{c_1 \ells}{(\vmin+\thres) |\taum-1|} $  and $\kappa_2\!=\!\frac{2\mu\tau}{ \betag \left(\vmax+\thres\right) +\mu\tau} (\betag \!-\!\frac{\betav}{4} )$  in \Adam~with a constant $c_1$.
\begin{thm}\label{upperlowerbound}
	Suppose Assumptions~\ref{function_assumption} and~\ref{algorithm_assumption} hold. Let $\Hui{\epsi{-1}}\!=\!\frac{2}{\alpha} \epsi{\alpha}$, $\rho_0\!=\!\frac{1}{16(1+c_2 \clip)}$ and $\ln\!\big(\frac{2\Delta}{\mu \epsi{1\!/3}}\big) \! \leq\! \kappa_2 \epsi{-\frac{1}{3}}$ with $\Delta \!=\!\Fm(\wmi{0}) $ $-\Fm(\wms)$ and a constant $c_2$. Then for any  $\wmi{0}\!\in\GG$, $u>\!-1$, $\epsi{}\in(0,\epsi{}_0]$, $\gamma\in (0,\gamma_0]$ and $\rho\in(0,\rho_0] $ satisfying $\epsi{\gamma}\leq \rho_0$ and $\lim_{\epsi{}\rightarrow 0}\rho = 0$, \Sgd~in~\eqref{assumption_stochastic_RMSP} and \Adam~in~\eqref{assumption_stochastic_adam} obey 
	\begin{equation*} 
	{ \frac{1-\rho }{1+u+\rho} }\leq	\EEi{\exp{\left(-um(\W) \Hui{\epsi{-1}}\te \right)}} \leq  {\frac{1+\rho }{1+u-\rho}}.
	\end{equation*}
\end{thm} 
See its proof in Appendix~\ref{proofofupperlowerbound}. By setting $\epsi{}$ small, Theorem~\ref{upperlowerbound} shows  that for both \Adam~and \Sgd, the upper and lower bounds of their expected escaping time $\te$  
are  at the order of  $\mathcal{O}\big( \frac{1}{m(\W) \Hui{\epsi{-1}}}\big)$. Note,  $m(\W)$ has different values for \Sgd~and \Adam~due to their different  $\Qmi{\wms}$ in Eqn.~\eqref{escapingset}.  If the escaping time $\te$ is very large, it  means that the algorithm cannot well escape from the basin $\G$ and would get stuck in $\G$. Moreover, given the same basin $\G$, if one algorithm has smaller escaping time $\te$ than other algorithms, then it is  more locally unstable and would faster escape from this basin  to others. In the following sections, we discuss   the effects of the geometry  adaptation and the gradient noise structure of \Adam~and \Sgd~to the escaping time $\te$ which are respectively reflected by the factors $m(\W)$ and $\Hui{\epsi{-1}}$. Our results show that \Sgd~has smaller escaping time  than~\Adam~and can better escape from local basins with small Radon measure to those with lager Radon measure.

\subsection{Preference to Flat Minima}\label{flatone}
To interpret Theorem~\ref{upperlowerbound},  we first define the  \textit{``flat" minima} in this work in terms of Radon measure.
\begin{defn}
	\!A minimum $\wms\!\in\!\Omegas$ is said to be  flat  if  its  basin $\Omegas$  has large nonzero  Radon measure.  \vspace{-0.6em}
\end{defn}
Due to  the  factor $m(\W)$ in Theorem~\ref{upperlowerbound},    both  \Adam~and \Sgd~have large escaping time $\te$ at the  {``flat" minima}. Specifically, if the  basin $\Omegas$  has  larger  Radon measure, then the  complementary  set   $\W^c=$ $\{\ym\! \in\! \Rs{d}\ | \ \Qmi{\wms}^{-1} \Sigmai{\wms}\ym \! \in\! \Grd\}$   of $\W$ also has  larger  Radon measure. Meanwhile, the Radon measure on $\W^c\!\cup\!\W$  is a constant, meaning the larger  $m(\W^c)$ the smaller $m(\W)$. So     \Adam~and \Sgd~have larger  escaping time at ``flat"  minima. Thus,  they would escape  ``sharp" minima due to their smaller escaping time, and  tend to converge to  ``flat" ones.   Since for  basin $\Omegas$, its  Radon measure    positively relies on its volume,   $m(\W)$ negatively depends on the volume of   $\Omegas$. So  \Adam~and \Sgd~are more stable at the minima with larger basin $\Omegas$ in terms of volume. This can be intuitively understood: for the process $\wmi{t}$, the volume of the basin determines the necessary jump size of  the \levyp~motion $\levyi{t}$ in the SDEs to escape, which means the larger the basin the harder for an algorithm to escape.

\begin{wrapfigure}{r}{0.37\linewidth}
	\vspace{-1.9em}
	\begin{center}
		\setlength{\tabcolsep}{0.1pt} 
		\begin{tabular}{c}
			\includegraphics[width=0.96\linewidth]{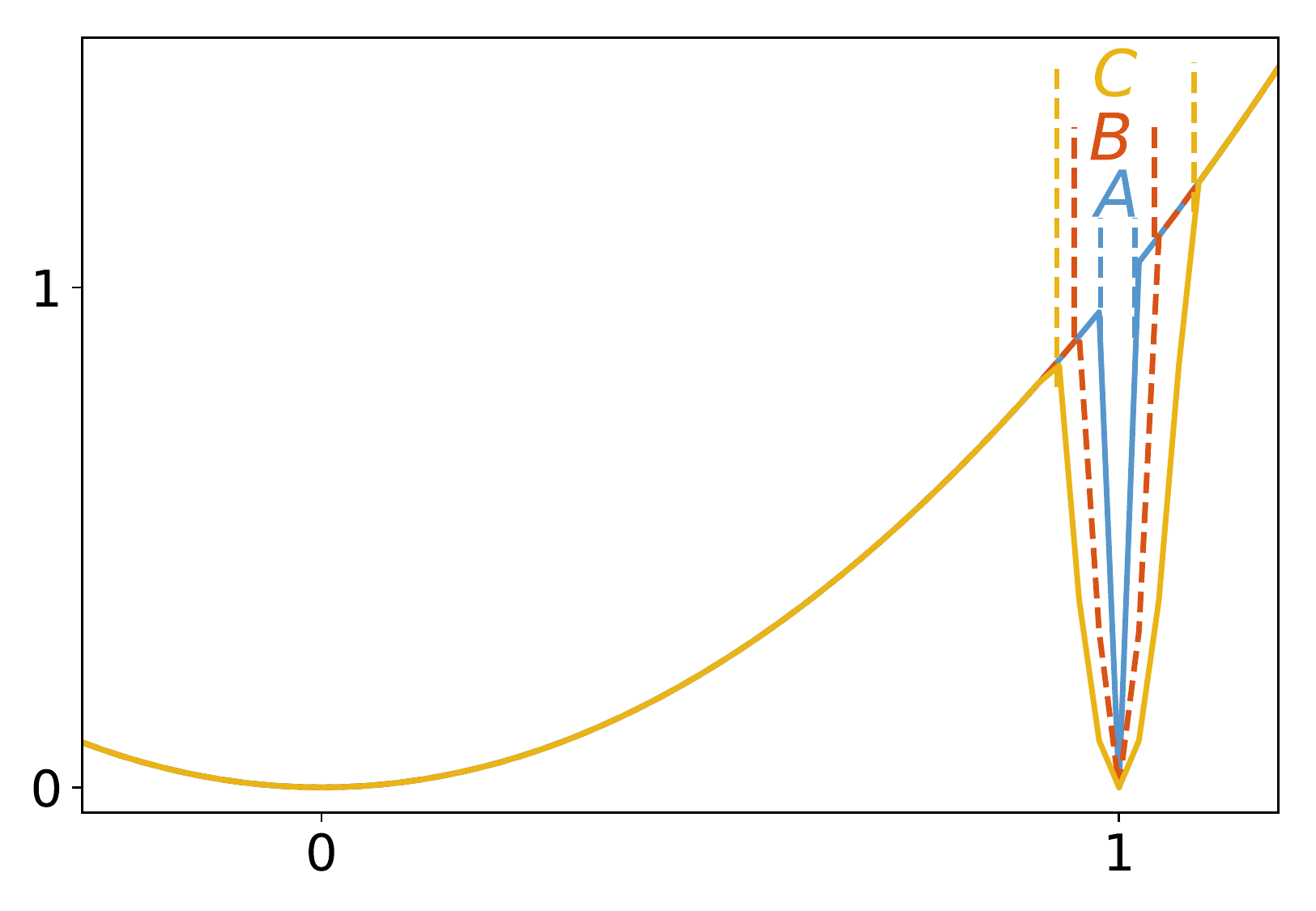}
		\end{tabular}
	\end{center}
	\vspace{-0.8em}
	\caption{$f(x)\!=\!\min(x^2, a(x-1)^2)$
	}
	\label{illustration_relations}
	\vspace{-1.1em}
\end{wrapfigure}
To investigate the above conclusion,  namely, positive dependence of the escaping time $\te$ to the Radon measure of a basin, we  construct a function  $f(x)=\min(x^2, a(x-1)^2)$ which has two local  basins at the points $x=0$ and $x=1$ as illustrated in Fig.~\ref{illustration_relations}.  By setting $a= 10^5, 500, 150$, we obtain three basins $A$, $B$ and $C$, where  their Radon measures obey $m(A)< m(B)<m(C)$ because  their volumes satisfies $\textsc{V}(A)< \textsc{V}(B)<\textsc{V}(C)$. Then we run SDE of SGD with initialization $x_0=1$ for 2000 iterations, and repeat 1000 times. For the three basins $A$, $B$ and $C$ with same \levyp~noise, the escaping probabilities of SDE for jumping outside are $100\%$, $65.6\%$ and $10.1\%$, and the average iterations for successful escaping on $A$, $B$ and $C$ are $122$, $457$ and $1898$.  For SDE of \Adam~and \Sgdm (SGD with momentum), we obtain similar observations and do not report them for avoid needless duplication. These results confirm our theory: the larger Radon measure of the basin, the harder to escape.

Note the ``flat" minima here is defined  on Radon measure, and differ from the conventional flat ones  whose local basins have no large curvature (no large eigenvalues in its Hessian matrix). In most cases, the flat minima here consist of the conventional flat ones and the minma at the  asymmetric basins/valleys since   local basins of these minima often have large volumes and thus larger Radon measures. Accordingly, our theoretical results can well explain the phenomenons observed in many works~\cite{keskar2017improving,merity2017regularizing,loshchilov2016sgdr,wilson2017marginal,sagun2017empirical,he2019asymmetric} that \Sgd~often converges to the minima at the flat or asymmetric valleys which is interpreted by our theory to have larger Radon measure and  attract \Sgd~to stay at these places.  In contrast,  the conventional flat argument cannot explain asymmetric valleys, as asymmetric valleys means sharp minima under the conventional  definition and should be avoided by \Sgd.

\subsection{Analysis of Generalization Gap between \Adam~and \Sgd}\label{generalizationgap}
Theorem~\ref{upperlowerbound}  can also well explain the generalization gap between  \Adam-alike adaptive  algorithms and \Sgd.  That is,  compared with~\Sgd, the minima given by \Adam~often 
suffer from worse test  performance~\cite{keskar2017improving,merity2017regularizing,loshchilov2016sgdr,wilson2017marginal,luo2019adaptive}. 
On one hand, the  observations in~\cite{keskar2016large,hochreiter1997flat,izmailov2018averaging,li2018visualizing}  show that the  minima at the flat or asymmetric basins/valleys  often enjoy better generalization performance than others.  On the other hand, Theorem~\ref{upperlowerbound} shows that  \Adam~and \Sgd~can escape sharp minima  to  flat ones with larger Radon measure. As aforementioned,  flat minima in terms of  Radon measure often refer to the minima  at the flat or asymmetric basins/valleys. This    implies that if one algorithm can escape the current minima faster, it is more likely for the algorithm to find flatter minima. These results together show the benefit of faster escaping behaviors of an algorithm to its generalization performance.

\begin{figure}[tb]
	\begin{center}
		\setlength{\tabcolsep}{0.0pt} 
		\begin{tabular}{cc}
			{\hspace{-2pt}}
			\includegraphics[width=0.5\linewidth]{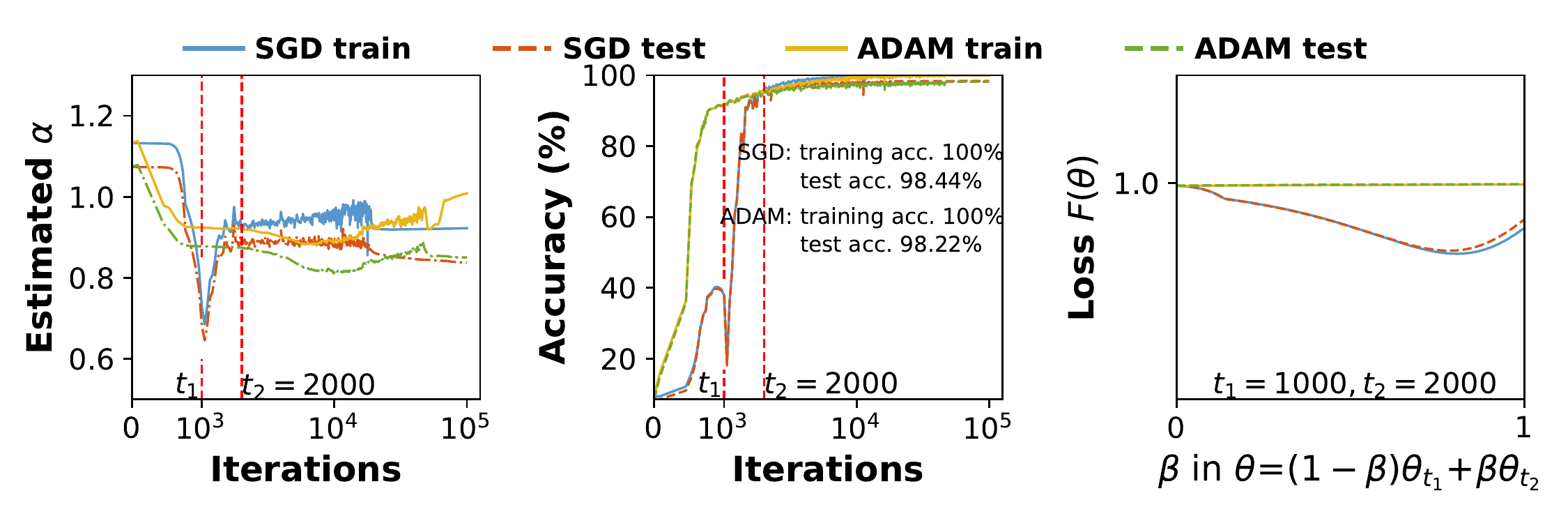}& \includegraphics[width=0.5\linewidth]{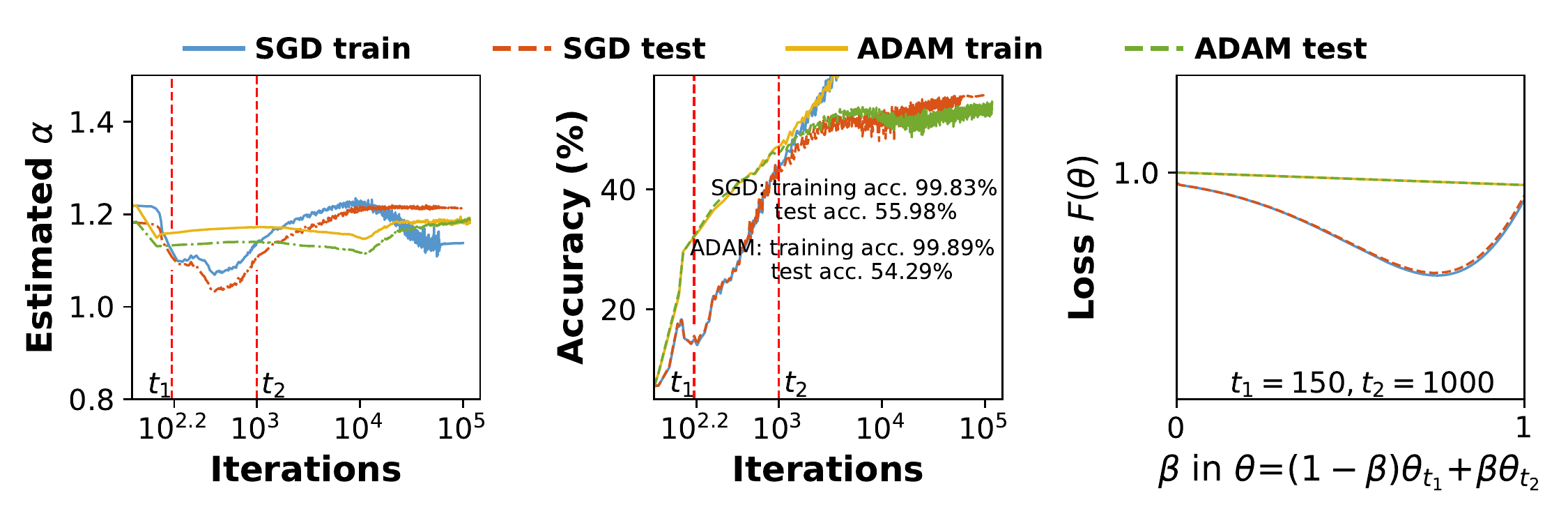}\vspace{-0.0em}\\
			{\small{(a) MNIST}} & {\small{(b) CIFAR10}} 
		\end{tabular}
	\end{center}
	\vspace{-0.3em}
	\caption{Behaviors illustration of \Sgd~and \Adam~on fully connected networks. In both (a) and (b), the left and middle  figures respectively report the estimated tail index $\alpha$ in $\Sas$ distribution and classification accuracies;  right figures show  possible barriers between the solutions $\wmi{1000}$ and $\wmi{2000}$ on MNIST, and $\wmi{150}$ and $\wmi{1000}$ on CIFAR10, respectively. \textbf{Best viewed in $\times$2 sized color pdf file.}
	}
	\label{illustration_iteration}
	\vspace{-1.0em}
\end{figure}

According to  Theorem~\ref{upperlowerbound},  two main factors, \ie~the gradient noise and geometry adaptation respectively reflected by the factors $\Hui{\epsi{-1}}\!=\!\frac{2}{\alpha} \epsi{-\alpha}$ and $m(\W)$,  affects the escaping time $\te$ of both \Adam~and \Sgd. We first look at the factor $\Hui{\epsi{-1}}$ in the escaping time $\te$.  As illustrated in Fig.~\ref{illustration_iteration} in Sec.~\ref{experiments}, the gradient noise in \Sgd~enjoys  very similar  tail index $\alpha$ with \Adam~for most optimization iterations, but it  has much smaller tail index $\alpha$ than \Adam~ for some  iterations, which means  \Sgd~has  larger  \levyp~discontinuous jumps in these iterations and thus enjoys smaller escaping time $\te$. This different tail property of gradient noise in these algorithms  are caused by the following reason.  \Sgd~assumes the gradient noise  $\umi{t}\!=\!\nabla \Fm(\wmi{t}) \!-\! \nabla \Fbi{t}$ at one iteration has heavy tails, while \Adam~considers the exponential gradient noise $\frac{1-\betag}{1-\betag^t}  \sum_{i=0}^{t} \!\betag^{t-i} \umi{i}$ which indeed smooths  gradient noise over the iteration trajectory  and  prevents large occasional gradient noise. In this way, \Sgd~reveals  heavier tails of gradient noise than \Adam~and thus has smaller tail index  $\alpha$ for some optimization iterations, helping escaping behaviors. Moreover, to guarantee convergence, \Adam~needs to use smaller learning rate $\eta$ than \Sgd~due to the geometry adaptation in \Adam, \eg~default learning rate $10^{-3}$ in \Adam~and $10^{-2}$ in \Sgd, leading to  smaller $\varepi{}\!=\!\eta^{(\alpha-1)/\alpha}$ and thus larger escaping time $\te$ in \Adam. Thus, compared with \Adam,  \Sgd~is more locally unstable and will converge to  flatter minima which often locate at the flat or asymmetric basins/valleys and enjoy better generalization performance~\cite{keskar2016large,hochreiter1997flat,izmailov2018averaging,li2018visualizing}.

Besides, the factor $m(\W)$ also plays an important role in the generalization degeneration phenomenon of \Adam. W.o.l.g., assume the minimizer $\wms=\bm{0}$ in the basin $\G$. 
As the local basin $\Omegas$ is often small, following~\cite{wu2018sgd,zhu2018anisotropic} we adopt second-order Taylor expansion to approximate  $\Omegas$ as a quadratic basin with center $\wms$, \ie~ $\Omegas\!=\!\big\{\ym\ |\  \Fm(\wms) \!+\! \frac{1}{2} \ym^T\Hm(\wms) \ym \!\leq\! h(\wms)\big\}$  
with a basin height $h(\wms)$ and Hessian matrix $\Hm(\wms)$ at $\wms$.   
Then for \Sgd, since $\Qmi{\wms}=\Imm$ in Eqn.~\eqref{escapingset}, its corresponding escaping set $\W$ is 
\begin{equation}\label{sgd_escaptingset} 
\W_{\Sgd}=\big\{\ym\in\Rs{d}\ | \ \ym^T   \Sigmai{\wms}\Hm(\wms)  \Sigmai{\wms} \ym \geq h_f^*\big\}
\end{equation}
with $h_f^*\!=\!2(h(\wms\!)\!-\!  \Fm(\wms\!))$, while according to Eqn.~\eqref{escapingset}, \Adam~has escaping set 
\begin{equation}\label{adamespcapingset} 
\W_{\!\Adam}=\{\ym\in\Rs{d}\ \! | \ \! \ym^T   \Sigmai{\wms}\Qmi{\wms}^{-1} \Hm(\wms)  \Qmi{\wms}^{-1} \Sigmai{\wms}\ym \geq h_f^* \}.
\end{equation} 
Then we  prove that for most time interval except the jump time, the current variable $\wmi{t}$ is indeed 
close to the minimum $\wms$. Specifically,  we first decompose the \levyp~process $\levyi{t}$ into two components  $\xii{t}$ and  $\zetai{t}$, \ie~$\levyi{t}  =  \xii{t}  +  \zetai{t}$, with the jump sizes $\|\xii{t}\|\!<\! \epsi{-\delta}$ and $\|\zetai{t}\|\!\geq \!\epsi{-\delta}$ ($\delta\!\in\!(0,1)$).  In this way, the stochastic process $\xii{}$ does not departure from $\wmi{t}$ a lot due to its limited jump size. The process $\zetai{}$ is a compound Poisson process with intensity
$\psiis{\epsi{}}{\delta}\!  =\! \larger\int_{\|\ym\|\geq \epsi{-\delta}} \!\nu(\ds \ym) \!=\! \larger{\int}_{\|\ym\|\geq \epsi{-\delta}} \frac{\ds \ym}{\|\ym\|^{1+\alpha}}\!=\!\frac{2}{\alpha} \epsi{\alpha\delta} $ and jumps distributed according to the law of $1/\psiis{\epsi{}}{\delta}$. Specifically, let $0\!=\!t_0\!<\!t_1\!<\!\cdots\!<\!t_k\!<\!\cdots$ denote the time  of successive jumps of $\zetai{}$. Then the inner-jump  time intervals $\jumptime{k}\!=$ $t_k-t_{k-1}$ are i.i.d. exponentially distributed random variables with mean value $\EE(\jumptime{k}) \!=\! \frac{1}{\psiis{\epsi{}}{\delta}}$ and  probability  function $\PPi{\jumptime{k} \!\geq\! x} $ $=\!\exp(-x \psiis{\epsi{}}{\delta})$. 
Based on this decomposition, we  state our results in Theorem~\ref{closedistance}.
\begin{thm}\label{closedistance}
	Suppose Assumptions~\ref{function_assumption} and~\ref{algorithm_assumption} hold.   Assume the process $\wgi{t}$ is produced by setting $\epsi{}=0$ in the \levyp-driven SDEs of \Sgd~and \Adam. 
	\\
	(1) $\wgi{t}$ exponentially converges to the minimizer $\wms$ in $\G$. Specifically,  by defining $\Delta\! =\! \Fm(\wmi{0})\!-\!\Fm(\wms)$, $\kappa_3\!=\!\frac{2\mu\tau}{ \betag \left(\vmax+\thres\right) +\mu\tau} \big(\betag -\frac{\betav}{4} \big)$ in \Adam~and  $\kappa_3=2\mu$ in \Sgd, for any $\bar{\rho}>0$, it satisfies 
	\begin{equation*} 
	\|\wgi{t}-\wms\|_2^2  \leq \epsi{\rhoa} \ \quad \text{if}\ \ t\geq \ve \triangleq  \kappa_3^{-1} \ln\big( 2\Delta\mu^{-1} \epsi{-\rhoa}\big).
	\end{equation*}
	(2) Assume 	$\delta \in (0,1)$,  $p=\min((\rhoa (1+c_3 \clip))/4, \bar{p})$,  $\rhoa=\frac{1-\delta}{16(1+c_4 \clip)}$, where $\kappa_1$ (in Theorem~\ref{upperlowerbound}), $\bar{p}$, $c_3$ and $c_4$ are four positive constants. When $\wmi{t}$ and $\wgi{t}$ have the same initialization $\wmi{0}= \wgi{0}$,  we have 
	\begin{equation*}\label{small_distance} 
	\sup\nolimits_{\wmi{0}\in\G} \mathbb{P}\big(\sup\nolimits_{0\leq t < \jumptime{1}} \|\wmi{t}- \wgi{t}\|_2\geq 2\epsi{\rhoa}\big) \leq 2\exp\big(-\epsi{-p}\big).
	\end{equation*}
\end{thm}
See its proof in Appendix~\ref{proofofclosedistance}. By inspecting the first part of Theorem~\ref{closedistance}, one can observe  that  the gradient-noise-free processes  $\wgi{t}$ produced by setting $\epsi{}=0$ in the \levyp-driven~SDEs of \Sgd~and \Adam~locate in a very small neighborhood of the minimizer $\wms$ in the local basin $\G$ after a very small time interval $ \ve =\kappa_3^{-1} \ln\big( 2\Delta\mu^{-1} \epsi{-\rhoa}\big)$. The second part of Theorem~\ref{closedistance} shows  that before the first jump time $t_1=\sigma_1$ of the jump $\zetai{}$  with size larger than $\epsi{-\delta}$ in \levyp~motion $\levyi{t}$, the distance between $\wmi{t}$ and $\wgi{t}$ is very small. So  these two parts together guarantee small distance between  $\wmi{t}$ and $\wms$ for the most time interval before the first big jump in the \levyp~motion $\levyi{t}$ since the mean jump time $\EE(\jumptime{1}) = \frac{\alpha}{2\epsi{\alpha\delta}}=\mathcal{O}(\epsi{-1})$ of the first big jump is much larger than $\ve=\mathcal{O}(\ln(\epsi{-1}))$ when $\epsi{}$ is  small. Next  after the first big jump, if $\wmi{t}$ does not escape from the local basin $\Omegas$,    by using the first part of   Theorem~\ref{closedistance}, after the time interval $\ve$, $\wmi{t}$ becomes close to $\wms$ again. This process will continue until the algorithm escapes from the  basin. So for most time interval before escaping from $\Omegas$, the stochastic process $\wmi{t}$ locates in a very small neighborhood of the minimizer $\wms$.    

The above analysis results on Theorem~\ref{closedistance} hold  for moderately ill-conditioned local basins (ICLBs). Specifically, the analysis  requires $ \ve\!\leq \!\sigma_1$ to guarantee small distance of current solution $\wm_t$ to $\wms$ before each big jump.  So if  $\mu$ of ICLBs is larger than $\mathcal{O}(\epsi{\alpha\delta})$ which is very small as $\epsi{}$ in SDE is often small to precisely mimic  algorithm behaviors, The above analysis results~\ref{closedistance} still hold.  Moreover,  to obtain the result (1) in Theorem~\ref{closedistance}, we assume the optimization trajectory goes along the eigenvector direction corresponding to $\mu$ which is the worse case and leads to the worst convergence speed. As the measure of  one/several eigenvector directions on high dimension is 0,  optimization trajectory cannot always go  along the eigenvector direction corresponding to $\mu$. So  $\ve$ is actually much larger than $\mathcal{O}\big(\frac{1}{\mu} \ln(\frac{1}{\mu  \epsi{\delta}})\big)$, largely improving applicability of our theory.   For extremely ICLBs ($\mu\rightarrow 0$ or $\mu=0$),  the above analysis does not hold which accords with the previous results that first-order gradient algorithms cannot escape from them provably~\cite{anandkumar2016efficient}. Fortunately,  $\mu\rightarrow 0$ and $\mu=0$ give   asymmetric basins which often  generalize well~\cite{he2019asymmetric,sagun2017empirical}  and are not needed to escape.

By using the above results, we have $\wmi{t}\!\approx\! \wms$ before  escaping and thus $\vmi{t} $ $= \!\lim_{\wmi{t}\rightarrow \wms}\! [\nabla \Fbi{t}]^2$. Considering the randomness of the mini-batch $\Bi{t}$, $\omegai{t} \!\approx\! 1$ and $\epsilon\!\approx\! 0$, we can  approximate 
\begin{equation*}\label{Qstar}
\EE[\Qmi{\wms}] \approx \EE\big[\lim\nolimits_{\wmi{t}\rightarrow \wms}  \mbox{diag}\big(\sqrt{\omegai{t}\vmi{t}}\big)\big]\approx \mbox{diag} \left({\sqrt{ \frac{1}{n} \sum\nolimits_{i=1}^n[\nabla f_i(\wms)]^2}}\ \right).
\end{equation*}  
Meanwhile,  since
$\Sigmai{\wms} = \frac{1}{ \batchsize}  \Sigmabi{\wms}$ because of   $\lim_{\wmi{t}\rightarrow\wms}\! \Fm(\wmi{t})=\bm{0}$ where $\Sigmabi{\wms} = \frac{1}{n}$ $ \sum\nolimits_{i=1}^n \! \nabla f_i(\wms) \nabla f_i(\wms)^T$, one  can approximately compute  $\EE[\Sigmai{\wms}\Qmi{\wms}^{-1}]  \approx  \frac{1}{\batchsize} \Imm.$   Plugging this result into the escaping set $\W_{\Adam}$ yields
\begin{equation*}\label{adap_escapingset}
\W_{\Adam}\approx   \Big\{\ym\in\Rs{d}\ \big| \  \ym^T  \Hm(\wms) \ym \!\geq\! \batchsize^2 h_f^* \Big\}.
\end{equation*}
Now we compare  the escaping sets $\W_\Sgd$ of \Sgd~and $\W_{\Adam}$ of \Adam. For clarity, we  re-write $\W_\Sgd$ in Eqn.~\eqref{sgd_escaptingset} as
\begin{equation*} 
\W_{\Sgd}= \Big\{\ym\in\Rs{d}\ \big| \ym^T   \Sigmabi{\wms}\Hm(\wms)  \Sigmabi{\wms} \ym \!\geq\! \batchsize^2 h_f^*\Big\}.
\end{equation*}
By comparison, one can observe that for \Adam, its gradient noise does not affect the escaping set $\W_{\Adam}$ due to the geometry adaptation via scaling each gradient coordinate, while for \Sgd,  its gradient noise plays an important role. Suppose $\bar{\Hm}(\wms)\!=\!\Sigmabi{\wms}\Hm(\wms)  \Sigmabi{\wms}$, and the singular values of $\Hm(\wms)$ and $\Sigmabi{\wms}$ are respectively $\lambda_1 \!\geq \!\lambda_2\!\geq\!\cdots\!\geq\! \lambda_d$ and $\varsigma_1\!\geq \!\varsigma_2\!\geq\!\cdots\!\geq \!\varsigma_d$.  Zhu~\et~\cite{zhu2018anisotropic} proved that  $\Sigmabi{\wms}$ of \Sgd~on deep neural networks well aligns the Hessian matrix $\Hm(\wms)$, namely the  top eigenvectors associated with large eigenvalues in $\Sigmai{\wms}$ have similar directions in those in $\Hm(\wms)$. Besides, for modern over-parameterized neural networks, both Hessian $\Hm(\wms)$ and the gradient covariance matrix $\Sigmai{\wms}$ are ill-conditioned and anisotropic near minima~\cite{sagun2017empirical,chaudhari2018stochastic}. Based on these results, we can approximate the singular values of $\bar{\Hm}(\wms)$ as $\lambda_1\varsigma_1^2 \!\geq\! \lambda_2\varsigma_2^2 \!\geq\!\cdots\!\geq\! \lambda_d\varsigma_d^2$, implying that $\bar{\Hm}(\wms)$ becomes  much more singular than $\Hm(\wms)$. Then the volume of the component set $\W_{\Adam}^c$ of $\W_{\Adam}$ is $\textsc{V}(\W_{\Adam}^c) \!=\! \zeta{\prod_{i=1}^d}  \lambda_i$  where $\zeta \!=\!  2d^{-1}(\pi\batchsize/h_f^*)^{d/2} g^{-1}(d/2)$ with a gamma function $g$. Similarly, we can obtain the volume $\textsc{V}(\W_{\Sgd}^c) $ $ =\!\zeta{\prod_{i=1}^d} \lambda_i \varsigma_i^2$  of the  component set $\W_{\Sgd}^c$  of $\W_\Sgd$. As aforementioned,   covariance matrix $\Sigmai{\wms}$ is ill-conditioned and anisotropic near minima and has only a few larger singular values~\cite{sagun2017empirical,chaudhari2018stochastic},  indicating ${\prod_{i=1}^d} \varsigma_i^2 \!\ll\! 1$.  So  $\textsc{V}(\W_{\Sgd}^c)$  is actually much smaller than $\textsc{V}(\W_{\Adam}^c)$. Hence $\W_{\Sgd}$   has larger volume than  $\W_{\Adam}$ and thus has larger Radon measure $m(\W_{\Sgd})$  than  $m(\W_{\Adam})$. Accordingly, \Sgd~has smaller  escaping time at the local basin $\Omegas$ than \Adam.  
Thus, \Sgd~would escape from $\Omegas$ and converges to flat minima whose local basins have large Radon measure, while \Adam~will get stuck in $\Omegas$. Since  flat minima with large Radon measure usually locate at the flat or asymmetric basins/valleys and generalize better~\cite{loshchilov2016sgdr,keskar2017improving,merity2017regularizing,wu2018group,wilson2017marginal}, \Sgd~often enjoys better testing performance.  From the above analysis, one  can also observe that for \Sgd, the  covariance matrix $\Sigmai{\wms}$  helps increase Radon measure $m(\W_{\Sgd})$ of $\W_{\Sgd}$. So anisotropic gradient noise~helps \Sgd~escape from the local basin but cannot help \Adam's escaping behaviors.

\textbf{Discussion on \Sgdm.} Our theory also indicates that \Sgd~with momentum (\Sgdm) can generalize better than \Adam. Here we discuss it in an intuitive way.  Specifically, as  \Sgdm~does not adapt the geometry,   under the same assumption,  it has the following  \levyp~SDE with $ \Qmi{t}\!=\!\Imm$:
	\begin{equation}\label{assumption_stochastic_adamsda}
		\mbox{d} \wmi{t} =  -\mui{t} \Qmi{t}^{-1}\mmi{t} \!+\! \varepi{}\Qmi{t}^{-1} \Sigmai{t}\mbox{d} \levyi{t}, \ \ 
		\mbox{d} \mmi{t} =   \betag(\nabla\Fm(\wmi{t})  \!-\! \mmi{t}),\ \ 
		\mbox{d} \vmi{t} =  \betav([\nabla \Fbi{t}]^2\!-\!\vmi{t}).
	\end{equation}
	Then we follow Eqn.~\eqref{escapingset}  and obtain  escaping set  $\W\!=\!\{\ym\!\in\!\Rs{d} |  \Qmi{\wms}^{-1} \Sigmai{\wms}\ym\! \notin\!\Grd\}$ of SGD-M, where $\Qmi{\wms}\!=\!\Imm$ and $\Sigmai{\wms}\!=\!\lim_{\wmi{t}\rightarrow \wms}\! \Sigmai{t}$.  Since Adam has the same SDE~\eqref{assumption_stochastic_adamsda} except  $\Qmi{t}\!=\!\diag{\sqrt{\omegai{t}\vmi{t}} +\thres}$ and same escaping set $\W$ except $\Qmi{\wms}\!\!=\!\lim_{\wmi{t}\rightarrow \wms}\! \Qmi{t}$, we can directly derive the escaping time $\Gamma=\mathcal{O}\big( \frac{1}{m(\W) \Hui{\epsi{-1}}}\big)$ of SGD-M with $\Hui{\epsi{-1}}\!=\!\frac{2}{\alpha} \epsi{\alpha}$. 

As SGD-M and Adam use the same gradient estimation $\mmi{t}$, their gradient noise have the same tail index $\alpha$ and thus the same factor $\Hui{\epsi{-1}}$.  For  $m(\W)$, due to  different escaping sets $\W_{\mbox{\tiny{SGD-M}}}$ of SGD-M and $\W_{\mbox{\tiny{Adam}}}$ of Adam,  $m(\W_{\mbox{\tiny{SGD-M}}})$ in SGD-M  differs from $m(\W_{\mbox{\tiny{Adam}}})$  in Adam. By observation,  $\W_{\mbox{\tiny{SGD-M}}}$ is as same as escaping set $\W_{\mbox{\tiny{SGD}}}$ of  SGD  in Eqn.~\eqref{escapingset} in manuscript, as  SGD(-M) have no geometry adaptation.   Then  Sec. 4.2  proves $\W_{\mbox{\tiny{SGD}}}$  has much larger volume than $\W_{\mbox{\tiny{Adam}}}$.   So  $m(\W_{\mbox{\tiny{SGD-M}}})$ is much larger than  $m(\W_{\mbox{\tiny{Adam}}})$. Thus, SGD-M has much smaller escaping time than Adam at the same basin, and  can better escape sharp minima to flat ones for better generalization. 

\section{Experiments}\label{experiments}
In this section, we first investigate the gradient noise in \Adam~and \Sgd, and then show  their iteration-based  convergence behaviors to testify the  implications of our escaping theory.  The code is available at  \url{https://panzhous.github.io}.

\textbf{Heavy Tails of Gradient Noise.}  We respectively use  \Sgd~and \Adam~to train AlexNet~\cite{krizhevsky2012imagenet} on CIFAR10, and  show the statistical behaviors of  gradient noise on CIFAR10.  
To fit the noise via $\Sas$ distribution, we consider covariance matrix $\Sigmai{t}$ and use the approach in~\cite{simsekli2019tail,mohammadi2015estimating} to estimate the tail index $\alpha$. 
Fig.~\ref{illustration_grdientnoise} in Sec.~\ref{introduction} and  Fig.~\ref{illustration_grdientnoise222} in Appendix~\ref{constructiondetails} show that the gradient noise in both \Sgd~and \Adam~usually reveal the heavy tails  and can be well characterized by  $\Sas$ distribution.  This  testifies the heavy tail assumption on the gradient noise in our theories.

\textbf{Escaping Behaviors.}  We investigate the iteration-based convergence behaviors of \Sgd~and \Adam, including their training and test accuracies and losses and tail index of their gradient noise.  For MNIST~\cite{lecun1998} and CIFAR10~\cite{krizhevsky2009learning}, we respectively use nine- and seven-layered fully-connected-networks. Each layer has 512 neurons and contains a linear layer and a ReLu layer.  
Firstly, the results in the middle figures show that \Sgd~usually has better generalization performance than \Adam-alike adaptive  algorithms which is consistent with the results in~\cite{wilson2017marginal,keskar2017improving,merity2017regularizing,luo2019adaptive}.

Moreover, from the   trajectories of the tail index $\alpha$ and  accuracy of \Sgd~on MNIST and CIFAR10 in Fig.~\ref{illustration_iteration}, one can observe  two distinct phases. Specifically, for the first 1000 iterations in MNIST and 150 iterations in CIFAR10, both the training and test accuracies  increase tardily, while the tail index parameter $\alpha$ reduces quickly. This process continues until $\alpha$ reaches its lowest value. When considering the barrier around  inflection point (\eg~a barrier between $\wmi{1000}$ and $\wmi{2000}$ on MNIST), it seems that the process of \Sgd~has a sudden jump from one basin to another one which leads to a sudden accuracy drop,  and then gradually converges. Accordingly,  the accuracies are  improved quickly.  In contrast, one cannot observe similar phenomenon in \Adam.  This is because as our theory suggested, \Sgd~is more locally unstable and converges to  flatter minima  than \Adam, which is caused by the geometry adaptation, exponential gradient average and smaller learning rate in \Adam.   All these results are consistent with our theories and also explain the well observed evidences in~\cite{keskar2017improving,merity2017regularizing,loshchilov2016sgdr,wu2018group,wilson2017marginal} that \Sgd~usually converges to  flat minima which often locate at the flat or asymmetric basins/valleys,  while \Adam~does not.  Because the empirical observations~\cite{keskar2016large,hochreiter1997flat,izmailov2018averaging,li2018visualizing} show that  minima at the flat or asymmetric basins/valleys often generalize better than sharp ones, our empirical and theoretical results can well explain the generalization gap between \Adam-alike algorithms and \Sgd.

\section{Conclusion}
In this work, we  analyzed the generalization performance degeneration of \Adam-alike adaptive  algorithms over \Sgd.  By looking into the local convergence behaviors of the  \levyp-driven SDEs of these algorithms through analyzing their escaping time, we  prove that for the same basin, \Sgd~has smaller escaping time than \Adam~and tends to converge to  flatter minima whose local basins have larger Radon measure, explaining its better generalization performance.  This result  is also consistent       with the widely observed convergence behaviors of \Sgd~and \Adam~in many literatures.   Finally our experimental results testify the heavy gradient noise assumption and implications in our theory.

\newpage
\section*{Broader Impacts}
\label{sec:impact}
This work  theoretically analyzes  a   fundamental problem in deep learning field, namely the generalization  gap between adaptive gradient algorithms and \Sgd, and reveals the essential reasons for the generalization degeneration of adaptive algorithms. The established theoretical understanding of these algorithms may inspire new algorithms with both fast convergence speed and good generalization performance, which  alleviate the need for computational resource and achieve state-of-the-art results. Yet it still needs more efforts to provide more insights to design practical algorithms.

{
\small
\bibliographystyle{unsrt}
\bibliography{referen}
}

 \newpage
 
 \appendix

\section{Structure of This Document} 
This supplementary document contains the technical proofs of convergence results and some additional numerical results of the  main draft   entitled ``Towards Theoretically Understanding Why  \Sgd~Generalizes Better Than \Adam~in Deep Learning''. It is structured as follows. In Appendix~\ref{constructiondetails}, we provides more construction details of the SDE for \Adam~and also conduct experiments which show very similar convergence behaviors of \Adam~(\Sgd) and its SDE.  Appendix~\ref{comparison} compares our work with the related work~\cite{dinh2017sharp,zhang2019adam} in more details.  
Appendix~\ref{notations} summarizes the notations throughout this document and also provides the auxiliary theories and lemmas for subsequent analysis whose proofs are deferred to Appendix~\ref{proofofAuxiliaryLemma}.  Then Appendix~\ref{proofofvolumeanalysis} gives the proofs of the main results in Sec.~\ref{volumeanalysis}, including  Theorem~\ref{upperlowerbound}  which analyzes the escaping time analysis of \levyp-driven SDEs and  Theorem~\ref{closedistance} which proves the processes with and without \levyp~motion are close to each other.  Finally, in Appendix~\ref{proofofAuxiliaryLemma} we presents the proofs of auxiliary theories and lemmas in Appendix~\ref{notations}, including Theorems~\ref{linearconvergenceSGD} $\sim$ \ref{linearconvergenceadam} and Lemmas~\ref{lemma2} $\sim$~\ref{lemm4}.

\section{More Discussion of SDE in~\Adam} \label{constructiondetails}
Here we provide more discussion and construction details for the SDE  in~\Adam.  We first investigate the second order moment of  the gradient noise in \Adam. Then we introduce the two types of randomness in the SDE of \Adam. Finally, we run experiments to investigate the validity of the constructed SDEs of \Adam~and \Sgd. 

\subsection{$\Sas$-distributed Gradient Noise in~\Adam}

\begin{figure*}[tb]
	\begin{center}
		\setlength{\tabcolsep}{0.8pt} 
		\begin{tabular}{cc}
			{\hspace{-2pt}}
			\includegraphics[width=0.5\linewidth]{fig1/Levy_fit_adam44.pdf} & 
			\includegraphics[width=0.5\linewidth]{fig1/Levy_fit_sgd44.pdf} \\
			{\small{(a) Gradient noise in  \Adam}} & {\small{(b) Gradient noise in  \Sgd}}  \\
			\includegraphics[width=0.5\linewidth]{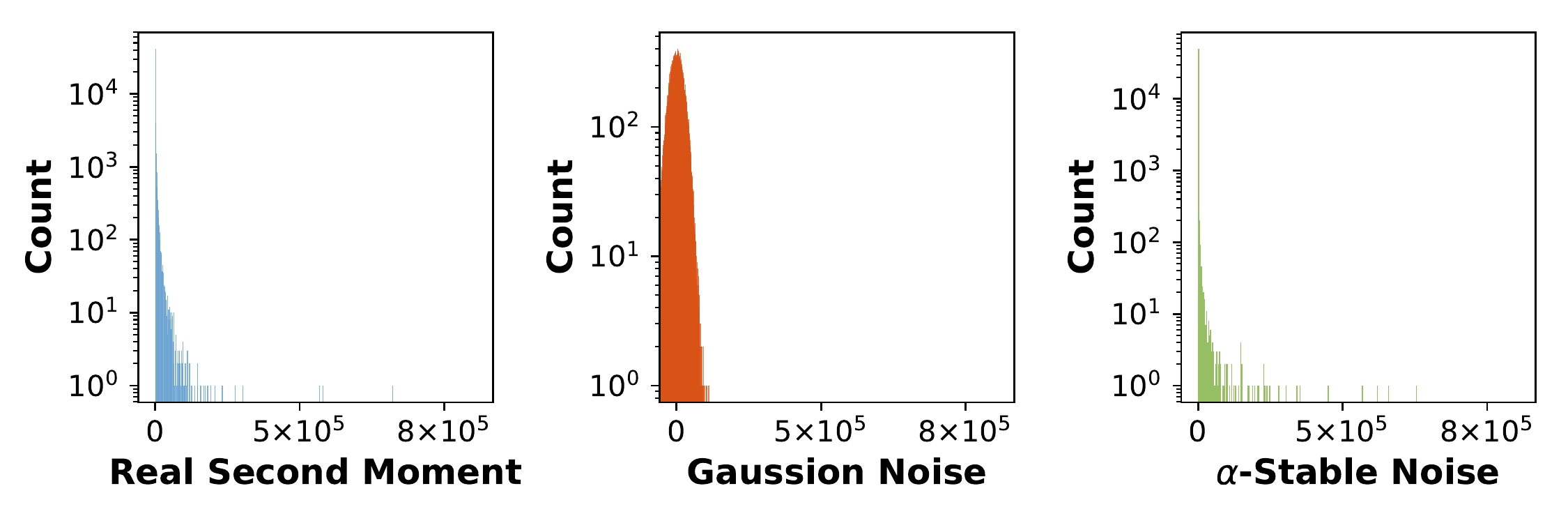} & 
			\includegraphics[width=0.5\linewidth]{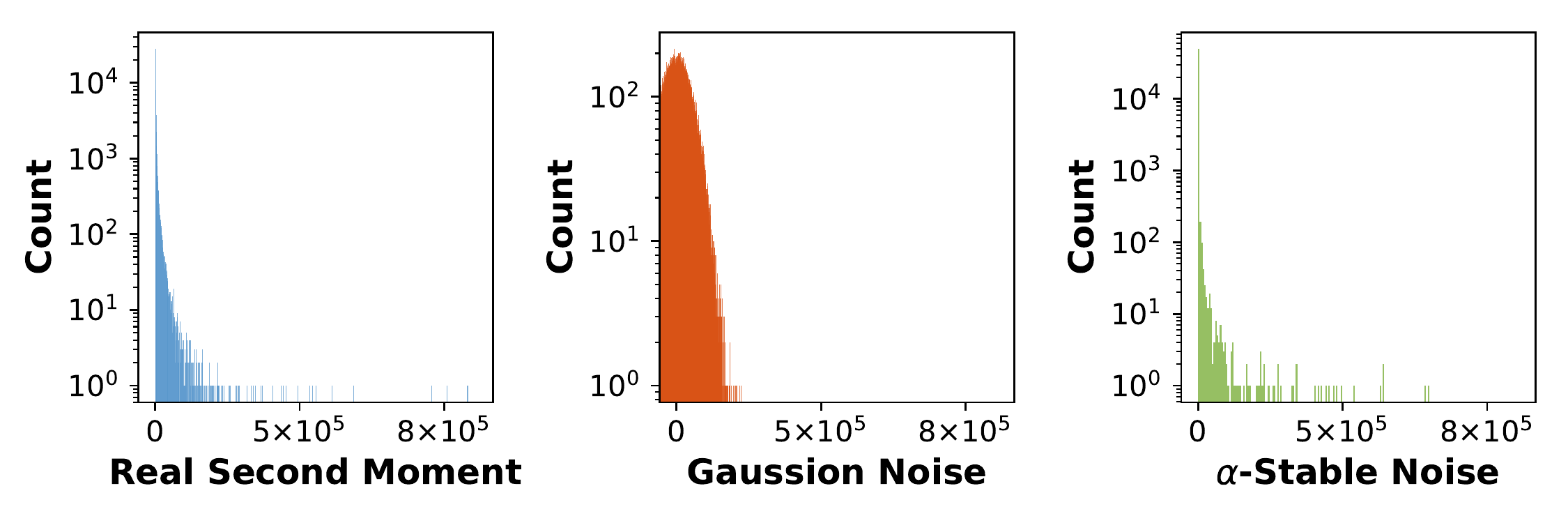} \\
			{\small{(c) Second-order moment of gradient noise in \Adam}} & {\small{(d) Second-order moment of gradient noise in \Sgd}}  \\
		\end{tabular}
	\end{center}
	\caption{Illustration of gradient noise in \Adam~and \Sgd. The left figures in (a) and (b) are the real gradient noise computed with AlexNet on CIFAR10. Similarly, the left figures in (c) and (d) are the second-order moment of gradient noise computed with AlexNet on CIFAR10. The middle and right figures in (a) $\sim$ (d)  are respectively the fitted Gaussian and systemic $\alpha$-stable noise. By comparison, $\alpha$-stable noise can better characterize real gradient noise in deep learning.}
	\label{illustration_grdientnoise222}
\end{figure*}

In the manuscript, we have shown the gradient noise itself to be $\Sas$-distributed. Here we further investigate the second-order moment of the gradient noise. From the bottom row of Figure~\ref{illustration_grdientnoise222}, one can observe that (1) both the second-order moment of the gradient noise also reveals heavy tails; (2) compared with Gaussian distribution, $\Sas$ distribution can better characterize this kind of  second-order moment of the gradient noise. All these results demonstrate that the gradient noise in both \Adam~and \Sgd~actually satisfies the $\Sas$ distribution. So the heavy-tailed gradient noise assumptions in our manuscript is very reasonable.

\subsection{Randomness in  SDE of \Adam}
The SDE of ADAM approximates gradient noise $\mmi{t}$ via the combination of full gradient and  \levyp~motion but does not approximate $\vmi{t}$. This SDE should be more accurate than the one which approximates both $\mmi{t}$  and the coefficients $\vmi{t}$. So the randomness in the SDE of \Adam~comes from the~\levyp~motion and also  $\vmi{t}$ caused by sampling a minibatch. But these two types of randomness actually do not depend on each other. Note that as shown in many literatures, \eg~\cite{bishop2019stability,kohatsu1997stochastic}, SDE allows randomness in coefficients and also enjoys many good properties, such as stability and unique solution. This type of SDE is usually called ``SDE with random coefficients", and usually appears in  stochastic jump systems~\cite{fang2002stabilization}, economics and finance~\cite{lim2002mean,turnovsky1973optimal}, 
biology~\cite{tiwari1976random,tsokos1974random}, mechanics and physics~\cite{finney1982random}, etc.  See more details of SDE with random coefficients in~\cite{bishop2019stability,kohatsu1997stochastic}.

\begin{figure*}[tb]
	\begin{center}
		\setlength{\tabcolsep}{0.8pt} 
		\begin{tabular}{cc}
			{\hspace{-2pt}}
			\includegraphics[width=0.95\linewidth]{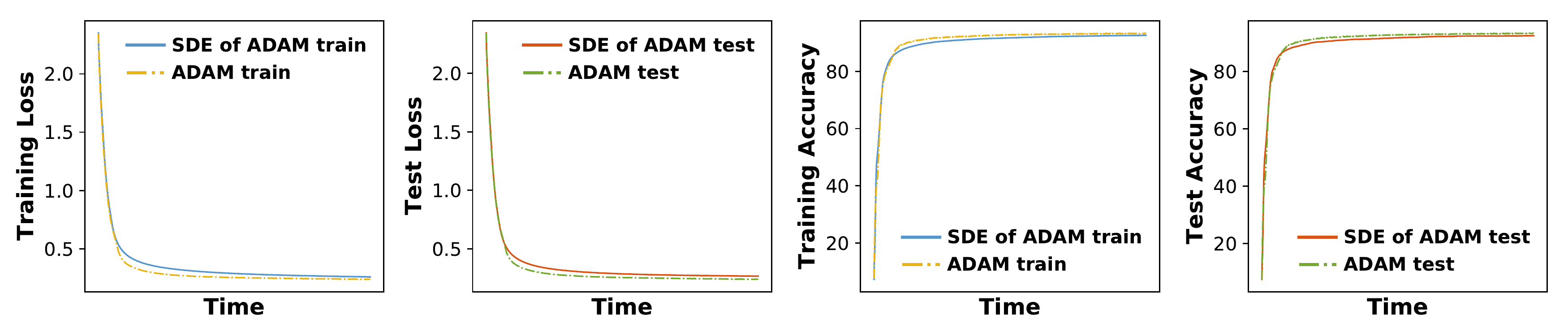} \vspace{0.8em}\\
			{\small{(a)  \Adam}}   \\
			\includegraphics[width=0.95\linewidth]{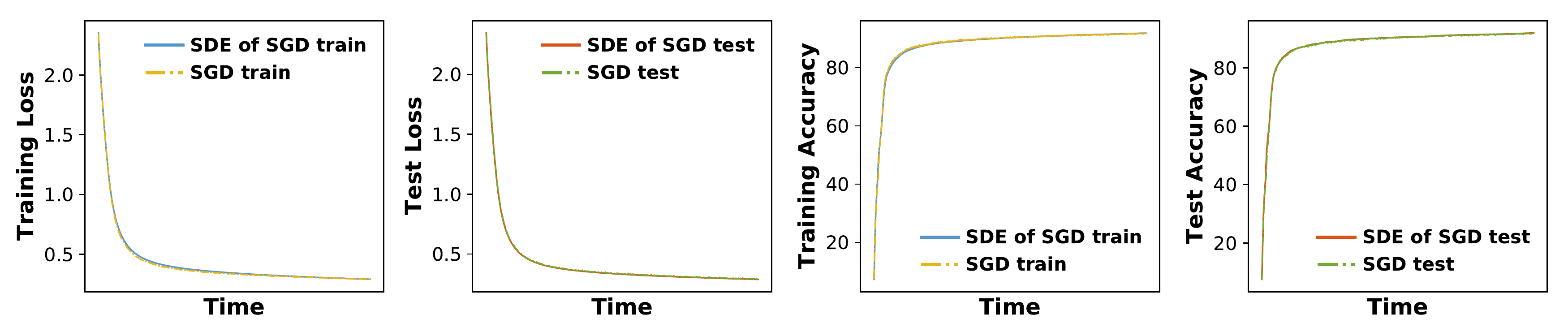}   \vspace{0.8em}\\
			{\small{(b) \Sgd}}   \\
		\end{tabular}
	\end{center}
	\caption{Illustration of convergence  trajectories of \Adam, \Sgd~and their SDEs. One can observe that for \Adam, it convergence trajectories are very similar to its SDE, which shows the validity of the SDE construction. Similarly, we can observe the same observations on \Sgd~and its SDE. }
	\label{illustration_trajectory}
\end{figure*}

\subsection{Convergence Behavior Comparison between Algorithm and Its SDE}
Here we conduct experiments on 784-10-10-sized networks  and report the convergence behvariors of \Adam~(\Sgd) and its SDE in Fig.~\ref{illustration_trajectory}. Note SDE actually equals to   injecting heavy tailed noise into   \Sgd~and \Adam~that use  full gradients. We use a relatively small network since simulating high-dimensional gradient noise $\umi{t}$ and computing the huge covariance matrix $\Sigmai{t}$ at each iteration are too computationally expensive to compute.   From the convergence trajectories of both \Adam~and its SDE in Fig.~\ref{illustration_trajectory} (a), one can observe that they have very similar convergence behaviors. Similarly, in Fig.~\ref{illustration_trajectory} (b) we can observe the same observations on \Sgd~and its SDE. So  injecting heavy tailed noise into   \Sgd~and \Adam~that use  full gradients   leads to   similar convergence behaviors  to  \Sgd~and \Adam~that use  stochastic gradients. These results well demonstrate  the validity of current SDE construction. Note that here we do not observe jump behaviors, since the networks are very small and may have not very sharp minima. But these results as aforementioned can  testify  the validity of current SDE construction. 

\section{Comparison to Related Works}\label{comparison}

Dinh et al.~\cite{dinh2017sharp} showed flat minimum can become sharp by scaling two layers at the same time. But with this scaling, sharp minimum cannot be arbitrarily flat, as if the eigenvalues of two parameters in the same layer has large ratio, this scaling cannot change this ratio. So flat and sharp minimum are not totally equivalent. Combining the observation in many works that flat minima could achieve better  generalization performance than sharper ones, one could conclude that flat minima can generalize well in most case, while sharp minima that can become flat one by linearly scaling two layers also can generalize but other sharp minima cannot. So analyzing the flat and sharp properties is still meaningful. Besides, the flatness in this work is defined on general non-zero Radon measure. If one finds an invariant measure to the  scaling in~\cite{dinh2017sharp}, the flatness is also invariant, providing more insights to generalization. So it is promising to  explore this invariant measure in the future.

\section{Notations and Auxiliary Lemmas}\label{notations}

\subsection{Notations}
For analyzing the uniform \levyp-driven SDEs in Eqn.~\eqref{assumption_stochastic_RMSP} and~\eqref{assumption_stochastic_adam}, we first decompose the \levyp process $\levyi{t}$ into two components $\xii{t}$ and  $\zetai{t}$, namely 
\begin{equation}\label{decompositionlevy}
\levyi{t} = \xii{t} + \zetai{t}
\end{equation}
whose characteristic functions are respectively defined as 
\begin{equation*}
\begin{split}
\EE [e^{i\langle \lam, \xii{t}\rangle }] =&e^{t  \mathlarger{\int}_{\Rs{d}\!\setminus\!\{\bm{0}\}} \zeta \indi{\|\ym\|\leq \frac{1}{\epsi{\delta}}}\nu(\ds \ym)},\\
\EE [e^{i\langle \lam, \zetai{t}\rangle }] =&e^{t\mathlarger{\int}_{\Rs{d}\!\setminus\!\{\bm{0}\}} \zeta \indi{\|\ym\|\geq \frac{1}{\epsi{\delta}}} \nu(\ds \ym)},
\end{split}
\end{equation*}
where $\zeta = e^{i\langle\lam,\ym \rangle}- 1 -i \langle \lam, \ym\rangle \indi{\|\ym\|\leq 1}$, $\epsi{}$  (in Eqn.~\eqref{assumption_stochastic_RMSP} and~\eqref{assumption_stochastic_adam}) and $\delta$ are two small constants satisfying $\epsi{-\delta} < 1$  and will be specified later. Define the  \levyp~measures $\nu$ as $\nu(\ds \ym) = \frac{1}{\|\ym\|^{1+\alpha}} \ds \ym$. 
Accordingly, the \levyp~measures $\nu$ of the stochastic processes $\xii{}$ and $\zetai{}$ are 
\begin{equation*}
\nu_{\xii{}} =\nu\big(\Am \cap \{0<\|\ym\|\leq \frac{1}{\epsi{\delta}}\}\big),\ \nu_{\zetai{}} =\nu\big(\Am \cap \{\|\ym\|\geq \frac{1}{\epsi{\delta}}\}\big),
\end{equation*}
where $\Am\in\mathcal{B}(\Rs{d})$. In this way, the stochastic process $\xii{}$ has infinite \levyp~measure with support $\{\ym\ |\ 0 <  \|\ym\|\leq \epsi{-\delta}\}$ and thus makes infinitely many jumps on any time interval. But the jump size does not exceed $\epsi{-\delta}$ and thus is small which actually does not help escape the current local basin. In contrast, the \levyp~measure $\nu_{\zetai{}}(\cdot)$ of $\zetai{}$ is finite and is computed as 
\begin{equation*}
\psiis{\epsi{}}{\delta}= \larger\int_{\|\ym\|\geq \epsi{-\delta}} \nu(\ds \ym) = \larger{\int}_{\|\ym\|\geq \epsi{-\delta}} \frac{\ds \ym}{\|\ym\|^{1+\alpha}}=\frac{2}{\alpha} \epsi{\alpha\delta}.
\end{equation*}
So the process $\zetai{}$ is a compound Poisson process with intensity $\psiis{\epsi{}}{\delta}$ and jumps distributed according to the law of $1/\psiis{\epsi{}}{\delta}$. Specifically, let $0=t_1<t_2\cdots<t_k<\cdots$ denote the times of successive jumps of $\zetai{}$ and $\jumpi{k}$ denote the jump size at the $k$-th jump. Then the inner-jump times $\jumptime{k}=t_k-t_{k-1}$ are i.i.d. exponentially distributed random variables with mean value $\EE(\jumptime{k}) = \frac{1}{\psiis{\epsi{}}{\delta}}$ and the probability distribution function $\PPi{\jumptime{k} \leq x}=1-\exp(-x \psiis{\epsi{}}{\delta})$. The probability law of $\jumpi{k}$ is also known explicitly in terms of the \levyp~measure $\nu$:
\begin{equation*}
\PPi{\jumpi{k}\!\in\!\Am}\!=\! \frac{1}{\psiis{\epsi{}}{\delta}} \nu\!\left(\Am\!\cap\!\{\ym\!\ |\ \!\|\ym\|\geq \epsi{-\delta}\}\right)\!,\ \Am\!\in\!\mathcal{B}(\Rs{d}).
\end{equation*}
So the main force for escaping the local basin comes from the big jumps in the process $\zetai{}$ which will be  rigorous  analyzed  in the following sections. 

Besides, for analysis, we usually need to consider affects of the \levyp~motion (noise) $\levyi{t}$ to the \levyp-driven SDEs of \Sgd~and \Adam~given in Eqn.~\eqref{assumption_stochastic_RMSP} and~\eqref{assumption_stochastic_adam}. So here we define two \levyp-free SDEs which  respectively correspond to Eqn.~\eqref{assumption_stochastic_RMSP} and~\eqref{assumption_stochastic_adam}:
\begin{equation}\label{deterministicversion_RMSP}
\ds \wgi{t} =  \nabla \Fm(\wgi{t}), 
\end{equation}
and 
\begin{equation}\label{deterministicversion_adam}
\begin{cases}
\ds \wgi{t} = & -\mui{t} \Qgi{t}^{-1}\mgi{t}, \\
\ds \mgi{t} = &  \betag(\nabla\Fm(\wgi{t})  - \mgi{t})\\
\ds \vgi{t} = & \betav(\nabla \Fbhi{t}^2 -\vgi{t}).\\
\end{cases}
\end{equation}
where $\Qgi{t}=\diag{\sqrt{\vgi{t}} +\thres}$.  Then by analyzing the distance $\|\wgi{t}-\wmi{t}\|$ between  the processes $\wgi{t}$ without \levyp~motion and $\wmi{t}$ with \levyp~motion, we can well know the effects of the \levyp~motion to the escaping behaviors.  
\subsection{Auxiliary Theories and Lemmas}

\begin{thm}\label{linearconvergenceSGD}
	Suppose Assumptions~\ref{function_assumption} and~\ref{algorithm_assumption} holds.  Then for  \levyp-driven \Sgd~SDE~\eqref{deterministicversion_RMSP} with $\Qgi{t}=\Imm$ and $\betav=0$, the Lyapunov function 
	$\LL(t) = \Fm(\wgi{t}) -\Fm(\wms)$ obeys 
	\begin{equation*}
	\begin{split}
	\LL(t)\leq \Delta \exp\left(- 2\mu t \right)
	\end{split}
	\end{equation*}
	where $\Delta=\Fm(\wgi{0}) -\Fm(\wms)$ with the optimum solution $\wms$ in the current local basin $\Omegas$. 
	The sequence $\{\wgi{t}\}$ produced by Eqn.~\eqref{deterministicversion_RMSP}  obeys
	\begin{equation*}
	\begin{split}
	\|\wgi{t}-\wms\|_2^2  \leq \frac{2\Delta}{\mu} \exp\left(- 2\mu t\right).
	\end{split}
	\end{equation*}
\end{thm}

See its proof in Appendix~\ref{linearconvergenceSGDproof}.

\begin{thm}\label{linearconvergenceadam}
	Suppose Assumptions~\ref{function_assumption} and~\ref{algorithm_assumption} holds.    Assume the sequence $\{(\wgi{t},\mgi{t},\vgi{t})\}$ are produced by Eqn.~\eqref{deterministicversion_adam}. Let $\smi{t}=\frac{h_t}{\mui{t}}\left(\sqrt{\omegai{t}\vgi{t}}+\thres \right)$ with $h_t= \betag$, $\mui{t}=(1-e^{-\betag t})^{-1}$ and $\omegai{t}=(1-e^{-\betav t})^{-1}$.  We define $\|\xm\|_{\ym}^2=\sum_i \ym_i\xm_i^2$.  Then for  \levyp-driven \Adam~SDEs in Eqn.~\eqref{deterministicversion_adam}, its    Lyapunov function 
	$\LL(t) = \Fm(\wgi{t}) -\Fm(\wms)+ \frac{1}{2} \|\mgi{t}\|_{\smi{t}^{-1}}$ with the optimum solution $\wms$ in the current local basin $\Omegas$ obeys 
	\begin{equation*}
	\begin{split}
	\LL(t)\leq \Delta \exp\left(-\frac{2\mu\tau}{ \betag \left( \vmax + \thres\right) +\mu\tau}   \left(\betag -\frac{\betav}{4} \right)t   \right)
	\end{split}
	\end{equation*}
	where $\Delta=\Fm(\wgi{0}) -\Fm(\wms)$ due to $\mgi{0}=\bm{0}$.
	The sequence $\{\wgi{t}\}$ produced by Eqn.~\eqref{deterministicversion_adam} obeys
	\begin{equation*}
	\begin{split}
	\|\wgi{t}-\wms\|_2^2  \leq \frac{2\Delta}{\mu} \exp\left(-\frac{2\mu\tau}{ \betag \left(\vmax + \thres\right) +\mu\tau}   \left(\betag -\frac{\betav}{4} \right)t   \right).
	\end{split}
	\end{equation*}
\end{thm}

See its proof in Appendix~\ref{linearconvergenceadamproof}.

\begin{lem}\label{lemma2}
	(1) The process $\xii{}$ in Eqn.~\eqref{decompositionlevy} can be decomposed into two processes $\xwi{}$ and linear drift, namely,
	\begin{equation}\label{adasfcsa}
	\xii{t}=\xwi{t}+\mue t,
	\end{equation}
	where $\xwi{}$ is a zero mean \levyp martingale with bounded jumps.  \\
	(2) Let $\delta \in(0,1)$, $\mue=\EE [\xii{1}]$ and $\Te=\epsi{-\theta}$ for some $\theta>0$,  $\rho_0=\rho_0(\delta)=\frac{1-\delta}{4}>0$ and $\theta_0=\theta_0(\delta)=\frac{1-\delta}{3}>0$.  Suppose $\epsi{}$ is sufficient small such that such that $\Hui{1}\leq \epsi{-\frac{1-\delta}{6}}$ and $\epsi{-\rho}-2(C+\Hui{1})\epsi{\frac{7}{6}(1-\delta)+\frac{\rho}{2}}\geq 1$ with a constant $C=\left|    \int_{0<u\leq 1}  u^2  \ds \Hui{u} \right| \in(0, +\infty)$.  Then for all $\delta\in(0,\delta_0)$, $\theta\in(0,\theta_0)$ there are $p_0=p_0(\delta)=\frac{\delta}{2}$ and $\epsi{}_0=\epsi{}_0(\delta,\rho)$ such that  the estimates 
	\begin{equation*}
	\|\epsi{}\xii{\Te}\| = \epsi{} \|\mue\|\Te < \epsi{2\rho}\quad \text{and}\quad \PPi{[\epsi{} \xii{}]_{\Te}^d\geq \epsi{\rho}} \leq \exp(-\epsi{-p})
	\end{equation*}
	hold for all $p\in(0,p_0]$ and $\epsi{}\in(0,\epsi{}_0]$. 
\end{lem}
See its proof in Appendix~\ref{proofoflemma2}. 

\begin{lem}\label{lemma3}
	Let $\delta\in (0,1)$ and $\gmi{t}_{t\geq 0}$ be a bounded adapted c\.{a}dl\.{a}g stochastic process with values in  $\Rs{d}$, $\Te=\epsi{-\theta}$, $\theta>0$. Suppose  $\sup_{t \geq 0} \|\gmi{t}\|$ is well bounded. Assume $\rho_0=\rho_0(\delta)=\frac{1-\delta}{16}>0$, $\theta_0=\theta_0(\delta)=\frac{1-\delta}{3}>0$,  $p_0=\frac{\rho}{2}$. For $ \xwi{t}$ in Eqn.~\eqref{adasfcsa}, there is  $\delta_0=\delta_0(\delta)>0$ such that for all $\rho\in(0,\rho_0)$ and $\theta\in(0,\theta_0)$, it holds 
	\begin{equation*}
	\PPi{\sup_{0\leq t \leq \Te}\epsilon \left| \sum_{i=1}^{d} \int_{0}^{t}  \gmii{s-}{i} \ds \xwii{s}{i} \right|\geq \epsi{\rho}} \leq 2\exp\left(-\epsi{-p}\right)
	\end{equation*}
	for all $p\in(0,p_0]$ and $0<\epsi{}\leq \epsi{}_0$ with $\epsilon_0 = \epsi{}_0(\rho)$, where $ \xwii{s}{i} $ denotes the $i$-th entry in $\xwi{s}$.
\end{lem}
See its proof of Appendix~\ref{proofoflemma3}.

\begin{lem}\label{lemm4} 
	Suppose Assumptions~\ref{function_assumption} and~\ref{algorithm_assumption} holds.     Assume 	$\delta \in (0,1)$,  $\rho_0=\rho_0(\delta)=\frac{1-\delta}{16(1+c_1 \clip)}>0$, $\theta_0=\theta_0(\delta)=\frac{1-\delta}{3}>0$,  $p_0=\min(\frac{\rhoa (1+c_1 \clip)}{2}, p)$,  $\frac{1}{c_2} \ln\left(\frac{2\Delta}{\mu \epsi{\rhoa}}\right)  \leq \epsi{-\theta_0}$ where $\clip=\ell$ and $c_2=2\mu$  in \Sgd,  	$\clip=\frac{c_2\ells}{(\vmin+\thres) |\taum-1|} $ and $c_3=\frac{2\mu\tau}{ \betag \left(\vmax+\thres\right) +\mu\tau} \big(\betag -\frac{\betav}{4} \big)$  in \Adam. Here  $c_1$ $\sim$ $c_3$ are positive constants. For all $\rhoa\in(0,\rho_0)$, $p\in(0,p_0]$,   $0<\epsi{}\leq \epsi{}_0$ with $\epsi{}_0 = \epsi{}_0(\rhoa)$, and $\wmi{0}= \wgi{0}$, we have 
	\begin{equation}\label{small_distance}
	\sup_{\wmi{0}\in\G} \PPi{\sup_{0\leq t < \jumptime{1}} \|\wmi{t}- \wgi{t}\|\geq 2\epsi{\rhoa}} \leq 2\exp(-\epsi{-p/2}),
	\end{equation}
	where  the sequences $\wmi{t}$ and $\wgi{t}$  are respectively produced by~Eqn.~\eqref{assumption_stochastic_adam} and~\eqref{deterministicversion_adam} in Adam or Eqn.~\eqref{assumption_stochastic_RMSP} and~\eqref{deterministicversion_RMSP} in \RMSProp~and \Sgd.  
\end{lem}

See its proof in Appendix~\ref{proofoflemm4}.

\section{Proof of Results in Sec.~\ref{volumeanalysis}}\label{proofofvolumeanalysis}
\subsection{Proof of Theorem~\ref{upperlowerbound}}\label{proofofupperlowerbound}
\begin{proof}
	Here we first briefly introduce our proof idea.   As we proved in Lemma~\ref{lemm4},
	for any $\delta \in (0,1)$, there exist $\rho_0$, $p_0$ and $\epsi{}_0$ such that for all $\rhoa\in(0,\rho_0)$, $p\in(0,p_0]$ and $0<\epsi{}\leq \epsi{}_0$,  we have 
	\begin{equation}\label{small_distancea}
	\sup_{\wmi{0}\in\G} \PPi{\sup_{0\leq t < \jumptime{1}} \|\wmi{t}- \wgi{t}\|\geq 2\epsi{\rhoa}} \leq 2\exp(-\epsi{-p/2}),
	\end{equation}
	where the sequences $\wmi{t}$ and $\wgi{t}$ share the same initialization $\wmi{0}= \wgi{0}$. Such a result holds for both \Sgd~and Adam. Besides, from Theorems~\ref{linearconvergenceSGD} and~\ref{linearconvergenceadam}, we know that the sequence $\{\wgi{t}\}$ produced by Eqn.~\eqref{deterministicversion_RMSP} or \eqref{deterministicversion_adam} (namely, the dynamic systems of \Sgd~and Adam) exponentially converges to the minimum $\wms$ of the current local basin $\G$. To escape the local basin $\G$, there are two possible choices, the small jumps in the process $\xii{}$ and the big jumps $\jumpi{k}$ in the process $\zetai{}$. As the small jumps in the process $\xii{}$ is well bounded, it is not very likely that these small jumps can help escape the local basin $\G$ which is verified by Eqn.~\eqref{small_distancea}.  We well prove this more rigorously latter. For the big jumps $\jumpi{}$,  since the expectation jump time $\EE(\jumptime{1})$ is $1/\psiis{\epsi{}}{\delta}$, such as $\EE(\jumptime{1})=\frac{2}{\alpha} \epsi{\alpha\delta}$ in the $\alpha$-stable ($\Sas$) distribution,  $\EE(\jumptime{1})$  is usually much larger than the necessary time $t=\mathcal{O}(\ln(1/\epsi{}))$ to achieve $\|\wgi{t}-\wms\|\leq \epsi{\delta}$. This means that before the jump time $\jumptime{1}$ the sequence $\wgi{t}$ is very close to the optimum of $\G$ and thus $\wmi{t}$ is very close to the minimum $\wms$. In this way, the escaping time $\te$ of the sequence $\{\wmi{t}\}$ most likely occurs at the time $\jumptime{1}$ if the big jump $\epsi{}\jumpi{1}$ in the process $\zetai{}$ is large. If the jump $\epsi{}\jumpi{1}$ is small and $\wmi{\jumptime{1}}$ does not escape $\G$, then $\wmi{t}$ will converge to the minimum $\wms$ exponentially and stay in the small neighborhood of $\wms$. Accordingly, before the second jump time $t_2=t_1+\jumptime{2}$, $\wmi{t_2}$ will jump. This process will continue during the time interval $[0, t]$.  Since for each jump time $t_k-$, $\wmi{t_k-}$ is very close to the optimum $\wms$, the big jump size $\epsi{} \Qmi{t_k}^{-1} \Sigmai{t_k} \jumpi{t_k} \approx  \epsi{} \Qmi{\wms}^{-1} \Sigmai{\wms} \jumpi{t_k}$. So we can use $\epsi{} \Qmi{t}^{-1} \Sigmai{t} \jumpi{t_k} \approx  \epsi{} \Qmi{\wms}^{-1} \Sigmai{\wms} \jumpi{t_k}\notin \G$ to judge whether at time $t_k$, $\wmi{t_k}$ escapes the local basin $\G$. The events $\{ \epsi{} \jumpi{1} \notin \W\}=\{\epsi{} \Qmi{\wms}^{-1} \Sigmai{\wms} \jumpi{t_k}\notin \G\}$, $\cdots$,  $\{ \epsi{} \jumpi{k-1} \notin \W\}=\{\epsi{} \Qmi{\wms}^{-1} \Sigmai{\wms} \jumpi{t_{k-}}\notin \G\}$, $\{ \epsi{} \jumpi{k} \notin \W\}=\{\epsi{} \Qmi{\wms}^{-1} \Sigmai{\wms} \jumpi{t_k}\notin \G\}$ are independent. 
	
	Now we prove the desired results from two aspects, namely establishing upper bound and lower bound of $\EEi{\exp\left(-um(\W) \Hui{\epsi{-1}}\te \right)}$ for any $u>-1$. Before that, we first establish basic inequalities for lower and upper bounds.
	
	\step{Basic inequalities for lower and upper bounds.}	 Since  $\jumptime{1}$ is exponentially distributed with the parameter $\psiis{\epsilon}{\delta}$, we  compute the Laplace transform of $m(\W)\Hui{\epsi{-1}}\jumptime{1}$ as follows:
	\begin{equation*}
	\begin{split}
	&\EE\left[e^{- u m(\W)\Hui{\epsi{-1}}\jumptime{1}}\right] = \EE\left[\int_{0}^{+\infty}e^{- u m(\W)\Hui{\epsi{-1}}\jumptime{1}} \cdot \psiis{\epsi{}}{\delta} e^{-\psiis{\epsi{}}{\delta}\jumptime{1}}\ds \jumptime{1}\right] \\
	= & \frac{\psiis{\epsi{}}{\delta}}{\psiis{\epsi{}}{\delta}+ u m(\W)\Hui{\epsi{-1}}} = \frac{1}{1+ u\aae},
	\end{split}
	\end{equation*}
	where $\aae=m(\W)\frac{\Hui{\epsi{-1}}}{\Hui{-\epsi{\delta}}}\ \text{and}\ \psiis{\epsi{}}{\delta}= \Hui{-\epsi{\delta}}.$
	Besides, for the probability law of the big jump we have
	\begin{equation*}
	\PPi{\Qmi{\wms}^{-1} \Sigmai{\wms}\epsi{}\jumpi{1}\notin\G} =\PPi{\epsi{}\jumpi{1}\in\W}=  \frac{\nu(\W/\epsi{})}{\psiis{\epsi{}}{\delta}}.
	\end{equation*}
	Since for the \levyp~measure, we have $
	m(\W) = \lim_{u \rightarrow +\infty} \frac{\nu(u\W)}{\Hui{u}}$ according to~\cite{pavlyukevich2011first}. 
	So for any $\delta'$, there always exists $\epsi{}$ such that it holds  
	\begin{equation}\label{afdsafascascsac}
	\aae(1-\delta')\leq \frac{\nu(\W/\epsi{})}{\psiis{\epsi{}}{\delta}} =  \frac{\nu(\W/\epsi{})}{\Hui{\epsi{-1}}} \frac{\Hui{\epsi{-1}}}{\psiis{\epsi{}}{\delta}}\overset{\text{\ding{172}}}{\approx} m(\W) \frac{\Hui{\epsi{-1}}}{\psiis{\epsi{}}{\delta}} = m(\W) \frac{\Hui{\epsi{-1}}}{\Hui{\epsi{-\delta}}} \leq \aae(1+\delta').
	\end{equation}
	where \ding{172} holds since $\epsi{}$ is enough small. 	Then with the help of the continuity of the function $(\wm,\zm)\rightarrow \Qmi{\wm}^{-1} \Sigmai{\wm}\zm$ both in $\wm$ and $\zm$. Indeed, for any $\delta'$ we can choose $R>0$ enough large such that for small $\epsi{}$ we have 
	\begin{equation*}
	\PPi{\|\epsi{}\jumpi{1}\|>R} \leq \frac{\delta'}{4} \frac{\Hui{\epsi{-1}}}{\Hui{\epsi{-\delta}}}.
	\end{equation*}
	Further, the function $(\wm,\zm)\rightarrow \Qmi{\wm}^{-1} \Sigmai{\wm}\zm$ is uniformly continuous in $\zm$ in the ball $\|\zm\|\leq R$ and is continuous in $\wm$ at the optimum $\wms$. Following~\cite{pavlyukevich2011first}, by using the scaling property of the jump measure $\nu$ and the fact that the limiting measure $m$ has no atoms we show that uniformly over $\|\wm-\wms\|\leq \epsi{\gamma}$:
	\begin{equation}\label{AAAasfasf}
	\begin{cases}
	\left|\PPi{\Qmi{\wm}^{-1} \Sigmai{\wm}\epsi{}\jumpi{k}\notin\G^{\pm\epsi{\gamma}} , \  \|\epsi{}\jumpi{k}\|\leq R} - \PPi{\Qmi{\wms}^{-1} \Sigmai{\wms}\epsi{}\jumpi{k}\notin\G,\ \|\epsi{}\jumpi{k}\|\leq R} \right|\leq \frac{\delta'}{4} \frac{\Hui{\epsi{-1}}}{\Hui{\epsi{-\delta}}},\\
	\left|\PPi{\Qmi{\wm}^{-1} \Sigmai{\wm}\epsi{}\jumpi{k}\notin\G, \  \|\epsi{}\jumpi{k}\|\leq R} - \PPi{\Qmi{\wms}^{-1} \Sigmai{\wms}\epsi{}\jumpi{k}\notin\G,\ \|\epsi{}\jumpi{k}\|\leq R} \right|\leq \frac{\delta'}{4} \frac{\Hui{\epsi{-1}}}{\Hui{\epsi{-\delta}}},\\
	\end{cases}
	\end{equation}
	At the same time, we also can establish 
	\begin{equation}  \label{ascasdcascas}
	\begin{split}
	& \PPi{\Qmi{\wms}^{-1} \Sigmai{\wms}\epsi{}\jumpi{k}\notin\G} -  \PPi{\Qmi{\wms}^{-1} \Sigmai{\wms}\epsi{}\jumpi{k}\notin\G,\ \|\epsi{}\jumpi{k}\|\leq R} \\
	=&   \PPi{\Qmi{\wms}^{-1} \Sigmai{\wms}\epsi{}\jumpi{k}\notin\G} -  \PPi{\|\epsi{}\jumpi{k}\|\leq R \ | \  \Qmi{\wms}^{-1} \Sigmai{\wms}\epsi{}\jumpi{k}\notin\G} \PPi{\Qmi{\wms}^{-1} \Sigmai{\wms}\epsi{}\jumpi{k}\notin\G}  \\
	=&  \PPi{\Qmi{\wms}^{-1} \Sigmai{\wms}\epsi{}\jumpi{k}\notin\G} (1-   \PPi{\|\epsi{}\jumpi{k}\|\leq R \ | \  \Qmi{\wms}^{-1} \Sigmai{\wms}\epsi{}\jumpi{k}\notin\G} ) \\
	=&  \PPi{\Qmi{\wms}^{-1} \Sigmai{\wms}\epsi{}\jumpi{k}\notin\G}  \PPi{\|\epsi{}\jumpi{k}\|> R \ | \  \Qmi{\wms}^{-1} \Sigmai{\wms}\epsi{}\jumpi{k}\notin\G}  
	\leq  \PPi{\|\epsi{}\jumpi{k}\|> R}  \leq \frac{\delta'}{4} \frac{\Hui{\epsi{-1}}}{\Hui{\epsi{-\delta}}}.
	\end{split}
	\end{equation}

	\textbf{Upper bound of $\EEi{\exp\left(-um(\W) \Hui{\epsi{-1}}\te \right)}$.}  
	In this part, we  consider both the big jumps in the process $\zetai{}$ and the small jumps in the process $\xii{}$ which may escape the local minimum $\wms$.  Instead of estimate the escaping time $\te$ from $\G$, we  first estimate the escaping time $\tea$ from $\G^{-\rhoa}$.  Here we define  the inner part of $\G$ as $\G^{-\rhoa}=\{\ym \in \G\ |\ \dis(\partial \G, \ym) \geq \rho\}$  and the outer $\rho$-neighborhood of $\G$ as $\G^{+\rhoa} = \{\ym\ |\ \dis(\partial \G, \ym)\geq \rhoa\}$. Then by setting $\rhoa\downarrow 0$, we can use $\tea$ to estimate $\te$ well. Let   $\rhoa=\epsi{\gamma}$ where $\gamma$ is a constant such that the results  of Lemmas~\ref{lemma2}$\sim$~\ref{lemm4} holds. Here we suppose the initial point $\wmi{0}\in \GG$.

	\step{Step 1.} In this step we give the formulation of the upper bound of $\EEi{e^{-um(\W) \Hui{\epsi{-1}}\te}}$. For any $u>-1$, we can compute the formula of the total probability as follows	
	\begin{equation*}
	\EEi{e^{-um(\W) \Hui{\epsi{-1}}\tea}} \leq \sum_{k=1}^{+\infty} \EE \left[ e^{-um(\W) \Hui{\epsi{-1}}t_k} \indi{\tea=t_k}  + \Resi{k}\right],
	\end{equation*}
	where 
	\begin{equation*}
	\Resi{k}\leq 
	\begin{cases}
	\EEi{e^{-um(\W) \Hui{\epsi{-1}}t_k}\indi{\tea\in(t_{k-1}, t_k)}},\quad &\text{if}\ u\in(-1, 0]\\
	\EEi{e^{-um(\W) \Hui{\epsi{-1}}t_{k-1}}\indi{\tea\in(t_{k-1}, t_k)}},\quad &\text{if}\ u\in(0, +\infty).
	\end{cases}
	\end{equation*}
	
	\step{Step 2.} In this step we specifically upper bounds the first term $\sum_{k=1}^{+\infty} \EE \left[ e^{-um(\W) \Hui{\epsi{-1}}t_k} \indi{\tea=t_k} \right]$. 
	For $k\geq 1$, we can use the strong Markov property and obtain
	\begin{equation*} 
	\begin{split}
	& \EE \left[e^{-um(\W) \Hui{\epsi{-1}}t_k} \indi{\tea=t_k}\right] = \EE \left[e^{-um(\W) \Hui{\epsi{-1}}t_k} \indi{\wmi{t}\in\G^{-\epsi{\gamma}}, t\in[0,t_k), \wmi{t_k}\notin \G^{-\epsi{\gamma}}} \right]\\ 
	= & \EE \left[  e^{-um(\W) \Hui{\epsi{-1}}\jumptime{k}} \indi{\wmi{t+t_{k-1}}\in\G^{-\epsi{\gamma}}, t\in[0,\jumptime{k})}\indi{ \wmi{t_k}\notin \G^{-\epsi{\gamma}}} \right. \\
	&\qquad \qquad \qquad \qquad \qquad \qquad \qquad \qquad\cdot \left.\prod_{i=1}^{k-1} e^{-um(\W) \Hui{\epsi{-1}}\jumptime{i}}\indi{\wmi{t+t_{i-1}}\in\G^{-\epsi{\gamma}}, t\in[0,\jumptime{k}]} \right]\\ 
	\leq  & \sup_{\wmi{0}\in\GG}\!\!\!\!\!\!\! \EE \left[ e^{-um(\W) \Hui{\epsi{-1}}\jumptime{1}}\indi{\wmi{t}\!\in\!\G^{-\epsi{\gamma}}\!, t\!\in\![0,\jumptime{1})} \indi{ \wmi{\jumptime{1}}\notin \G^{-\epsi{\gamma}}} \right]\\
	&\qquad \qquad \qquad \qquad \qquad \qquad \qquad \qquad \cdot \sup_{\wmi{0}\in\GG} \!\!\!\!\!\!\! \EE \left[  e^{-um(\W) \Hui{\epsi{-1}}\jumptime{1}}\indi{\wmi{t}\!\in\!\G^{-\epsi{\gamma}}\!, t\in[0,\jumptime{1}]} \right]^{k-1}\!\!\!.
	\end{split} 
	\end{equation*}

	Recall  $\rhoa=\epsi{\gamma}$ where $\gamma$ is a constant such that the results  of Lemmas~\ref{lemma2}$\sim$~\ref{lemm4} holds. The escaping  from the basin $\G^{-\epsi{\gamma}}$ with a big jump $\epsi{}\jumpi{1}$ occurs when $\Qmi{\jumptime{1}-}^{-1} \Sigmai{\jumptime{1}-}\epsi{}\jumpi{1}\in\G^{-\epsi{\gamma}}$. Furthermore, $\sup_{0\leq t<\jumptime{1}} \| \wmi{t}-\wgi{t}\| \leq \frac{1}{2} \epsi{\gamma}$ with probability exponentially close to 1 (verified by Lemma~\ref{lemm4}). Meanwhile  $\jumptime{1}=\frac{2}{\alpha} \epsi{\alpha\delta}$ in the $\alpha$-stable ($\Sas$) distribution is much larger than $\ve=\mathcal{O}(\ln(1/\epsi{}))$ with sufficient small $\epsi{}$, $\wgi{t}$ reaches a $\frac{1}{2}\epsi{\gamma}$-neighborhood of the optimum $\wms$ which only requires  time $\ve$. So this actually means $\sup_{0\leq t<\jumptime{1}} \| \wmi{t}-\wms\| \leq  \epsi{\gamma}$.  In this way, to obtain the final upper bound results, we only need to estimate the escaping probability $\PPi{\Qmi{\wm}^{-1} \Sigmai{\wm}\epsi{}\jumpi{1}\in\G^{-\epsi{\gamma}}}$ and 
	$\PPi{\Qmi{\wm}^{-1} \Sigmai{\wm}\epsi{}\jumpi{1}\notin\G^{-\epsi{\gamma}}}$ uniformly over $\|\wm-\wms\|\leq \epsi{\gamma}$. 
	Then we first give two important inequalities which will used to bound each component  later:
	\begin{equation*}
	\begin{split}
	&\sup_{\|\wm-\wms\|\leq \epsi{\gamma}} \!\!\!\!\PPi{\Qmi{\wm}^{-1} \Sigmai{\wm}\epsi{}\jumpi{k}\notin\G^{-\epsi{\gamma}} }\\ 
	=& \!\! \sup_{\|\wm-\wms\|\leq \epsi{\gamma}} \!\!\!\! \PPi{\Qmi{\wm}^{-1} \Sigmai{\wm}\epsi{}\jumpi{k}\notin\G^{-\epsi{\gamma}}\!,  \|\epsi{}\jumpi{k}\|\!\leq\! R} \!+\! \PPi{\Qmi{\wm}^{-1} \Sigmai{\wm}\epsi{}\jumpi{k}\!\notin\!\G^{-\epsi{\gamma}}\!, \|\epsi{}\jumpi{k}\|\! >\! R}\\
	\ged{172} & \PPi{\Qmi{\wms}^{-1} \Sigmai{\wms}\epsi{}\jumpi{k}\notin\G\!, \|\epsi{}\jumpi{k}\|\!\leq \!R} \! -\!   \frac{\delta'}{4} \frac{\Hui{\epsi{-1}}}{\Hui{\epsi{-\delta}}} \!+\! \PPi{\Qmi{\wm}^{-1} \Sigmai{\wm}\epsi{}\jumpi{k}\notin\G^{-\epsi{\gamma}},\ \|\epsi{}\jumpi{k}\| \!> \!R} \\
	\geq  & \PPi{\Qmi{\wms}^{-1} \Sigmai{\wms}\epsi{}\jumpi{k}\notin\G,\ \|\epsi{}\jumpi{k}\|\leq R}  -   \frac{\delta'}{4} \frac{\Hui{\epsi{-1}}}{\Hui{\epsi{-\delta}}}  \\
	\ged{173} & \PPi{\Qmi{\wms}^{-1} \Sigmai{\wms}\epsi{}\jumpi{k}\notin\G}  -  \frac{\delta'}{2} \frac{\Hui{\epsi{-1}}}{\Hui{\epsi{-\delta}}} \\
	\ged{174} & m(\W)  \left(1-\delta' - \frac{\delta'}{2 m(\W)}\right) \frac{\Hui{\epsi{-1}}}{\Hui{\epsi{-\delta}}} \ged{175} m(\W)  (1-\rho) \frac{\Hui{\epsi{-1}}}{\Hui{\epsi{-\delta}}},
	\end{split}
	\end{equation*}
	where \ding{172} uses the  result in Eqn.~\eqref{AAAasfasf}, \ding{173} uses Eqn.~\eqref{ascasdcascas}, \ding{174} uses  Eqn.~\eqref{afdsafascascsac}, and in \ding{175} we set $\delta'$ enough small such that $\rho\geq \delta' + \frac{\delta'}{2 m(\W)}$. So in this way, for any $\rho$ we choose $\delta'>0$ small enough to lower bound $\sup_{\|\wm-\wms\|\leq \epsi{-\gamma}} \PPi{\Qmi{\wm}^{-1} \Sigmai{\wm}\epsi{}\jumpi{k}\in\G^{-\epsi{\gamma}} } $ as follows:
	\begin{equation*}
	\sup_{\|\wm-\wms\|\leq \epsi{\gamma}} \PPi{\Qmi{\wm}^{-1} \Sigmai{\wm}\epsi{}\jumpi{k}\in\G^{-\epsi{\gamma}} } =1- \sup_{\|\wm-\wms\|\leq \epsi{\gamma}} \PPi{\Qmi{\wm}^{-1} \Sigmai{\wm}\epsi{}\jumpi{k}\notin\G^{-\epsi{\gamma}} }  \geq  1-\aae(1-\rho).
	\end{equation*}
	Similarly, we only need to upper bound the remaining term  $\sup_{\|\wm-\wms\|\leq \epsi{\gamma}} \PPi{\Qmi{\wm}^{-1} \Sigmai{\wm}\epsi{}\jumpi{k}\notin\G }$ as follows:
	\begin{equation*}
	\begin{split}
	&\sup_{\|\wm-\wms\|\leq \epsi{\gamma}} \!\!\!\!  \PPi{\Qmi{\wm}^{-1} \Sigmai{\wm}\epsi{}\jumpi{k}\notin\G^{-\epsi{\gamma}} }\\
	=& \!\!\sup_{\|\wm-\wms\|\leq \epsi{\gamma}}\!\!\!\! \PPi{\Qmi{\wm}^{-1} \Sigmai{\wm}\epsi{}\jumpi{k}\notin\G^{-\epsi{\gamma}}\!,   \|\epsi{}\jumpi{k}\|\!\leq\! R}\!+\! \PPi{\Qmi{\wm}^{-1} \Sigmai{\wm}\epsi{}\jumpi{k}\!\notin\!\G^{-\epsi{\gamma}}\!,   \|\epsi{}\jumpi{k}\|\!>\! R} \\
	\led{172} & \!\!\sup_{\|\wm-\wms\|\leq \epsi{\gamma}}\!\!\!\! \PPi{\Qmi{\wm}^{-1} \Sigmai{\wm}\epsi{}\jumpi{k}\notin\G^{-\epsi{\gamma}}\!,   \|\epsi{}\jumpi{k}\|\!\leq\! R}\!+\!  \frac{\delta'}{4} \frac{\Hui{\epsi{-1}}}{\Hui{\epsi{-\delta}}}  \\
	\led{173} & \!\!\sup_{\|\wm-\wms\|\leq \epsi{\gamma}}\!\!\!\! \PPi{\Qmi{\wms}^{-1} \Sigmai{\wms}\epsi{}\jumpi{k}\notin\G,\ \|\epsi{}\jumpi{k}\|\leq R} + \frac{\delta'}{2} \frac{\Hui{\epsi{-1}}}{\Hui{\epsi{-\delta}}} \\
	\leq &\!\!\sup_{\|\wm-\wms\|\leq \epsi{\gamma}}\!\!\!\!    \PPi{\Qmi{\wms}^{-1} \Sigmai{\wms}\epsi{}\jumpi{k}\notin\G} + \frac{\delta'}{2} \frac{\Hui{\epsi{-1}}}{\Hui{\epsi{-\delta}}} \\
	\led{174} & m(\W)  \left(1+\delta' + \frac{\delta'}{2m(\W)}\right)  \frac{\Hui{\epsi{-1}}}{\Hui{\epsi{-\delta}}} \leq m(\W)  (1+\rho/3) \frac{\Hui{\epsi{-1}}}{\epsi{-\delta}}= \aae(1+\rho/3),
	\end{split}
	\end{equation*}
	where \ding{172} uses $\PPi{\Qmi{\wm}^{-1} \Sigmai{\wm}\epsi{}\jumpi{k}\!\notin\!\G^{-\epsi{\gamma}}\!,   \|\epsi{}\jumpi{k}\|\!>\! R}\leq \PPi{\|\epsi{}\jumpi{k}\|\!>\! R}\leq  \frac{\delta'}{4} \frac{\Hui{\epsi{-1}}}{\Hui{\epsi{-\delta}}} $, \ding{173} uses the  result in Eqn.~\eqref{AAAasfasf}, \ding{173} uses  Eqn.~\eqref{ascasdcascas}, and \ding{174} uses Eqn.~\eqref{afdsafascascsac}.
	
	Next, for any $\rho>0$ and $\epsi{}$ we can obtain the Laplace transforms for any $u>-1$ as follows:
	\begin{equation}\label{uppboundpro}
	\begin{split}
	& \sup_{\wmi{0}\in \GG}\!\!\!\!\! \EE\left[e^{-um(\W)\Hui{\epsi{-1}}\jumptime{1}} \indi{\wmi{t}\!\in\!\G^{-\epsi{\gamma}}\!,\ t\!\in\![0,\jumptime{1}]}\right] \\
	\!\leq & [1-\aae(1+\rho)] \EE\left[\int_{0}^{+\infty}\!\!\!\!\!e^{- u m(\W)\Hui{\epsi{-1}}\jumptime{1}} \cdot \psiis{\epsi{}}{\delta} e^{-\psiis{\epsi{}}{\delta}\jumptime{1}}\ds \jumptime{1}\right] \\
	= &  \frac{1-\aae(1-\rho)}{1+u\aae}.
	\end{split}
	\end{equation}
	and 
	\begin{equation*}
	\begin{split}
	\sup_{\wmi{0}\in \GG} \!\! &\EE\left[e^{-um(\W)\Hui{\epsi{-1}}\jumptime{1}} \indi{\wmi{t}\in\G^{-\epsi{\gamma}}\!, t\in[0,\jumptime{1})}  \indi{\wmi{\jumptime{1}}\notin\G^{-\epsi{\gamma}}}\right] \\
	&\qquad  \leq  \aae\left(1+\frac{\rho}{3}\right) \EE\left[\int_{0}^{+\infty}\!\! e^{- u m(\W)\Hui{\epsi{-1}}\jumptime{1}} \cdot \psiis{\epsi{}}{\delta} e^{-\psiis{\epsi{}}{\delta}\jumptime{1}}\ds \jumptime{1}\right]  
	=   \frac{\aae(1-\rho/3)}{1+u\aae}.
	\end{split}
	\end{equation*}
	Here we summarize the above results such that we can upper  bound the first term  $\sum_{k=1}^{+\infty} \EE \left[e^{-um(\W)\Hui{\epsi{-1}}t_k}\indi{\te=t_k} \right]$:
	\begin{equation*}
	\begin{split}
	\mathcal{R}_1= \sum_{k=1}^{+\infty} \EE \left[e^{-um(\W)\Hui{\epsi{-1}}t_k}\indi{\te=t_k} \right]
	\leq & \frac{\aae(1+\rho/3)}{1+u\aae} \sum_{k=1}^{+\infty}\left(\frac{1-\aae(1-\rho)}{1+u\aae}\right)^{k-1} \\
	\leq & \frac{\aae(1+\rho/3)}{1+u\aae} \sum_{k=0}^{+\infty}\left(\frac{1-\aae(1-\rho)}{1+u\aae}\right)^{k-1}  
	= \frac{1+\rho/3}{1+u -\rho}.
	\end{split}
	\end{equation*}
	
	\step{Step 3.} In this step we specifically upper bounds the second term $\sum_{k=1}^{+\infty} \EE \left[ \Resi{k}\right]$. Specifically, we establish upper bound for each $\EE \left[ \Resi{k}\right]$ as follows. We first consider the case where $k=1$:
	\begin{equation*}
	\begin{split}
	\Resi{1}\leq &
	\begin{cases}
	\EEi{e^{-um(\W) \Hui{\epsi{-1}})t_1}\indi{\te\in(0, t_1)}},\qquad\qquad \qquad\qquad \qquad\qquad \quad \ \ \ &\text{if}\ u\in(-1, 0]\\
	\EEi{\indi{\te\in(0, t_1)}},\quad &\text{if}\ u\in(0, +\infty).
	\end{cases}\\
	= &
	\begin{cases}
	\EEi{e^{-um(\W) \Hui{\epsi{-1}})\jumptime{1}}\indi{ \exists t \in(0, \jumptime{1}):\ \wmi{t} \notin\G^{-\epsi{\gamma}}}},\quad\qquad\qquad\qquad\quad &\text{if}\ u\in(-1, 0]\\
	\EEi{\indi{ \exists t \in(0, \jumptime{1}):\ \wmi{t} \notin\G^{-\epsi{\gamma}}}},\quad &\text{if}\ u\in(0, +\infty).
	\end{cases}\\
	\leq &
	\begin{cases}
	\EEi{e^{-um(\W) \Hui{\epsi{-1}})\jumptime{1}}\sup_{\wmi{0}\in\GG}\indi{ \exists t \in(0, \jumptime{1}):\ \wmi{t} \notin\G^{-\epsi{\gamma}}}},\quad \qquad\qquad&\text{if}\ u\in(-1, 0]\\
	\EEi{\sup_{\wmi{0}\in\GG} \indi{ \exists t \in(0, \jumptime{1}):\ \wmi{t} \notin\G^{-\epsi{\gamma}}}},\quad\qquad &\text{if}\ u\in(0, +\infty).
	\end{cases}\\
	\end{split}
	\end{equation*}
	For $k\geq 2$, it needs more efforts to be upper bounded:
	\begin{equation*}
	\begin{split}
	&\Resi{k}\\\leq &
	\begin{cases}
	\EEi{e^{-um(\W) \Hui{\epsi{-1}})t_k}\indi{\te\in(t_{k-1}, t_k)}},\ \qquad \qquad\qquad\qquad\qquad\qquad \quad\qquad\qquad\qquad\qquad &\text{if}\ u\in(-1, 0]\\
	\EEi{e^{-um(\W) \Hui{\epsi{-1}})t_{k-1}}\indi{\te\in(t_{k-1}, t_k)}},\quad &\text{if}\ u\in(0, +\infty).
	\end{cases}\\
	= &
	\begin{cases}
	\EEi{e^{-um(\W) \Hui{\epsi{-1}})t_k}\indi{t \in[0, t_{k-1}]:\ \wmi{t}\in\G^{-\epsi{\gamma}}} \indi{\exists t \in(t_{k-1}, t_{k}):\ \wmi{t}\notin\G^{-\epsi{\gamma}}}} ,\quad &\text{if}\ u\in(-1, 0]\\
	\EEi{e^{-um(\W) \Hui{\epsi{-1}})t_{k-1}}\indi{t \in[0, t_{k-1}]:\ \wmi{t}\in\G^{-\epsi{\gamma}}} \indi{\exists t \in(t_{k-1}, t_{k}):\ \wmi{t}\notin\G^{-\epsi{\gamma}}}},\quad &\text{if}\ u\in(0, +\infty).
	\end{cases}
	\end{split}
	\end{equation*}
	In this case, for all $u>0$ we can upper 	bound $\Resi{k} $ as 
	\begin{equation*}
	\begin{split}
	\Resi{k}  \leq & \left[\EEi{\!x \sup_{\wmi{0}\in \GG} \!\!\!\!\indi{t \!\in\! [0, \jumptime{1}]: \wmi{t}\!\in\!\G^{-\epsi{\gamma}}}} \right]^{k-2} \!\! \!\!\\
	&\qquad  \qquad  \qquad  \EEi{\!x \sup_{\wmi{0}\in \GG}\!\!\!\! \indi{t \!\in\![0, \jumptime{1}]: \wmi{t}\!\in\!\G^{-\epsi{\gamma}}}  \indi{\exists t \in(0, \jumptime{1}): \wmi{t}\!\notin\!\G^{-\epsi{\gamma}}}}.
	\end{split}
	\end{equation*}
	where $x=e^{-um(\W) \Hui{\epsi{-1}})\jumptime{1}}$.  Let the event $E= \{\sup_{0\leq t < \jumptime{1}} \|\wmi{t}- \wgi{t}\|\leq \epsi{\gamma}\}$. Now we bound each term in the above inequalities:
	\begin{equation}\label{safasfcsadcfsa}
	\begin{split}
	&\EEi{e^{-um(\W) \Hui{\epsi{-1}})\jumptime{1}}\sup_{\wmi{0}\in\GG} \indi{ \exists t \!\in\!(0, \jumptime{1})\!: \wmi{t} \!\notin\!\Grd}}\\
	= &  \EEi{e^{-um(\W) \Hui{\epsi{-1}})\jumptime{1}}\sup_{\wmi{0}\in\GG}\indi{ \exists t \in(0, \jumptime{1}):\ \wmi{t} \notin\Grd}(\indi{E}+\indi{E^c})}\\
	\leq &  \EEi{e^{-um(\W) \Hui{\epsi{-1}})\jumptime{1}}\sup_{\wmi{0}\in\GG}\indi{ \exists t \in(0, \jumptime{1}):\ \wmi{t} \notin\Grd} \indi{E^c}}\\
	\led{172} &  \EEi{e^{-um(\W) \Hui{\epsi{-1}})\jumptime{1}} \exp(-\epsi{-p})} = \frac{\psiis{\epsi{}}{\delta}}{\psiis{\epsi{}}{\delta}+u m(\W) \Hui{\epsi{-1}}}\cdot 2\exp(-\epsi{-p})\\
	=&   \frac{1}{1+u\aae } \exp(-\epsi{-p})   \led{173}  \frac{\rho/3}{1+u-\rho},
	\end{split}
	\end{equation}
	where   \ding{172} uses the fact that $ \sup_{\wmi{0}\in\GG}\indi{ \exists t \in(0, \jumptime{1}):\ \wmi{t} \notin\G} \leq 1$ and the sequence $\wgi{t}$ obeys $\GG$ due to $\wmi{0}\in\GG$ and the results in Lemma~\ref{lemm4}:
	\begin{equation*}
	\sup_{\wmi{0}\in\G} \PPi{\sup_{0\leq t < \jumptime{1}} \|\wmi{t}- \wgi{t}\|\geq \epsi{\gamma}} \leq 2 \exp(-\epsi{-p}),
	\end{equation*}
	where the sequences $\wmi{t}$ and $\wgi{t}$ share the same initialization $\wmi{0}= \wgi{0}$. In \ding{173} we set $\epsi{}$ small enough such that $2\exp(-\epsi{-p})\leq    \frac{\rho/3}{1+u-\rho}$. Similarly, we can upper bound 
	\begin{equation}\label{afcsvsdrtwafszcv}
	\begin{split}
	\EEi{\sup_{\wmi{0}\in\GG}\!\!\!\!\indi{ \exists t \!\in\!(0, t_1): \wmi{t} \!\notin\!\Grd}}\!\leq & \EEi{\sup_{\wmi{0}\in\GG}\!\!\!\!\indi{ \exists t \!\in\!(0, t_1): \wmi{t} \notin\Grd}(\indi{E}\!+\!\indi{E^c})}\\
	\leq&  \exp(-\epsi{-p})\!\leq\!  \frac{\rho/3}{1+u-\rho}.
	\end{split}
	\end{equation}
	Since $p$ is much smaller than 1, then we have for $k=2,\cdots, k$
	\begin{equation*}
	\begin{split}
	&	\Resi{k}\leq  
	\left[\EEi{e^{-um(\W) \Hui{\epsilon^{-1}})\jumptime{1}}\sup_{\wmi{0}\in \GG} \indi{t \in[0, \jumptime{1}]:\ \wmi{t}\in\Grd}} \right]^{k-2} 
	\EE \Bigg[ e^{-um(\W) \Hui{\epsilon^{-1}})\jumptime{1}} \\
	&  \sup_{\wmi{0}\in \GG} \indi{t \in[0, \jumptime{1}]:\ \wmi{t}\in\Grd}   \indi{\exists t \in(0, \jumptime{1}):\ \wmi{t}\notin\Grd}\Bigg]  
	\leq  \left[\frac{1-\aae(1-\rho)}{1+u\aae}\right]^{k-2} \frac{\aae(1+\rho/3)}{1+u\aae}.
	\end{split}
	\end{equation*}
	where we use the above results, namely, $\sup_{\wmi{0}\in \GG} \EE\left[e^{-um(\W)\Hui{\epsi{-1}}\jumptime{1}} \indi{\wmi{t}\in\Grd,\ t\in[0,\jumptime{1}]}\right] 
	\leq   \frac{1-\aae(1-\rho)}{1+u\aae}$ 
	and $\sup_{\wmi{0}\in \GG} \EE\left[e^{-um(\W)\Hui{\epsi{-1}}\jumptime{1}} \indi{\wmi{t}\in\Grd,\ t\in[0,\jumptime{1})}  \indi{\wmi{\jumptime{1}}\notin\Grd}\right] 
	\leq  \frac{\aae(1+\rho/3)}{1+u\aae}$.  So in this case, we have 
	\begin{equation*}
	\begin{split}
	\mathcal{R}_2=& \sum_{k=1}^{+\infty} \EE \left[\Resi{k}\right]
	\leq  \frac{\rho/3}{1+u-\rho} + \sum_{k=2}^{+\infty} \left[\frac{1-\aae(1-\rho)}{1+u\aae}\right]^{k-2} \frac{\aae(1+\rho/3)}{1+u\aae} = \frac{1+2\rho/3}{1+u-\rho}\\
	\end{split}
	\end{equation*}
	Therefore, for any $\wmi{0}\in\GG$ we can upper bound 
	\begin{equation*}
	\EEi{e^{-um(\W) \Hui{\epsi{-1}})\te }} \leq \mathcal{R}_1 + \mathcal{R}_2 \leq \frac{1+\rho }{1+u-\rho},
	\end{equation*}
	where $\rho\downarrow 0$ as $\epsi{}\downarrow 0$.

	\textbf{Lower bound of $\EEi{\exp\left(-um(\W) \Hui{\epsi{-1}}\te \right)}$.} 
	In this part, we only consider the big jumps in the process $\zetai{}$ which may escape the local minimum $\wms$, and ignore the possibility of the small jumps in the process $\xii{}$ which may also help escape local minimum $\wms$. Here we consider the result under $\wmi{0}\in \Grd$ which is stronger than the results under $\wmi{0}\in\GG$ due to $\GG\subset \Grd$.
	
	\step{Step 1.} In this step we give the formulation of the lower bound of $\EEi{e^{-um(\W) \Hui{\epsi{-1}}\te}}$. 
	For any $u>-1$, we can compute the formula of the total probability as follows
	\begin{equation*}
	\EEi{e^{-um(\W) \Hui{\epsi{-1}}\te}} \geq \sum_{k=1}^{+\infty} \EE \left[ e^{-um(\W) \Hui{\epsi{-1}} t_k }\indi{\te=t_k} \right].
	\end{equation*}
	This inequality holds, since we ignore the small jumps in the process $\xii{}$ which may also help escape local minimum $\wms$. 
	
	For any small $\rhoa>0$, we define the inner part of $\G$ as $\G^{-\rhoa}=\{\ym \in \G\ |\ \dis(\partial \G, \ym) \geq \rho\}$  and the outer $\rho$-neighborhood of $\G$ as $\G^{+\rhoa} = \{\ym\ |\ \dis(\partial \G, \ym)\geq \rhoa\}$. For $k\geq 1$, we can use the strong Markov property and obtain
	\begin{equation}\label{adasfdasfa}
	\begin{split}
	& \EE \left[ e^{-um(\W) \Hui{\epsi{-1}} t_k }\indi{\te=t_k} \right] = \EE \left[e^{-um(\W) \Hui{\epsi{-1}} t_k } \indi{\wmi{t}\in\G, t\in[0,t_k), \wmi{t_k}\notin \G} \right]\\ 
	= & \EE \left[  e^{-um(\W) \Hui{\epsi{-1}} \jumptime{k}} \indi{\wmi{t+t_{k-1}}\in\G, t\in[0,\jumptime{k})}\indi{ \wmi{t_k}\notin \G} \right.\\
	&\qquad \qquad\qquad \qquad	\qquad \qquad  \qquad \qquad\left. \cdot \prod_{i=1}^{k-1} e^{-um(\W) \Hui{\epsi{-1}} \jumptime{i} } \indi{\wmi{t+t_{i-1}}\in\G, t\in[0,\jumptime{i}]} \right]\\ 
	\geq  &\! \inf_{\wmi{0}\in\G^{-\rhoa}} \!\! \EE \left[ e^{-um(\W) \Hui{\epsi{-1}}\jumptime{1}}\indi{\wmi{t}\in\G^{-\rhoa}, t\!\in\![0,\jumptime{1})} \indi{ \wmi{\jumptime{1}}\notin \G} \right]\! \\
	&\qquad \qquad \qquad \qquad \qquad \qquad\qquad \qquad \cdot \inf_{\wmi{0}\in\G^{-\rhoa}}  \!\! \EE \left[  e^{-um(\W) \Hui{\epsi{-1}}\jumptime{1}}\indi{\wmi{t}\in\G^{-\rho}, t\!\in\![0,\jumptime{1}]} \right]^{k-1}\!\!.
	\end{split} 
	\end{equation}	
	
	\step{Step 2.} In this step we specifically lower bounds each terms in the lower bound of $\EEi{e^{-um(\W) \Hui{\epsi{-1}}\te}}$. 
	Recall  $\rhoa=\epsi{\gamma}$ where $\gamma$ is a constant such that the results  of Lemmas~\ref{lemma2}$\sim$~\ref{lemm4} holds. The escaping  from the basin $\G$ with a big jump $\epsi{}\jumpi{1}$ occurs when $\Qmi{\jumptime{1}-}^{-1} \Sigmai{\jumptime{1}-}\epsi{}\jumpi{1}\in\G$. Furthermore, $\sup_{0\leq t<\jumptime{1}} \| \wmi{t}-\wgi{t}\| \leq \frac{1}{2} \epsi{\gamma}$ with probability exponentially close to 1 (verified by Lemma~\ref{lemm4}). Meanwhile  $\jumptime{1}=\frac{2}{\alpha} \epsi{\alpha\delta}$ in the $\alpha$-stable ($\Sas$) distribution is much larger than $\ve=\mathcal{O}(\ln(1/\epsi{}))$ with sufficient small $\epsi{}$, $\wgi{t}$ reaches a $\frac{1}{2}\epsi{\gamma}$-neighborhood of the optimum $\wms$ which only requires  time $\ve$. So this actually means $\sup_{0\leq t<\jumptime{1}} \| \wmi{t}-\wms\| \leq  \epsi{\gamma}$.  In this way, to obtain the final lower bound results, we only need to estimate the escaping probability $\PPi{\Qmi{\wm}^{-1} \Sigmai{\wm}\epsi{}\jumpi{1}\in\G^{-\epsi{\gamma}}}$ and 
	$\PPi{\Qmi{\wm}^{-1} \Sigmai{\wm}\epsi{}\jumpi{1}\notin\G}$ uniformly over $\|\wm-\wms\|\leq \epsi{\gamma}$.

	Based on the results in Eqn.~\eqref{AAAasfasf} and~\eqref{ascasdcascas} which provides the upper bound of $\PPi{\Qmi{\wms}^{-1} \Sigmai{\wms}\epsi{}\jumpi{1}\notin\G}$ and some important inequalities, we 	first upper bound the term $\inf_{\|\wm-\wms\|\leq \epsi{-\gamma}} \PPi{\Qmi{\wm}^{-1} \Sigmai{\wm}\epsi{}\jumpi{k}\notin\G^{-\epsi{-\gamma}} }$ as follows:
	\begin{equation*}
	\begin{split}
	&	\inf_{\|\wm-\wms\|\leq \epsi{\gamma}} \!\!\!\!  \PPi{\Qmi{\wm}^{-1} \Sigmai{\wm}\epsi{}\jumpi{k}\notin\G^{-\epsi{\gamma}} }
	\\	=& \inf_{\|\wm-\wms\|\leq \epsi{\gamma}} \!\! \!\!  \PPi{\Qmi{\wm}^{-1} \Sigmai{\wm}\epsi{}\jumpi{k}\notin\G^{-\epsi{\gamma}}\!,  \|\epsi{}\jumpi{k}\|\leq R}\! +\! \PPi{\Qmi{\wm}^{-1} \Sigmai{\wm}\epsi{}\jumpi{k}\notin\G^{-\epsi{\gamma}}\!,  \|\epsi{}\jumpi{k}\| \!> \! R}\\
	\led{172} & \PPi{\Qmi{\wms}^{-1} \Sigmai{\wms}\epsi{}\jumpi{k}\notin\G\!,  \|\epsi{}\jumpi{k}\|\!\leq\! R} \! +\!  \frac{\delta'}{4} \frac{\Hui{\epsi{-1}}}{\Hui{\epsi{-\delta}}}\! +\! \PPi{\Qmi{\wm}^{-1} \Sigmai{\wm}\epsi{}\jumpi{k}\notin\G^{-\epsi{-\gamma}}\!, \|\epsi{}\jumpi{k}\|\! > \!R} \\
	\leq & \PPi{\Qmi{\wms}^{-1} \Sigmai{\wms}\epsi{}\jumpi{k}\notin\G}  +  \frac{\delta'}{4} \frac{\Hui{\epsi{-1}}}{\Hui{\epsi{-\delta}}} + \PPi{\|\epsi{}\jumpi{k}\| > R} \\
	\led{173} &  m(\W)  (1+\delta') \frac{\Hui{\epsi{-1}}}{\Hui{\epsi{-\delta}}} +  \frac{\delta'}{4} \frac{\Hui{\epsi{-1}}}{\Hui{\epsi{-\delta}}} +  \frac{\delta'}{4} \frac{\Hui{\epsi{-1}}}{\Hui{\epsi{-\delta}}} \\
	= & m(\W)  (1+\delta' + \frac{\delta'}{2 m(\W)}) \frac{\Hui{\epsi{-1}}}{\Hui{\epsi{-\delta}}} \led{174} m(\W)  (1+\rho) \frac{\Hui{\epsi{-1}}}{\Hui{\epsi{-\delta}}},
	\end{split}
	\end{equation*}
	where \ding{172} uses the  result in Eqn.~\eqref{AAAasfasf}, \ding{173} uses  Eqn.~\eqref{afdsafascascsac}, and in \ding{174} we set $\delta'$ enough small via setting small $\epsi{}$ such that $\rho\geq \delta' + \frac{\delta'}{2 m(\W)}$. So for any $\rho$ we choose $\delta'>0$ small enough to upper bound 
	\begin{equation*}
	\inf_{\|\wm-\wms\|\leq \epsi{\gamma}} \PPi{\Qmi{\wm}^{-1} \Sigmai{\wm}\epsi{}\jumpi{k}\in\G^{-\epsi{\gamma}} } =1- \inf_{\|\wm-\wms\|\leq \epsi{\gamma}} \PPi{\Qmi{\wm}^{-1} \Sigmai{\wm}\epsi{}\jumpi{k}\notin\G^{-\epsi{\gamma}} }  \geq  1-\aae(1+\rho).
	\end{equation*}
	Similarly, we only need to lower bound the remaining term  $\inf_{\|\wm-\wms\|\leq \epsi{\gamma}} \PPi{\Qmi{\wm}^{-1} \Sigmai{\wm}\epsi{}\jumpi{k}\notin \G }$ as follows:
	\begin{equation*}
	\begin{split}
	&\inf_{\|\wm-\wms\| \!\leq\! \epsi{\gamma}}\!\!\!\! \PPi{\Qmi{\wm}^{-1} \Sigmai{\wm}\epsi{}\jumpi{k}\notin\G } 
	\\=\!& \inf_{\|\wm-\wms\|\!\leq\! \epsi{\gamma}}\!\!\!\! \PPi{\Qmi{\wm}^{-1} \Sigmai{\wm}\epsi{}\jumpi{k}\notin\G\!, \|\epsi{}\jumpi{k}\|\leq R} \!+ \! \PPi{\Qmi{\wm}^{-1} \Sigmai{\wm}\epsi{}\jumpi{k}\notin\G\!,  \|\epsi{}\jumpi{k}\|> R} \\
	\ged{172}\! & \inf_{\|\wm-\wms\|\!\leq\! \epsi{\gamma}} \!\!\!\!\PPi{\Qmi{\wms}^{-1} \Sigmai{\wms}\epsi{}\jumpi{k}\!\notin\!\G^{-\epsi{\gamma}}\!, \|\epsi{}\jumpi{k}\|\!\leq\! R}\! -\! \frac{\delta'}{4} \frac{\Hui{\epsi{-1}}}{\Hui{\epsi{-\delta}}} \!+\! \PPi{\Qmi{\wm}^{-1} \Sigmai{\wm}\epsi{}\jumpi{k}\!\in\!\G\!,   \|\epsi{}\jumpi{k}\|\!> \!R} \\
	\geq \! & \inf_{\|\wm-\wms\|\!\leq\! \epsi{\gamma}} \!\!\!\!\PPi{\Qmi{\wms}^{-1} \Sigmai{\wms}\epsi{}\jumpi{k}\!\notin\!\G\!, \|\epsi{}\jumpi{k}\|\!\leq\! R}\! -\! \frac{\delta'}{4} \frac{\Hui{\epsi{-1}}}{\Hui{\epsi{-\delta}}} \!+\! \PPi{\Qmi{\wm}^{-1} \Sigmai{\wm}\epsi{}\jumpi{k}\!\in\!\G\!,   \|\epsi{}\jumpi{k}\|\!> \!R} \\
	\geq &  \PPi{\Qmi{\wms}^{-1} \Sigmai{\wms}\epsi{}\jumpi{k}\notin\G,\ \|\epsi{}\jumpi{k}\|\leq R} - \frac{\delta'}{4} \frac{\Hui{\epsi{-1}}}{\Hui{\epsi{-\delta}}} \\
	\ged{173} &  \PPi{\Qmi{\wms}^{-1} \Sigmai{\wms}\epsi{}\jumpi{k}\notin\G} - \frac{\delta'}{2} \frac{\Hui{\epsi{-1}}}{\Hui{\epsi{-\delta}}} \\
	\ged{174} & m(\W)  \left(1-\delta' - \frac{\delta'}{2m(\W)}\right)  \frac{\Hui{\epsi{-1}}}{\Hui{\epsi{-\delta}}} \geq m(\W)  (1-\rho) \frac{\Hui{\epsi{-1}}}{\epsi{-\delta}}= \aae(1-\rho),
	\end{split}
	\end{equation*}
	where \ding{172} uses the result in Eqn.~\eqref{AAAasfasf}, \ding{173} uses  Eqn.~\eqref{ascasdcascas}, and \ding{174} uses Eqn.~\eqref{afdsafascascsac}.
	
	Next, for any $\rho>0$ and $\epsi{}$ we can obtain  Laplace transforms for any $u>-1$ as follows:
	\begin{equation}\label{lowerboundPross}
	\begin{split}
	&\inf_{\wmi{0}\in \G^{-\epsi{\gamma}}}\!\! \EE\left[e^{-um(\W)\Hui{\epsi{-1}}\jumptime{1}} \indi{\wmi{t}\in\G^{-\epsi{\gamma}}\!, t\in[0,\jumptime{1}]}\right] \\
	\geq & [1-\aae(1+\rho)] \EE\left[\int_{0}^{+\infty}\!\!e^{- u m(\W)\Hui{\epsi{-1}}\jumptime{1}} \cdot \psiis{\epsi{}}{\delta} e^{-\psiis{\epsi{}}{\delta}\jumptime{1}}\ds \jumptime{1}\right]  \\
	=&  \frac{1-\aae(1+\rho)}{1+u\aae},
	\end{split}
	\end{equation}
	and 
	\begin{equation*}
	\begin{split}
	\inf_{\wmi{0}\in \G^{\epsi{-\gamma}}} \!\! &\EE\left[e^{-um(\W)\Hui{\epsi{-1}}\jumptime{1}} \indi{\wmi{t}\in\G^{-\epsi{\gamma}}\!, t\in[0,\jumptime{1})}  \indi{\wmi{\jumptime{1}}\notin\G}\right] \\
	&  \geq  [1-\aae(1+\rho)] \EE\left[\int_{0}^{+\infty}\!\! e^{- u m(\W)\Hui{\epsi{-1}}\jumptime{1}} \cdot \psiis{\epsi{}}{\delta} e^{-\psiis{\epsi{}}{\delta}\jumptime{1}}\ds \jumptime{1}\right]  
	=   \frac{\aae(1-\rho)}{1+u\aae}.
	\end{split}
	\end{equation*}
	
	\step{Step 3.} Here we summarize the results in Steps 1 and 2 such that we can lower bound the desired results $\EE\left[e^{-um(\W)\Hui{\epsi{-1}}\te} \right] $.  Specifically, from Eqn.~\eqref{adasfdasfa}, for any $\wmi{0}\in\Grd$ we can lower bound 
	\begin{equation*}
	\begin{split}
	\EE\left[e^{-um(\W)\Hui{\epsi{-1}}\te} \right] \geq \frac{\aae(1-\rho)}{1+u\aae} \sum_{k=1}^{+\infty}\left(\frac{1-\aae(1+\rho)}{1+u\aae}\right)^{k-1} = \frac{1-\rho}{1+u +\rho},
	\end{split}
	\end{equation*}	
	where $\rho\downarrow 0$ as $\epsi{}\downarrow 0$.  	The proof is completed. 
\end{proof}

\subsection{Proof of Theorem~\ref{closedistance}}\label{proofofclosedistance}

\begin{proof}
	In this step we prove the sequence  $\{\wgi{t}\}$ produced by Eqn.~\eqref{deterministicversion_RMSP} or \eqref{deterministicversion_adam} locates in a very small neighborhood of the optimum solution $\wms$ of the local basin $\G$ after a very small time interval.
	
	\step{Step 1.} In this step, we prove the first part of Theorem~\ref{closedistance}. Since we assume the function is locally strongly convex, by using Lemmas~\ref{linearconvergenceSGD} and~\ref{linearconvergenceadam}, we know that the sequence $\{\wgi{t}\}$ produced by Eqn.~\eqref{deterministicversion_RMSP} or \eqref{deterministicversion_adam} exponentially converges to the minimum $\wms$ at the current local basin $\G$. So for any initialization $\wmi{0}\in\G$, we have  
	\begin{equation*}
	\begin{split}
	\|\wgi{t}-\wms\|_2^2  \leq c_1\exp\left(-c_2 t \right),
	\end{split}
	\end{equation*}
	where $c_1= \frac{2\Delta}{\mu} $ and $c_2=\frac{2\mu\tau}{ \betag \left(\vmax+\thres\right) +\mu\tau} \left(\betag -\frac{\betav}{4} \right)$ in Adam,  $c_1= \frac{2\Delta}{\mu} $ and $c_2= 2\mu$ in \Sgd.   Therefore, for any initialization $\wmi{0}\in\G$ and sufficient small $\epsi{}$, we can obtain
	\begin{equation*}
	\begin{split}
	\|\wgi{t}-\wms\|_2^2  \leq \epsi{\rhoa} \ \text{if}\ t\geq \ve = \frac{1}{c_3} \ln\big(\frac{2\Delta}{\mu \epsi{\rhoa}}\big).
	\end{split}
	\end{equation*}
	where $c_3=\frac{2\mu\tau}{ \betag \left(\vmax+\thres\right) +\mu\tau} \big(\betag -\frac{\betav}{4} \big)$ in \Adam,  $c_3 =2\mu$ in \Sgd, and  $\Delta = \Fm(\wmi{0})-\Fm(\wms)$.

	\step{Step 2.} In this step, we prove the second part of Theorem~\ref{closedistance}. By replacing $p$ with $p/2$ in Lemma~\ref{proofoflemm4}, we can directly obtain the results.
\end{proof}

\section{Proofs of Auxiliary Theories and Lemmas in Appendix~\ref{notations}}\label{proofofAuxiliaryLemma}
Before analysis, we first introduce two useful lemmas which will be used in subsequent analysis. 

\begin{lem}[Grönwall's Lemma~\cite{gronwall1919note}]\label{Gronwallsinequality}
	Suppose $g(s): [0,t_0]$ is a non-negative continues function. If for almost $s\in[0,t_0]$
	\begin{equation*}
	g'(s)\leq q(s) g(s)
	\end{equation*}
	where $q(s)$ is a continuous function, then we have 
	\begin{equation*}
	g(t)\leq g(0) \exp\left(\int_{0}^t q(s) \ds s \right).
	\end{equation*}
\end{lem}

\begin{lem}[Theorem 5.3 in~\cite{papapantoleon2008introduction}]\label{setlemma}
	Consider a set $\Am\in\mathcal{B}(\Rs{}\setminus 0)$ with $0\in \bar{\Am}$ and a function $f: \Rs{}\rightarrow \Rs{}$ with Borel measurable and finite on $\Am$. Then we have \\
	(1) The process $(\int_{0}^{t} \int_{\Am} f(x) \nu(\ds s,\ds x))_{0\leq t \leq T}$ is a compound Poisson process with characteristic function
	\begin{equation*}
	\EE \left(\exp\left(i\lambda \int_{0}^{t} \int_{\Am} f(x) \mu^{\levy}(\ds s,\ds x)\right) \right) = \exp\left(t \int_{\Am} (e^{i \lambda f(x)}-1)\nu(\ds x)\right).
	\end{equation*} 
	(2) If $f\in L^1(\Am)$, then
	\begin{equation*}
	\EE \left(\int_{0}^{t} \int_{\Am} f(x) \mu^{\levy}(\ds s,\ds x)\right) = t \int_{\Am}  f(x) \nu(\ds x).
	\end{equation*} 
\end{lem}

\subsection{Proof of Theorem~\ref{linearconvergenceSGD} for the Linear Convergence of \levyp-driven \Sgd~SDE~\eqref{deterministicversion_RMSP}}\label{linearconvergenceSGDproof}

\begin{proof}
	\step{Step 1.} In this step, we upper bound the gradient norm of the Lyapunov function 	$\LL(t) = \Fm(\wgi{t}) -\Fm(\wms)$ of~\eqref{deterministicversion_RMSP} with $\Qgi{t}=\Imm$ and $\betav=0$. More specifically, we can upper bound  $\ds \LL(t)$ as follows:
	\begin{equation}\label{Lyapunov_SGD}
	\ds \LL(t) = \langle \nabla \Fm(\wgi{t}), \ds \wgi{t}\rangle = \left\langle \nabla \Fm(\wgi{t}), - \nabla \Fm(\wgi{t})\right\rangle =- \|\nabla\Fm(\wgi{t})\|^2_2.
	\end{equation} 
	
	\step{Step 2.} Here we prove the linear convergence behavior of $\LL(t) = \Fm(\wgi{t}) -\Fm(\wms)$ by using the results in Step 1. Since $\Fm(\wm)$ is locally $\mu$-strongly convex,  then we have
	\begin{equation*}
	\begin{split}
	\Fm(\ym) \geq \Fm(\wm) +\langle \nabla \Fm(\wm), \ym-\wm \rangle + \frac{\mu}{2} \| \ym-\wm\|^2_2.
	\end{split}
	\end{equation*}
	Next, by minimizing $\ym$ on both side ($\ym=\wmi{*}$ for the left side and $\ym=\wm-\frac{1}{\mu} \nabla \Fm(\wm)$ for the right side), it yields
	\begin{equation}\label{ustronglyconvex1}
	\begin{split}
	\|\nabla \Fm(\wm)\|_2^2 \geq 2 \mu(\Fm(\wm)- \Fm(\wmi{*})).
	\end{split}
	\end{equation}
	Hence, plugging the above equation into~Eqn.~\eqref{Lyapunov_SGD}  gives
	\begin{equation*}
	\ds \LL(t) \leq  - 2\mu  (\Fm(\wm)- \Fm(\wmi{*}))= - 2\mu \LL(t).
	\end{equation*} 
	
	In this way, by using the result in Lemma~\ref{Gronwallsinequality}, we can easily obtain
	\begin{equation*}
	\begin{split}
	\LL(t)\leq \LL(0) \exp\left(-\int_{0}^t  2\mu \ds s \right)
	\leq \Delta \exp\left(- 2\mu t  \right),
	\end{split}
	\end{equation*}
	where we use $\LL(0)=\Fm(\wgi{0}) -\Fm(\wms)=\Delta$ where $\wms$ is the optimum of the current basin.
	
	\step{Step 3.} Finally, we explore the local strong-convexity of $\Fm(\wm)$ to show the linear  convergence of $\|\wgi{t}-\wms\|_2^2$. Specifically, by using the strongly convex property of $\Fm(\wm)$, we can obtain 
	\begin{equation*}
	\begin{split}
	\Fm(\wm)-\Fm(\wms) \geq \frac{\mu}{2} \|\wm-\wms\|_2^2.
	\end{split}
	\end{equation*}
	So this gives 
	\begin{equation*}
	\begin{split}
	\|\wgi{t}-\wms\|_2^2  \leq \frac{2\Delta}{\mu} \exp\left(- 2\mu t \right).
	\end{split}
	\end{equation*}
	The proof is completed. 
\end{proof}

\subsection{Proof of Theorem~\ref{linearconvergenceadam} for the Linear Convergence of \levyp-driven \Adam~SDE~\eqref{deterministicversion_adam}}\label{linearconvergenceadamproof}
\begin{proof}
	\step{Step 1.} In this step, we upper bound the gradient norm of the Lyapunov function of~\eqref{deterministicversion_adam} defined as  
	\begin{equation}\label{Lyapunov}
	\LL(t) = \Fm(\wgi{t}) -\Fm(\wms)+ \frac{1}{2} \|\mgi{t}\|_{\smi{t}^{-1}}^2,
	\end{equation}
	where $\smi{t}=\frac{h_t}{\mui{t}}\left(\sqrt{\omegai{t}\vgi{t}}+\thres \right)$ with $h_t= \betag$, $\mui{t}=(1-e^{-\betag t})^{-1}$ and $\omegai{t}=(1-e^{-\betav t})^{-1}$. Here we define $\|\xm\|_{\ym}^2=\sum_i \ym_i\xm_i^2$. Then we can compute the derivative of Lyapunov function as 
	\begin{equation}\label{ddLyapunov}
	\ds \LL(t) =\underbrace{\langle \nabla \Fm(\wgi{t}), \ds \wgi{t}\rangle  + \sum_{i=1}^{d}    \frac{1}{\smi{t,i}}\mgi{t,i} \ds \mgi{t,i} -\sum_{i=1}^{d}   \frac{1}{2\smi{t,i}^2} \mgi{t,i}^2 \nabla_{\vgi{t}} \smi{t,i} \ds \vgi{t,i}}_{P_1}  \underbrace{-\sum_{i=1}^{d}   \frac{1}{2 \smi{t,i}^2} \mgi{t,i}^2  \nabla_t \smi{t,i}}_{P_2}, 
	\end{equation}
	where $\mgi{t,i}$, $\vgi{t,i}$ and $\smi{t,i}$ respectively denote the $i$-th entries of $\mgi{t}$, $\vgi{t}$ and $\smi{t}$.

	We first consider Adam in which $h_t= \betag$, $\mui{t}=(1-e^{-\betag t})^{-1}$, and $\omegai{t}=(1-e^{-\betav t})^{-1}$. We also assume $\betag \leq \betav\leq 2\betag$ which is consistent with the practical setting where $\betag=0.9$ and $\betav=0.999$. Let $[\nabla \Fbhi{t} ^2]_i$ denotes the $i$-th entry of the vector $\nabla \Fbhi{t} ^2$. Under this setting, we can first upper bound the first term $P_1$ as follows: 
	\begin{equation*}
	\begin{split}
	&P_1 \\
	=&  \langle \nabla \Fm(\wgi{t}), \ds \wgi{t}\rangle  + \sum_{i=1}^{d}    \frac{1}{\smi{t,i}}\mgi{t,i} \ds \mgi{t,i}  -\sum_{i=1}^{d}   \frac{1}{2\smi{t,i}^2} \mgi{t,i}^2 \nabla_{\vgi{t}} \smi{t,i} \ds \vgi{t,i} \\
	= &  \langle \nabla \Fm(\wgi{t}),  -\frac{\mui{t}\mgi{t}}{\sqrt{\omegai{t}\vgi{t}}+\thres} \rangle  + \betag \sum_{i=1}^{d}    \frac{1}{\smi{t,i}}\mgi{t,i} (\nabla\Fm_i(\wmi{t})  - \mgi{t,i}) -\sum_{i=1}^{d}   \frac{\betav}{2\smi{t,i}^2} \mgi{t,i}^2 ([\nabla \Fbhi{t} ^2]_i - \vgi{t,i}) \nabla_{\vgi{t}} \smi{t,i}\\
	= & - \betag \left\langle \nabla \Fm(\wgi{t}),  \frac{\mgi{t}}{\smi{t}}\right\rangle  + \betag \left\langle \nabla \Fm(\wgi{t}),  \frac{\mgi{t}}{\smi{t}}\right\rangle  -  \betag \sum_{i=1}^{d}    \frac{1}{\smi{t,i}}\mgi{t,i}^2   -\betav\sum_{i=1}^{d}   \frac{1}{2 \smi{t,i}^2} \mgi{t,i}^2   ([\nabla \Fbhi{t} ^2]_i - \vgi{t,i})\nabla_{\vgi{t}} \smi{t,i}\\
	= & -  \betag \sum_{i=1}^{d}    \frac{1}{\smi{t,i}}\mgi{t,i}^2   -\betav\sum_{i=1}^{d}   \frac{1}{2 \smi{t,i}^2} \mgi{t,i}^2   ([\nabla \Fbhi{t} ^2]_i - \vgi{t,i})\nabla_{\vgi{t}} \smi{t,i}.
	\end{split}
	\end{equation*}
	Next, we plug the specific formulation of $\nabla_{\vgi{t}} \smi{t,i}= \frac{\betag\sqrt{\omegai{t}}}{2\mui{t}\sqrt{\vgi{t,i}}}$ into the above equation and obtain:
	\begin{equation*}
	\begin{split}
	P_1 	= & -  \betag \sum_{i=1}^{d}    \frac{1}{\smi{t,i}}\mgi{t,i}^2   -\betav\sum_{i=1}^{d}   \frac{1}{2 \smi{t,i}^2} \mgi{t,i}^2   ([\nabla \Fbhi{t} ^2]_i - \vgi{t,i})  \frac{\betag\sqrt{\omegai{t}}}{2\mui{t}\sqrt{\vgi{t,i}}} \\
	= & - \betag^2 \sum_{i=1}^{d}    \frac{\mgi{t,i}^2}{\mui{t} \smi{t,i}^2} \left( \thres + (1-\frac{\betav}{4\betag}) \sqrt{\omegai{t}\vgi{t,i}}  + \frac{\betav}{4\betag}\frac{[\nabla \Fbhi{t} ^2]_i  \sqrt{\omegai{t}}}{\sqrt{\vgi{t,i}}} \right)\\
	= & - \betag \sum_{i=1}^{d}    \frac{\mgi{t,i}^2}{\smi{t,i}} \left(1-\frac{\betav}{4\betag} +   \frac{\betav \thres}{4\betag ( \thres+\sqrt{\omegai{t}\vgi{t,i}} )}  + \frac{\betav}{4\betag}\frac{[\nabla \Fbhi{t} ^2]_i  \sqrt{\omegai{t}}}{\sqrt{\vgi{t,i}} ( \thres+\sqrt{\omegai{t}\vgi{t,i}} )} \right)\\
	\leq  & - \left(\betag -\frac{\betav}{4}\right) \sum_{i=1}^{d}    \frac{\mgi{t,i}^2}{\smi{t,i}} = - \left(\betag -\frac{\betav}{4}\right) \|\mgi{t}\|^2_{\smi{t}^{-1}}.\\
	\end{split}
	\end{equation*}
	Then we consider the second term $P_2$ under the setting $\smi{t}=\frac{\betag}{\mui{t}}\left(\sqrt{\omegai{t}\vgi{t}}+\thres \right)$ with   $\mui{t}=(1-e^{-\betag t})^{-1}$ and $\omegai{t}=(1-e^{-\betav t})^{-1}$. Similarly, we can upper bound $P_2$ as 
	\begin{equation*}
	\begin{split}
	P_2 =&  -\sum_{i=1}^{d}   \frac{1}{2 \smi{t,i}^2} \mgi{t,i}^2  \nabla_t \smi{t,i} \\
	= &  -\betag\sum_{i=1}^{d}   \frac{1}{2 \smi{t,i}^2} \mgi{t,i}^2 \left( \betag e^{-\betag t} \left(\thres+ \sqrt{\frac{\vgi{t,i}}{1-e^{-\betav t}}}\right)-\frac{1}{2} \betav e^{-\betav t} \frac{1-e^{-\betag t}}{1-e^{-\betav t}} \sqrt{ \frac{\vgi{t,i}}{1-e^{-\betav t}}} \right)\\
	= & - \frac{\betag^2}{2} \sum_{i=1}^{d}    \frac{\mgi{t,i}^2}{ \mui{t}  \smi{t,i}^2} \frac{e^{-\betag t}}{1-e^{-\betag t}}\left( \thres + \left(1-\frac{\betav e^{-\betav t} (1-e^{-\betag t})}{2\betag e^{-\betag t} (1- e^{-\betav t})}\right) \sqrt{\frac{\vgi{t,i}}{1-e^{-\betav t}}} \right)\\
	\led{172} & - \frac{\betag^2}{2} \sum_{i=1}^{d}    \frac{\mgi{t,i}^2}{ \mui{t}  \smi{t,i}^2} \frac{e^{-\betag t}}{1-e^{-\betag t}}\left( \thres + \left(1-\frac{\betav }{2\betag }\right) \sqrt{\frac{\vgi{t,i}}{1-e^{-\betav t}}} \right)\\
	= & - \frac{\betag}{2} \sum_{i=1}^{d}    \frac{\mgi{t,i}^2}{  \smi{t,i} } \frac{e^{-\betag t}}{1-e^{-\betag t}}\left( 1-\frac{\betav }{2\betag }  + \frac{\betav \thres}{2\betag (\thres + \sqrt{\omegai{t}\vgi{t,i}})} \right)\\
	\leq  & - \frac{1}{2} \left(\betag-\frac{\betav}{2}\right)  \frac{e^{-\betag t}}{1-e^{-\betag t}} \sum_{i=1}^{d}    \frac{\mgi{t,i}^2}{  \smi{t,i} }=- \frac{1}{2} \left(\betag-\frac{\betav}{2}\right)  \frac{e^{-\betag t}}{1-e^{-\betag t}}  \|\mgi{t}\|^2_{\smi{t}^{-1}}\led{173} 0,
	\end{split}
	\end{equation*}
	where \ding{172} uses  $\frac{\betav e^{-\betav t} (1-e^{-\betag t})}{2\betag e^{-\betag t} (1- e^{-\betav t})} \leq \frac{\betav}{2\betag}$ since $\betav \geq \betag$; in \ding{173} we assume  $\betag-\frac{\betav}{2}>0$. 	Therefore, by combining the upper bounds of $P_1$ and $P_2$ we can upper bound
	\begin{equation}\label{tempresult}
	\begin{split}
	\ds \LL(t) \leq - \left[ \betag -\frac{\betav}{4}   \right] \|\mgi{t}\|^2_{\smi{t}^{-1}}.
	\end{split}
	\end{equation}
	On the other hand, noting $h_t= \betag$, $\mui{t}=(1-e^{-\betag t})^{-1}$ and $\omegai{t}=(1-e^{-\betav t})^{-1}$, we have
	\begin{equation*}
	\begin{split}
	\smi{t,i}=&\frac{h_t}{\mui{t}}\left(\thres+\sqrt{\omegai{t}\vgi{t,i}}\right) = \betag (1-e^{-\betag t})\left(\thres + \sqrt{\frac{\vgi{t,i}}{1-e^{-\betav t}}} \right)\leq \betag \left(\thres + \frac{1-e^{-\betag t}}{\sqrt{1-e^{-\betav t}}} \sqrt{\vgi{t,i}} \right)\\
	\led{172}& \betag \left(\thres + \frac{1-e^{-\betag t}}{1-e^{-\betav t/2}} \sqrt{\vgi{t,i}} \right) \led{173} \betag \left(\thres + \vmax\right),
	\end{split}
	\end{equation*}
	where \ding{172} uses $\sqrt{1-x}\geq  1- \sqrt{x}$ for $0\leq x\leq 1$ and \ding{173} holds since  $ \sqrt{\vgi{t,i}} \leq \vmax$.  By using the assumption $\|\mgi{t}\|^2\geq \tau \|\nabla \Fm(\wgi{t})\|^2$, we can establish 
	\begin{equation}\label{mgradient}
	\begin{split}
	\|\mgi{t}\|^2_{\smi{t}^{-1}} \geq \frac{1}{\betag \left(\thres + \vmax\right)} \|\mgi{t}\|_2^2 \geq \frac{\tau}{\betag \left(\thres + \vmax\right)}  \|\nabla \Fm(\wgi{t})\|^2_2.
	\end{split}
	\end{equation}
	Then from the locally $\mu$-strongly convex property Eqn.~\eqref{ustronglyconvex1}: 
	\begin{equation*} 
	\begin{split}
	\|\nabla \Fm(\wm)\|_2^2 \geq 2 \mu(\Fm(\wm)- \Fm(\wmi{*})).
	\end{split}
	\end{equation*}
	then we plug the above inequality into~Eqn.~\eqref{mgradient} and establish
	\begin{equation*}
	\begin{split}
	\|\mgi{t}\|^2_{\smi{t}^{-1}} \geq \frac{1}{\betag \left(\thres + \vmax\right)} \|\mgi{t}\|_2^2 \geq \frac{2 \mu\tau}{\betag \left(\thres + \vmax\right)}  (\Fm(\wgi{t})- \Fm(\wmi{*})).
	\end{split}
	\end{equation*}
	
	Finally, we can write Eqn.~\eqref{tempresult} as 
	\begin{equation*}\label{finalresult}
	\begin{split}
	\ds \LL(t) \leq &  -  \frac{2\mu\tau}{ \betag \left(\thres + \vmax\right) +\mu\tau} \left[ \betag -\frac{\betav}{4}    \right]  \left(\frac{1}{2}+\frac{\betag \left(\thres + \vmax\right)}{2 \mu\tau}  \right)\|\mgi{t}\|^2_{\smi{t}^{-1}}\\
	\leq &-\frac{2\mu\tau}{ \betag \left(\thres + \vmax\right) +\mu\tau} \left[ \betag -\frac{\betav}{4}    \right] \left(\Fm(\wgi{t}) -\Fm(\wms)+\frac{1}{2} \|\mgi{t}\|^2_{\smi{t}^{-1}}\right)\\
	=& - c_1\LL(t) ,
	\end{split}
	\end{equation*}
	where $c_1=\frac{2\mu\tau}{ \betag \left(\thres + \vmax\right) +\mu\tau} \left[ \betag -\frac{\betav}{4}  \right]$. 
	
	\step{Step 2.} Here we prove the linear convergence behavior of $\LL(t) = \Fm(\wgi{t}) -\Fm(\wms)$ by using the results in Step 1. More specifically, by using the result in Lemma~\ref{Gronwallsinequality}, we can easily obtain
	\begin{equation*}
	\begin{split}
	\LL(t)\leq& \LL(0) \exp\left(\int_{0}^t c_1 \ds s \right)= \LL(0) \exp\left(-\frac{2\mu\tau}{ \betag \left(\thres + \vmax\right) +\mu\tau}   \left(\betag -\frac{\betav}{4} \right)t  \right)\\
	\led{172} &(\Fm(\wgi{0}) -\Fm(\wms)) \exp\left(-\frac{2\mu\tau}{ \betag \left(\thres + \vmax\right) +\mu\tau}   \left(\betag -\frac{\betav}{4} \right)t \right),
	\end{split}
	\end{equation*}
	where \ding{172} uses $\LL(0)=\Fm(\wgi{0}) -\Fm(\wms)= \Delta$ due to $\mgi{0}=\bm{0}$.
	
	\step{Step 3.} Finally, we explore the local strong-convexity of $\Fm(\wm)$ to show the linear  convergence of $\|\wgi{t}-\wms\|_2^2$. Specifically, by using the strongly convex property of $\Fm(\wm)$, we can obtain 
	\begin{equation*} 
	\begin{split}
	\Fm(\wm)-\Fm(\wms) \geq \frac{\mu}{2} \|\wm-\wms\|_2^2.
	\end{split}
	\end{equation*}
	So this gives 
	\begin{equation*}
	\begin{split}
	\|\wgi{t}-\wms\|_2^2  \leq \frac{2\Delta}{\mu} \exp\left(-\frac{2\mu\tau}{ \betag \left(\thres + \vmax\right) +\mu\tau}  \left(\betag -\frac{\betav}{4} \right)t    \right).
	\end{split}
	\end{equation*}
	The proof is completed. 
\end{proof}

\subsection{Proof of Lemma~\ref{lemma2}}\label{proofoflemma2}
\begin{proof}
	To begin with, the process $\xii{}$ is defined as $\xii{t}=\sum_{s\leq t} \Delta \levyi{s} \indi{\|\levyi{s}\|\leq \epsi{-\delta}}$. Then by setting the set $\Am=\{\ym\ |\ \|\ym\|\leq \epsi{-\delta}\}$ in Lemma~\ref{setlemma} and noting $f(x)=x\in L^1(\Am)$, one can find $\EE[\xii{t}]=t \int_{\Am}  f(x) \nu(\ds x)$. Therefore, we can decompose the process $\xii{}$ into two processes $\xwi{}$ and linear drift, namely,
	\begin{equation*}
	\xii{t}=\xwi{t}+\mue t,
	\end{equation*}
	where $\xwi{}$ is a zero mean \levyp martingale with bounded jumps.  Then we prove our results in two steps. 
	
	\step{Step 1.} We first estimate the value of $\mue$. Since $\xii{}$ is a \levyp process, by \levyp-It$\hat{\text{o}}$  decomposition theory~\cite[Theorem~6.1]{papapantoleon2008introduction} its characteristic function is of form $$\EE [e^{i\langle \lam, \xii{t}\rangle }] =\exp\left(  t  \mathlarger{\int}_{\Rs{d}\setminus \{\bm{0}\}} \left( e^{i\langle\lam,\ym \rangle}- 1 -i \langle \lam, \ym\rangle \indi{\|\ym\|\leq 1}\right)\indi{\|\ym\|\leq \epsi{-\delta}}\ds \ym\right),$$ 
	which can be further split into two \levyp processes $\xii{(1)}$ and $\xii{(2)}$ with characteristic functions 
	\begin{equation*}
	\begin{split}
	\EE [e^{i\langle \lam, \xii{(1), t}\rangle }] =\exp\left(  t  \mathlarger{\int}_{0<\|\ym\|<1} \left( e^{i\langle\lam,\ym \rangle}- 1 -i \langle \lam, \ym\rangle \right)\ds \ym\right) 
	\end{split}
	\end{equation*}
	and 
	\begin{equation*}
	\begin{split}
	\EE [e^{i\langle \lam, \xii{(2), t}\rangle }] =\exp\left(  t  \mathlarger{\int}_{1\leq \|\ym\|\leq \epsi{-\delta}} \left( e^{i\langle\lam,\ym \rangle}- 1 \right)\ds \ym\right).
	\end{split}
	\end{equation*}
	Let we consider $\xii{}$ on the set $\{\ym\ |\ 0<\|\ym\|\leq1 \}$.  We construct a compensated compound Poisson process 
	\begin{equation*}
	\levy_t'= \sum_{s\leq t} \Delta \levy_s' \indi{1>\|\Delta\levy_s\|>\threslemma}- t\mathlarger{\int}_{1>\|\ym\|>\threslemma} \ym\nu(\ds \ym)= \mathlarger{\int}_{0}^t \mathlarger{\int}_{1>\|\ym\|>\threslemma}  \ym \mu^{\levy} (\dm \ym, \ds s) - t\mathlarger{\int}_{1>\|\ym\|>\threslemma} \ym\nu(\ds \ym),
	\end{equation*}
	where $\threslemma$ is a very small constant. 
	By applying Lemma~\ref{setlemma} on $\sum_{s\leq t} \Delta \levy_s' \indi{1>\|\Delta\levy_s\|>\threslemma}$, the characteristic function of $\levy_t'$ is 
	$$\EE [e^{i\langle \lam, \levy_t' \rangle }] =\exp\left(  t  \mathlarger{\int}_{\threslemma<\|\ym\|<1} \left( e^{i\langle\lam,\ym \rangle}- 1 -i \langle \lam, \ym\rangle \right)\ds \ym\right).$$ 
	This means that there exists a \levyp process $\levy'$ which is a square integral martingale such that $\levy'\rightarrow\xii{(1)}$ as $\threslemma \rightarrow 0$.  As   $\levy'$  is a square integral martingale, we have $\EE(\xii{(1)})=\EE(\levy')=\bm{0}$, which means that $\mue$ is only related to $\xii{(2)}$. Therefore, we have 
	\begin{equation*}
	\begin{split}
	&\mue^i =\EE[\xii{(2)}^i] = \mathlarger{\int}_{1\leq \|\ym\|\leq \epsi{-\delta}} \ymi{i} \nu(\ds \ym),\quad (i=1,\cdots,d)\\
	&\|\mue\|^2 = \mathlarger{\int}_{1\leq \|\ym\|\leq \epsi{-\delta}} \|\ym\|^2 \nu(\ds \ym)=- \mathlarger{\int}_1^{\epsi{-\delta}} u^2 \ds \Hui{u}=-u^2 \Hui{u}\big|_{1}^{\epsi{-\delta}} + 2\mathlarger{\int}_1^{\epsi{-\delta}} u  \Hui{u} \ds \leq \epsi{-2\delta}\Hui{1}.
	\end{split}
	\end{equation*}
	Thus, we can bound $\|\mue\|\leq \epsi{-\delta} \sqrt{\Hui{1}}$.  Finally, by setting $\theta_0 = (1-\delta)/3$ and $\rho_0=(1-\delta)/4$ we can obtain $\epsi{}\|\mue\|\Te = \epsi{1-\delta - \theta}\sqrt{\Hui{1}}\leq \epsi{2\rho}$ by setting $\epsi{}$ sufficient small such that $\Hui{1}\leq \frac{1}{\epsi{1-2\rho -\delta -\theta}}$.

	\step{Step 2.}  Since the increment is non-negative, the quadratic variation process $[\epsi{} \xwi{}]_{t}^d$ is a \levyp subordinator, namely,
	\begin{equation*}
	[\epsi{} \xwi{}]_{t}^d = \epsi{2} \sum_{s\leq t} \|\Delta \xwi{s}\|^2 = \epsi{2} \int_o^t \int_{0<\|\ym\|\leq \epsi{-\delta}} \|\ym\|^2 N(\ds \ym, \ds s),
	\end{equation*}
	where $\Delta\xwi{s}=\xwi{s} -\xwi{s-}$ where $ \xwi{s-} =\lim_{t\uparrow s} \xwi{t}$. 
	
	Since the jumps of $[\epsi{} \xwi{}]^d$ are bounded, its Laplace transform is well-defined for all $\lambda \in \Rs{}$:
	\begin{equation*}
	\begin{split}
	\EE e^{\lambda[\epsi{} \xwi{}]_{t}^d} = \exp\left( t  \int_{0<\|\ym\|\leq \epsi{-\delta}} (e^{\lambda \epsi{2} \|\ym\|^2}-1 )  \nu(\ds \ym) \right)= \exp\left(- t  \int_{0<u \leq \epsi{-\delta}} (e^{\lambda \epsi{2} u^2}-1 )  \ds \Hui{u} \right). 
	\end{split}
	\end{equation*}
	For any $\lambda>0$, the exponential Chebyshev inequality indicates 
	\begin{equation}\label{adas}
	\begin{split}
	\PPi{[\epsi{} \xwi{}]_{\Te}^d >\epsi{\rho}} =& \PPi{e^{\lambda [\epsi{} \xwi{}]_{\Te}^d} > e^{\lambda \epsi{\rho}}} \leq  e^{-\lambda \epsi{\rho}} \EE[e^{\lambda [\epsi{} \xwi{}]_{\Te}^d}]\\
	= & \exp\left(-\lambda \epsi{\rho}- \Te \int_{0<u \leq \epsi{-\delta}} (e^{\lambda \epsi{2} u^2}-1 )  \ds \Hui{u} \right). 
	\end{split}
	\end{equation}
	For $\lambda =\lame  = \epsi{-2\rho}$ with $0< \rho < \rho_{0} = (1-\delta)/4$ we have $\max_{0\leq u \leq \epsi{-\delta}} \lambda \epsi{2}u^2 \leq \lame \epsi{2(1-\delta)} \leq \epsi{\frac{3}{2} (1-\delta)} \downarrow 0$ as $\epsi{}\downarrow 0$. With help of the elementary inequality $e^x-1\leq 2 x$ for small positive $x$ the second summand appearing in the exponent in right-hand side of~\eqref{adas} can be now established as 
	\begin{equation*} 
	\begin{split}
	\left| \Te \int_{0<u \leq \epsi{-\delta}} (e^{\lame \epsi{2} u^2}-1 )  \ds \Hui{u} \right| \leq & \left| 2\Te\lame \epsi{2}   \left( \int_{0<u\leq 1} + \int_{1<u\leq \epsi{-\delta}} \right) u^2  \ds \Hui{u} \right| \\
	\leq & 2\Te\lame \epsi{2}  \left|    \int_{0<u\leq 1}  u^2  \ds \Hui{u} \right| + 2\Te\lame \epsi{2(1-\delta)}  \left|    \int_{1<u\leq \epsi{-\delta}}   \ds \Hui{u} \right|\\
	\leq & 2C\Te\lame \epsilon^2   + 2\Hui{1} \Te \lame \epsi{2(1-\delta)}
	\end{split}
	\end{equation*}
	where $C=\left|    \int_{0<u\leq 1}  u^2  \ds \Hui{u} \right| \in(0, +\infty)$ is a constant. Consequently, for all $0<\rho\leq \rho_0$ and $0< \theta <\theta_0$ we see that the exponential inequality 
	\begin{equation*} 
	\begin{split}
	\PPi{[\epsi{} \xwi{}]_{\Te}^d > \epsi{\rho}} \leq    \exp\left(-\lame \epsi{\rho}+ 2C\Te\lame \epsi{2}   + 2\Hui{1} \Te \lame \epsi{2(1-\delta)} \right)\leq \exp(-\epsi{-\rho/2})
	\end{split}
	\end{equation*}
	holds for small enough $\epsi{}$ with $p\in(0,\rho/2)$. This is because 
	\begin{equation*} 
	\begin{split}
	&-\lame \epsi{\rho}+ 2C\Te\lame \epsi{2}   + 2\Hui{1} \Te \lame \epsi{2(1-\delta)}=  -\epsi{-\rho} + 2C \epsi{2-\frac{1-\delta}{3}-\frac{1-\delta}{2}}+2 \Hui{1} \epsi{2(1-\delta)-\frac{1-\delta}{3}-\frac{1-\delta}{2}}\\
	\leq & -\epsi{-\rho} + 2(C+\Hui{1})   \epsi{2(1-\delta)-\frac{1-\delta}{3}-\frac{1-\delta}{2}}
	\leq  -\epsi{-\rho} + 2(C+\Hui{1})   \epsi{\frac{7}{6}(1-\delta)}\led{172} -\epsi{-\rho/2},
	\end{split}
	\end{equation*}
	where \ding{172} holds by setting $\epsi{}$ enough small such that $(\epsi{-\rho} - 2(C+\Hui{1})   \epsi{\frac{7}{6}(1-\delta)})/\epsi{-\rho/2}\geq \epsi{-\rho/2} - 2(C+\Hui{1})   \epsi{\frac{7}{6}(1-\delta)+\frac{\rho}{2}}\geq 1$.
	The proof is completed.
\end{proof}

\subsection{Proof of Lemma~\ref{lemma3}}\label{proofoflemma3}
\begin{proof}
	\step{Step 1.} Suppose $\sup_{t\geq 0} \|\gmi{t}\|\leq c_g$ for some constant $c_g>0$. Then we consider the one-dimensional martingale 
	\begin{equation*}
	\Mmi{t} =  \sum_{i=1}^{d} \int_{0}^{t}  \gmii{s-}{i} \ds \xwii{s}{i} .
	\end{equation*}  
	We estimate the probability of a deviation of the size $\epsi{\rho}$ of $\epsi{}\Mmi{t}$ from zero with help of the exponential inequality for martingales, see Theorem 26.17 (i) in~\cite{kallenberg2006foundations}. Indeed for any $\rho>0$ and $\theta>0$, we have 
	\begin{equation*}
	\PPi{\sup_{t\leq \Te} |\epsi{} \Mmi{t}| \geq \epsi{\rho}} \leq \PPi{\sup_{t\leq \Te}|\epsi{}  \Mmi{t}| \geq \epsi{\rho}\ \big|\ [\epsi{}  \Mmi{}]_{\Te} \leq \epsi{4\rho}} + \PPi{\ [\epsi{}  \Mmi{}]_{\Te} > \epsi{4\rho}}.
	\end{equation*}
	Inspecting the proofs of Lemma 26.19 and Theorem 26.17 (i) in~\cite{kallenberg2006foundations} we get that for any $\lambda>0$
	\begin{equation*}
	\PPi{\sup_{t\leq \Te}|\epsi{}  \Mmi{t}| \geq \epsi{\rho}\ \big|\ [\epsi{}  \Mmi{}]_{\Te} \leq \epsi{4\rho}} \leq \exp\left(-\lambda \epsi{\rho} + \lambda^2 h(\lambda c_g \epsi{1-\delta})\epsi{4\rho}\right),
	\end{equation*}
	where $h(x)=-(x+\ln(1-x)_{+})x^{-2}$. For any $0< \rho <\rho_1=(1-\delta)/2$ we set $\lambda=\lame=\epsi{-2\rho}$ so that $h(\lame c_g \epsi{1-\delta})\rightarrow 1/2$ as $\epsi{}\rightarrow 0$ by using LHopital's rule. Hence we obtain the estimate 
	\begin{equation*}
	\PPi{\sup_{t\leq \Te}|\epsi{} \Mmi{t}| \geq \epsi{\rho}\ \big|\ [\epsi{} \Mmi{}]_{\Te} \leq \epsi{4\rho}} \leq \exp\left(- \epsi{-\rho} +\frac{1}{2}\right) \led{172}  \exp\left(- \epsi{-\rho/2} \right)\leq \exp\left(- \epsi{-p} \right),
	\end{equation*}
	which holds for small enough $\epsi{}$ and $p\in(0,\rho/2]$. In \ding{172}, we set $\epsi{}$ enough small such that $0<\epsi{-\rho/2}-\epsi{\rho/2}\leq 1$.  
	
	\step{Step 2.} Since $\|\gmi{t}\|\leq c_g$ is well bounded, then there is a constant $c_1$ with 
	\begin{equation*}
	[\epsi{} \Mmi{}]_{t} =\int_{0}^{t} \gmii{s-}{2} \ds [\epsi{} \xwi{}]_s^d \leq c_1 [\epsi{} \xwi{}]_t^d.
	\end{equation*}
	Then we can use Lemma~\ref{lemma2} to upper bound:
	\begin{equation*}
	\PPi{[\epsi{} \Mmi{}]_{\Te} \geq \epsi{4\rho}} \leq \PPi{c_1 [\epsi{} \xwi{}]_t^d \geq \epsi{4\rho}}\led{172} \exp{(-p)},
	\end{equation*}
	where \ding{172} uses $\rho<\rho_2 < \frac{\rho_0}{4}$ with $\rho_0=\frac{1-\delta}{4}$ in Lemma~\ref{lemma2} and sets $\epsi{}$ sufficient small such that $\epsi{\rho_{0}-4\rho} \leq c_1$. This is because if $\epsi{\rho_0} \leq \frac{\epsi{4\rho}}{c_1}$, then it yields  $\PPi{[\epsi{} \xwi{}]_t^d \geq \epsi{4\rho}/c_1} \leq \exp{(-p)}$ due to $\PPi{[\epsi{} \xwi{}]_t^d \geq \epsi{\rho_0}}\leq \exp{(-p)}$. So the result in this lemma holds with $\rho_0=\min(\rho_0=\frac{1-\delta}{4}, \rho_1, \rho_2) =\frac{1-\delta}{16}$, $p_0=\min(p_0=\frac{\rho}{2}, p_1)=\frac{\delta}{2}$. The parameters $\rho_0$ and $p_0$ in the operator $(\cdot)$ are from Lemma~\ref{lemma2} as the results here is based on Lemma~\ref{lemma2}. Under this setting, we have 
	\begin{equation*}
	\PPi{\sup_{0\leq t \leq \Te}\epsi{} \left| \sum_{i=1}^{d} \int_{0}^{t}  \gmii{s-}{i} \ds \xwii{s}{i} \right|\geq \epsi{\rho}} \leq 2\exp\left(-\epsi{-p}\right)
	\end{equation*}
	The proof is completed. 
\end{proof}

\subsection{Proof of Lemma~\ref{lemm4}}\label{proofoflemm4}

\begin{proof}
	\step{Step 1.} In this step we prove the sequence  $\{\wgi{t}\}$ produced by Eqn.~\eqref{deterministicversion_RMSP} or \eqref{deterministicversion_adam} locates in a very small neighborhood of the optimum solution $\wms$ of the local basin $\G$ after a very small time interval. Since we assume the function is locally strongly convex, by using Theorems~\ref{linearconvergenceSGD} and~\ref{linearconvergenceadam}, we know that the sequence $\{\wgi{t}\}$ produced by Eqn.~\eqref{deterministicversion_RMSP} or \eqref{deterministicversion_adam} exponentially converges to the minimum $\wms$ at the current local basin $\G$. So for any initialization $\wmi{0}\in\G$, we have  
	\begin{equation*}
	\begin{split}
	\|\wgi{t}-\wms\|_2^2  \leq c_1\exp\left(-c_2 t \right),
	\end{split}
	\end{equation*}
	where $c_1= \frac{2\Delta}{\mu} $ and $c_2=\frac{2\mu\tau}{ \betag \left(\vmax+\thres\right) +\mu\tau} \left(\betag -\frac{\betav}{4} \right)$ in \Adam,   
	$c_1= \frac{2\Delta}{\mu} $ and $c_2=2\mu$ in \Sgd. Therefore, for any initialization $\wmi{0}\in\G$ and sufficient small $\epsi{}$, we can obtain
	\begin{equation*}
	\begin{split}
	\|\wgi{t}-\wms\|_2^2  \leq \epsi{\rhoa} \ \text{when}\ t\geq \ve = \frac{1}{c_2} \ln\left(\frac{c_1}{\epsi{\rhoa}}\right).
	\end{split}
	\end{equation*}

	\step{Step 2.} Here we prove that for the time $t\in[0,\ve]$, the sequence $\{\wmi{t}\}$ is always very close to the sequence $\{\wgi{t}\}$ when they are with the same initialization $\wmi{0}$ in the absence of the big jumps $\jumpi{k}$ in the stochastic process $\levy$.
	
	To begin with, according to the updating rule in \Sgd, we have
	\begin{equation}\label{afsafvcsacsa}
	\begin{split}
	\|\wmi{\ttt-}-\wgi{\ttt-}\|=& \left| \int_{0}^{\ttt-}\left(- \nabla \Fm(\wmi{s}) +   \nabla \Fm(\wgi{s}) \right)\ds s+ \int_{0}^{\ttt-}  \varepi{} \Sigmai{s}\ds \levyi{s} \right|\\
	\led{172} & \ell \int_0^{\ttt-} \|\wmi{s}-\wgi{s}\|\ds s + \epsi{}\left\|\int_0^{\ttt-} \Sigmai{s} \ds \levyi{s} \right\|,
	\end{split}
	\end{equation}
	where in \ding{172}, $\Fm(\wm)$ is $\ells$-smooth, namely $\|\nabla \Fm(\wmi{1})-\nabla \Fm(\wmi{2})\|\leq \ells\|\wmi{1}-\wmi{2}\|$ for any $\wmi{1}$ and $\wmi{2}$ in the local basin $\G$. 
	
	Then we consider \Adam~which needs more efforts.   According to the dynamic system of \Adam, we can first establish 
	\begin{equation*}
	\begin{split}
	\mmi{t} - \mgi{t} =  \int_{0}^{t}  (\nabla \Fm(\wmi{s}) - \nabla \Fm(\wgi{s})) \ds s - \int_{0}^{t}  (\mmi{s}-\mgi{s}) \ds s.
	\end{split}
	\end{equation*}
	Therefore, with the assumption $\|\mmi{t}-\mgi{t}\|\leq \taum \|\int_{0}^{t}(\mmi{s}-\mgi{s})\ds s\|$, it yields
	\begin{equation*}
	\begin{split}
	|1-\taum|\cdot \left\|\int_{0}^{t}  (\mmi{s}-\mgi{s}) \ds s \right\| \leq & \left\|\mmi{t} - \mgi{t} + \int_{0}^{t}  (\mmi{s}-\mgi{s}) \ds s \right\| =  \left\|\int_{0}^{t}  (\nabla \Fm(\wmi{s}) - \nabla \Fm(\wgi{s})) \ds s \right\|\\
	\leq \ells \int_{0}^{t}   \left\|\wmi{s} - \wgi{s}  \right\|\ds s.
	\end{split}
	\end{equation*}
	Moreover, we can upper bound $\frac{\mui{s}}{\sqrt{\omegai{s}\vmi{s}}+ \thres}=\frac{\sqrt{1-e^{-\betav t}}}{1-e^{-\betag t}} \cdot \frac{1}{1+ \epsi{} \sqrt{1-e^{-\betav t}}}$. Then let $q(x)=\frac{\sqrt{1-e^{-\betav t}}}{1-e^{-\betag t}}\leq c_4 =\min(q(0), q(+\infty), q(t^*))$,  where $t^*$ is a time such that $q'(t^*)=0$. Since $q(0)=\frac{\betav}{2\betag}$ by LHopital's rule,   $q(+\infty)=1$ and $q(t^*)<\infty$ is a constant,  $c_4<\infty$ is a constant. So there exists a constant $c_5$ such that $\frac{\mui{s}}{\sqrt{\omegai{s}\vmi{s}}+ \thres}\leq \frac{c_5}{\vmin+\thres}$. Then similarly, in \Adam, we also can establish 
	\begin{equation*}
	\begin{split}
	\|\wmi{\ttt-}-\wgi{\ttt-}\|=& \left| \int_{0}^{\ttt-}\left(-\frac{\mui{s}\mmi{s}}{\sqrt{\omegai{s}\vmi{s}}+ \thres}  +\frac{\mui{s}\mgi{s}}{\sqrt{\omegai{s}\vgi{s}}+\thres}\right)\ds s+ \int_{0}^{\ttt-}  \varepi{}\Qmi{s}^{-1} \Sigmai{s}\ds \levyi{s} \right|\\
	\led{172} & \frac{c_5\ell}{(\vmin+\thres) |\taum-1|} \int_0^{\ttt-} \|\wmi{s}-\wgi{s}\|\ds s + \epsi{}\left\|\int_0^{\ttt-} \Qmi{s}^{-1} \Sigmai{s} \ds \levyi{s} \right\|.
	\end{split}
	\end{equation*}
	Next,  we can employ Gronwall's to estimate 
	\begin{equation*}
	\begin{split}
	\sup_{0\leq t \leq \jumptime{1} \land \ve }\|\wmi{t}-\wgi{t}\|\leq \exp{\left(\clip\ve\right)}  \sup_{0\leq t \leq   \ve } \epsi{}\left\|\int_0^{t} \Qmi{s}^{-1} \Sigmai{s} \ds \xii{s} \right\|,
	\end{split}
	\end{equation*}
	where $\clip=\ell$ in \Sgd,  and $\clip=\frac{c_5\ells}{(\vmin+\thres) |\taum-1|} $ in \Adam. Since when $\epsi{}$ is small enough, $\ve= \frac{1}{c_2} \ln\left(\frac{c_1}{\epsi{\rhoa}}\right)$ is much smaller than $\Te=\epsi{-\theta}$ when $\epsi{}$ is sufficient small. It yields 
	\begin{equation*}
	\begin{split}
	\PPi{ \sup_{0\leq t \leq \jumptime{1} \land \ve }\|\wmi{t}-\wgi{t}\|\geq \epsi{\rhoa}} \leq& \PPi{\exp{\left(\clip\ve\right)}  \sup_{0\leq t \leq   \ve } \epsi{}\left\|\int_0^{t} \Qmi{s}^{-1} \Sigmai{s} \ds \xii{s} \right\|\geq \epsi{\rhoa}} \\
	\led{172}& \PPi{ \sup_{0\leq t \leq   \ve } \epsi{}\left\|\int_0^{t} \Qmi{s}^{-1} \Sigmai{s} \ds \xwi{s}   \right\|  +\epsi{} \|\mue \| \Te \geq \epsi{\rhoa+c_3\clip \rhoa}}\\
	= & \PPi{ \sup_{0\leq t \leq   \ve } \epsi{}\left\|\int_0^{t} \Qmi{s}^{-1} \Sigmai{s} \ds \xwi{s}   \right\|  \geq \epsi{\rho}(\epsi{\rhoa(1+c_3\clip)- \rho}-\epsi{\rho}}\\
	\led{173}&\exp(-p),
	\end{split}
	\end{equation*}
	where \ding{172} uses Lemma~\ref{lemma2}: (1) the process $\xii{}$ can be decomposed into two processes $\xwi{}$ and linear drift, namely, $\xii{t}=\xwi{t}+\mue t,$ 
	where $\xwi{}$ is a zero mean \levyp martingale with bounded jumps;  (2)  $
	\|\epsi{}\xii{\Te}\| = \epsi{} \|\mue\|\Te < \epsi{2\rho}$.  	In \ding{173}, (1) we set $\rhoa(1+c_3\clip) < \rho$ and also set $\epsi{}$ sufficient small such that $\epsi{\rhoa(1+c_3\clip)- \rho}-\epsi{\rho}\geq 1$; (2) by assume $\rho_0=\rho_0(\delta)=\frac{1-\delta}{16}>0$, $\theta_0=\theta_0(\delta)=\frac{1-\delta}{3}>0$ and $p_0=p_0(\rho)=\frac{\rho}{2}$, we use Lemma~\ref{lemma3} by setting $\gmi{t}=\Qmi{t}^{-1} \Sigmai{t}$ and obtain $\PPi{\sup_{0\leq t \leq \Te}\epsi{}\left| \sum_{i=1}^{d} \int_{0}^{t}  \gmii{s-}{i} \ds \xwii{s}{i} \right|\geq \epsi{\rho}} \leq 2\exp\left(-\epsi{-p}\right)$ for all $p\in(0, p_0]$  and $0<\epsi{}\leq \epsi{}_0$ with $\epsi{}_0 = \epsi{}_0(\rho)$.

	\step{Step 3.} In the first step, we have analyzed that the sequence $\{\wgi{t}\}$ will converge to the optimum $\wms$ of the basin $\G$. Moreover, in the second step, we prove that  $\wmi{t}$ is very close to $\wgi{t}$.  In this step, we show that in absence of the big jumps of the driving process $\levy$ the sequence $\wmi{t}$ is close to $\wms$. For brevity, we set $\wms=\bm{0}$.   	Then we define a function $h(\wm)=\ln(1+\Fm(\wm))\geq 0$. Since for a small local convex basin $\G$, the function $\Fm(\wm)$ can be well approximated by a quadratic function. In this way, for small $\wm$ one can always estimate $c_6\|\wm\|^2\leq h(\wm)\leq c_7 \|\wm\|^2$ for some positive constants $c_6$ and $c_7$. Furthermore, the derivatives $\partial_i h(\wm)=\frac{\partial_i \Fm(\wm)}{1+\Fm(\wm)}$ and $\partial_i \partial_j h(\wm) = \frac{\partial_{ij}\Fm(\wm) (1+\Fm(\wm)) - \partial_{i}\Fm(\wm)\partial_j \Fm(\wm)}{(1+\Fm(\wm))^2}$ are bounded since the assumptions on the function $\Fm(\wm)$, namely $\Fm(\wm)$ being upper bounded, $\ells$-smooth. Next we can apply the It$\hat{\text{o}}$ formulation to the process $h(\wmi{t})$:
	\begin{equation*}
	\begin{split}
	0\leq & h(\wmi{\ttte-}) = h(\wm) + \sum_{i=1}^{d} \int_{0}^{\ttte-} \partial_i h(\wmi{s-}) \ds \wmi{s-}^i + \frac{1}{2} \sum_{i,j=1}^{d} \int_{0}^{\ttte-} \partial_i \partial_j h(\wmi{s-}) \ds [\wm^i, \wm^j]_s^c\\
	&\qquad \qquad \qquad \quad+ \sum_{s < \ttte} \left(h(\wmi{s})-h(\wmi{s-}) - \sum_{i=1}^d\partial_i h(\wmi{s-})\Delta\wmi{s}^i\right)\\
	\led{172} & h(\wm) -\int_{0}^{\ttte-} \left\langle \frac{\nabla \Fm(\wmi{s-})}{1+ \Fm(\wmi{s-})}, \frac{\mui{t}\mmi{s-}}{\thres+\sqrt{\omegai{s-}\vmi{s-}}}  \right\rangle\ds s + \int_{0}^{\ttte-}  \frac{\varepi{} (\nabla \Fm(\wmi{s-}))^T\Qmi{s-}^{-1} \Sigmai{s-}}{1+ \Fm(\wmi{s-})} \ds \levyi{s} \\
	&\qquad \qquad \qquad \quad+ \sum_{s < \ttte} \left(h(\wmi{s})-h(\wmi{s-}) - \sum_{i=1}^d\partial_i h(\wmi{s-})\Delta\wmi{s}^i\right),
	\end{split}
	\end{equation*}
	where \ding{172} uses $\ds \wmi{s} = -\frac{\mui{s}\mmi{s}}{\thres+\sqrt{\omegai{s}\vmi{s}}} + \varepi{}\Qmi{s}^{-1} \Sigmai{s}\ds \levyi{s}$ and the path-by-path continuous part $[\wm^i, \wm^j]_s^c=0$ of the quadratic covariation of $\wm^i$ and $\wm^j$.   Since in Adam by assumption $\int_{0}^{\ttte-} \left\langle \frac{\nabla \Fm(\wmi{s-})}{1+ \Fm(\wmi{s-})}, \frac{\mui{t}\mmi{s-}}{\thres+\sqrt{\omegai{s-}\vmi{s-}}}  \right\rangle\ds s\geq 0$, the second term is non-negative due to $\Fm(\wm)\geq 0$. Note in  \Sgd, $\mmi{s}=\nabla \Fm(\wmi{s})$. So in \Sgd~we do not make the  assumption $\langle \nabla \Fm(\wmi{t}), \mmi{t}\rangle \geq 0$. In \Sgd, $\thres+\sqrt{\omegai{s-}\vmi{s-}}$ equals to one.  In this way, we can estimate the last term as 
	\begin{equation*}
	\begin{split}
	& \sum_{s < \ttte} \left|h(\wmi{s})-h(\wmi{s-}) - \sum_{i=1}^d\partial_i h(\wmi{s-})\Delta\wmi{s}^i\right| \\
	\leq&  \frac{1}{2} \sum_{i,j=1}^d \sum_{s < \ttte} \left|\int_{0}^{1} (1-v) \partial_i\partial_j h(\wmi{s-} + v \Delta \wmi{s}) \ds v  \right|  \cdot |\Delta \wmi{s}^i\Delta \wmi{s}^j|\leq c_8 \sum_{s\leq t} \|\Delta \wmi{s}\|^2 =c_8 [ \wmi{}]_t^d,
	\end{split}
	\end{equation*}
	holds with some $c_8>0$. Furthermore, since $\vmi{t}$ and $\Sigmai{t}$ are assumed to be bounded, then we can upper bound $[ \wmi{}]_t^d$ as follows:
	\begin{equation*}
	\begin{split}
	[ \wmi{}]_t^d \leq c_9 [\epsi{}\levy]_t^d \lee{172} c_9 [\epsi{}\xii{}]_t^d
	\end{split}
	\end{equation*}
	hold for some constant $c_9$ for all $t\leq \jumptime{1}$. \ding{172} holds since we assume there is no big jump during $t\leq \jumptime{1}$. Then by combining all the results and letting $\gmi{s}=\frac{ (\nabla \Fm(\wmi{s}))^T\Qmi{s}^{-1} \Sigmai{s}}{1+ \Fm(\wmi{s})}$ and considering $F(\wm)\leq c_7 \|\wm\|$, we can obtain the following results when $\|\wm\|=\|\wmi{0}\|\leq \epsi{\rhoa}$ with enough small $\epsi{}$:
	\begin{equation*}
	\begin{split}
	0\leq \|\wmi{\ttte-}\|^2 \leq \frac{1}{c_6} h(\wmi{\ttte-})  \leq c_{10} \left(\epsi{2\rhoa} + \epsi{} \sup_{0\leq t\leq \Te} \left|\int_{0}^{t} \gmi{s-}\ds \xwi{s} \right| + \epsi{} \|\mue\|\Te +\epsi{2} [\xii{}]_{\Te}^d\right). 
	\end{split}
	\end{equation*}
	where $c_{10}$ is a certain constant. Combining the above results gives 
	\begin{equation*}
	\begin{split}
	\PPi{\sup_{0\leq t \leq \Te \land \jumptime{1}} \|\wmi{t}\|\geq \epsi{\rhoa} } \leq &\PPi{\epsi{2\rhoa} \geq \frac{\epsi{\rhoa}}{4c_{10}}}  + \PPi{ \epsi{} \sup_{0\leq t\leq \Te} \left|\int_{0}^{t} \gmi{s-}\ds \xwi{s} \right|\geq \frac{\epsi{\rhoa}}{4c_{10}}} \\
	&+ \PPi{ \epsi{} \|\mue\|\Te \geq \frac{\epsi{\rhoa}}{4c_{10}}}  + \PPi{\epsi{2} [\xii{}]_{\Te}^d \geq \frac{\epsi{\rhoa}}{4c_{10}}}.\\
	\end{split}
	\end{equation*}	
	Then by setting $\rhoa<\rho$ and sufficient small $\epsi{}$ such that $\frac{\epsi{\rhoa-\rho}}{4c_{10}}\geq 1$ giving $\frac{\epsi{\rhoa}}{4c_{10}}\geq \epsi{\rho}$. 
	Then let the results in Lemma~\ref{lemma2} and \ref{lemma3} hold simultaneously by setting $\rho_0=\rho_0(\delta)=\frac{1-\delta}{16}>0$, $\theta_0=\theta_0(\delta)=\frac{1-\delta}{3}>0$,  $p_0=\frac{\rho}{2}$, and small enough $\epsilon$, we have 	$\|\epsi{}\xii{\Te}\| = \epsi{} \|\mue\|\Te < \epsi{2\rho}\quad \text{and}\quad \PPi{[\epsi{} \xii{}]_{\Te}^d\geq \epsi{\rho}} \leq \exp(-\epsi{-p})$ in Lemma~\ref{lemma2}, and $\PPi{\sup_{0\leq t \leq \Te}\epsilon \left| \sum_{i=1}^{d} \int_{0}^{t}  \gmii{s-}{i} \ds \xwii{s}{i} \right|\geq \epsi{\rho}} \leq 2\exp\left(-\epsi{-p}\right)$ in Lemma~\ref{lemma3}. By using these results, we have 
	\begin{equation*}
	\begin{split}
	\PPi{\sup_{0\leq t \leq \Te \land \jumptime{1}} \|\wmi{t}\|\geq \epsi{\rhoa} } \leq  4\exp(-\epsi{-p}).
	\end{split}
	\end{equation*}	
	for all $p\in(0,p_0]$ and $0<\epsi{}\leq \epsi{}_0$ with $\epsilon_0 = \epsi{}_0(\rho)$.

	\step{Step 4.} In Steps 1 and 2, we guarantee $\PPi{\sup_{0\leq t \leq \ve \land \jumptime{1}} \|\wmi{t}-\wgi{t}\|\geq \epsi{\rhoa} } \leq 4 \exp(-\epsi{-p})$. Then after $\ve$ time, we have $\|\wmi{t}\|\leq \epsi{\rhoa}$ for all $t\geq \ve$. In this way, the result in Step 4 holds. So in this step, we combine the results in Steps 1, 2 and 3 and extend the initialization in Step 3 to all possible parameter in $\wmi{0}\in\G$:
	\begin{equation*}
	\begin{split}
	\PPi{\sup_{0\leq t \leq \ve \land \jumptime{1}} \|\wmi{t}-\wgi{t}\|\geq \epsi{\rhoa} } \leq 4\exp(-\epsi{-p}), 
	\end{split}
	\end{equation*}	
	for all $p\in(0,p_0]$ and $0<\epsi{}\leq \epsi{}_0$ with $\epsi{}_0 = \epsi{}_0(\rho)$  by setting  $\rho_0=\rho_0(\delta)=\frac{1-\delta}{16}>0$, $\theta_0=\theta_0(\delta)=\frac{1-\delta}{3}>0$,  $p_0=\frac{\rho}{2}$, $\rhoa(1+c_3\clip) < \rho$ and small enough $\epsi{}$.  Note here we can remove the extra factor $\rho$ by setting $\epsi{}_0 = \epsi{}_0(\rhoa)$, $\rho_0=\rho_0(\delta)=\frac{1-\delta}{16(1+c_3\clip)}>0$, $\theta_0=\theta_0(\delta)=\frac{1-\delta}{3}>0$,  $p_0=\frac{\rhoa (1+c_3\clip)}{2}$, $p\in(0,p_0)$.

	\step{Step 5.}  In this step, we extend the result in Step 4 from the time interval $[0,\Te\land \jumptime{1})$ to the time interval $[0, \jumptime{1})$.

	Let $\wmxi{t}$ denote the sequence produced by  \Sgd~\eqref{assumption_stochastic_RMSP} or  Adam~\eqref{assumption_stochastic_adam}  driven by the process $\xii{}$. Then it is easy to check that for any $t< \jumptime{1}$, we have $\wmxi{t} = \wmi{t}$, since there are no big jumps in $\wmi{t}$. Then consider any $\wmi{0}\in\G$ and $k\geq 1$, we have for any $\rhoa>0$ and $\theta>0$
	\begin{equation*}
	\begin{split}
	&\PPi{\sup_{0\leq t < \jumptime{1}} \|\wmi{t}-\wgi{t}\|\geq  \epsi{\rhoa}} \\
	\leq & \PPi{\sup_{0\leq t < k\Te \land \jumptime{1}} \|\wmi{t}-\wgi{t}\|\geq  \epsi{\rhoa}\ \big| \ k\Te < \jumptime{1}} + \PPi{\sup_{0\leq t < \jumptime{1}} \|\wmi{t}-\wgi{t}\|\geq  \epsi{\rhoa}\ \big| \ k\Te \geq \jumptime{1}}\\
	\leq & \PPi{\sup_{0\leq t < k\Te \land \jumptime{1}} \|\wmi{t}-\wgi{t}\|\geq  \epsi{\rhoa}} + \PPi{k\Te \geq \jumptime{1}}\\
	\leq & \PPi{\sup_{0\leq t <  k\Te} \|\wmi{t}-\wgi{t}\|\geq  \epsi{\rhoa}} + \PPi{k\Te \geq \jumptime{1}}.
	\end{split}
	\end{equation*}	
	Besides, by using the linear convergence results of  $\wgi{t}$ to the optimum solution $\wms=\bm{0}$ in the local basin $\G$, for enough small  $\epsi{}$ we have $\|\wgi{\Te}\|\leq \epsi{2\rhoa}$ with initialization $\wmi{0}\in\G$. Then we  let $\wgti{t}{\wm}$ denote the sequence $\wgi{t}$ but with initialization $\wm$ and define 
	\begin{equation*}
	\begin{split}
	\Emi{i} = \left\{\sup_{t \in [i\Te, (i+1)\Te]} \|\wmxi{t}-\wgti{t-i\Te}{\wmxi{i\Te}}\|< \epsi{\rhoa}\right\},\quad 0\leq i \leq k-1.
	\end{split}
	\end{equation*}	
	Note that the probability of $\Emi{0}^c= \left\{\sup_{t \in [0, \Te]} \|\wmxi{t}-\wgti{t}{\wmxi{0}}\|\geq  \epsi{\rhoa}\right\}$ is given in Step 4 where $\wmxi{0}=\wmi{0}$. Furthermore for any $k\geq 1$, we have 
	\begin{equation*}
	\begin{split}
	\bigcap_{i=0}^{k-1} \Emi{i} \subseteq \left\{\sup_{t \in [0,  k\Te]} \|\wmxi{t}-\wgi{t}\|<2 \epsi{\rhoa}\right\}.
	\end{split}
	\end{equation*}	
	As a result, we can obtain
	\begin{equation*}
	\begin{split}
	\PPi{ \sup_{t \in [0,  k\Te]} \|\wmxi{t}-\wgi{t}\|\geq 2 \epsi{\rhoa}} \leq & \PPi{\bigcup_{i=0}^{k-1} \Emi{i}^c} 
	= \PPi{\Emi{0}^c \bigcup (\Emi{0}\Emi{1}^c) \bigcup (\Emi{0}\Emi{1}\Emi{2}^c) \bigcup \cdots (\bigcup_{i=0}^{k-2} \Emi{i} \Emi{k-1}^c)} \\
	\leq &  \sum_{i=0}^{k-1} \PPi{\Emi{i}^c, \wmxi{i\Te} \in \G} \leq k \sup_{\wmi{0}\in\G} \PPi{\Emi{0}^c}.
	\end{split}
	\end{equation*}	
	For $k=\ke=\epsi{-2r}$ and any $\theta>0$ we have 
	\begin{equation*}
	\begin{split}
	\PPi{\jumptime{1}\geq \ke \Te} =\exp(-\ke \Te \psiis{\epsilon}{\delta} )\leq \exp(-\epsi{r\delta-\theta-2r} \Hui{\epsi{-\delta}}) \leq \exp(-\epsi{-p})
	\end{split}
	\end{equation*}	
	for all $0<p\leq (2-\delta)r$ with enough small $\epsi{}$. On the other hand, we have 
	\begin{equation*}
	\begin{split}
	\PPi{ \sup_{t \in [0,  k\Te]} \|\wmxi{t}-\wgi{t}\|\geq 2 \epsi{\rhoa}} \leq k \sup_{\wmi{0}\in\G} \PPi{\Emi{0}^c} \leq \epsi{-2r} \exp(-\epsi{-p}) \leq \exp(-\epsi{-p/2})
	\end{split}
	\end{equation*}	
	for any $p\leq \frac{2\log(r\log(\epsi{})))}{\log(\epsi{})}$. Therefore, the result in this lemma holds 
	\begin{equation*}
	\begin{split}
	&\PPi{ \sup_{t \in [0,  \jumptime{1})} \|\wmxi{t}-\wgi{t}\|\geq 2 \epsi{\rhoa}} \\
	= & \PPi{ \sup_{t \in [0,  \jumptime{1})} \|\wmxi{t}-\wgi{t}\|\geq 2 \epsi{\rhoa},\ \jumptime{1} < k \Te}  +\PPi{ \sup_{t \in [0,  \jumptime{1})} \|\wmxi{t}-\wgi{t}\|\geq 2 \epsi{\rhoa},\ \jumptime{1} \geq k \Te}\\
	\leq  & \PPi{ \sup_{t \in [0,  k\Te ]} \|\wmxi{t}-\wgi{t}\|\geq 2 \epsi{\rhoa}}  +\PPi{ \jumptime{1} \geq k \Te} \leq 2 \exp(-\epsi{-p/2}).
	\end{split}
	\end{equation*}	
	for all $p\in(0,p_0]$ and $0<\epsi{}\leq \epsi{}_0$ with $\epsi{}_0 = \epsi{}_0(\rhoa)$  by setting  $\rho_0=\rho_0(\delta)=\frac{1-\delta}{16(1+c_3\clip)}>0$, $\theta_0=\theta_0(\delta)=\frac{1-\delta}{3}>0$,  $p_0=\min(\frac{\rhoa (1+c_3\clip)}{2}, p)$ with $p>0$ and small enough $\epsi{}$.  Besides, we also require $\ve= \frac{1}{c_2} \ln\left(\frac{c_1}{\epsi{\rhoa}}\right) = \frac{1}{c_2} \ln\left(\frac{2\Delta}{\mu \epsi{\rhoa}}\right)  \leq \epsi{-\theta_0}$ where $c_1= \frac{2\Delta}{\mu} $ and $c_2=\frac{2\mu\tau}{ \betag \left(\vmax+\thres\right) +\mu\tau} \left(\betag -\frac{\betav}{4} \right)$ in \Adam,  	$c_1= \frac{2\Delta}{\mu} $ and $c_2=2\mu$ in \Sgd. That is,  The proof is completed. 
\end{proof}

\end{document}